\documentclass{article} 
\usepackage{iclr2027_conference,times}

\usepackage{amsmath,amsfonts,bm}

\def\eqref#1{equation~\ref{#1}}

\def\1{\bm{1}}

\DeclareMathAlphabet{\mathsfit}{\encodingdefault}{\sfdefault}{m}{sl}
\SetMathAlphabet{\mathsfit}{bold}{\encodingdefault}{\sfdefault}{bx}{n}

\usepackage{url}

\usepackage{wrapfig}
\usepackage{amssymb}
\usepackage{amsmath}
\usepackage{multirow}
\usepackage{makecell}
\usepackage{siunitx}
\usepackage{bbding}
\usepackage{blindtext}
\usepackage{multicol}
\usepackage{graphicx}
\usepackage{booktabs}
\usepackage{subfloat}
\usepackage{caption}
\usepackage{subfiles}
\usepackage{subcaption}
\usepackage{enumitem}
\usepackage{tabu}
\usepackage[table]{xcolor}
\usepackage[linesnumbered,ruled]{algorithm2e}

\usepackage{natbib}
\usepackage{bibentry}

\definecolor{lightred}{rgb}{0.9, 0.47, 0.37}

\definecolor{paleblue}{RGB}{207, 240, 255}
\definecolor{palerowgreen}{RGB}{235,255,235}
\definecolor{resultgreen}{RGB}{45, 150, 105}

\makeatletter
\newcommand{\ssymbol}[1]{^{\@fnsymbol{#1}}}
\makeatother

\usepackage{tocloft}
\usepackage{etoc}
\usepackage{titletoc}
\usepackage{calc}
\usepackage{diagbox}
\usepackage{slashbox}

\usepackage{pifont}

\newcommand{\cmark}{\ding{51}}
\newcommand{\xmark}{\ding{55}}
\newcommand{\redx}{\textcolor{red}{\xmark}}

\usepackage{alphalph}

\usepackage{hyperref}

\title{DocLayout-YOLO: Axis-Attentive Document Layout Detector with Mesh-Guided \\ Synthetic Pretraining}

\author{Zhiyuan Zhao, Hengrui Kang, Guang Liang, Bin Wang, Conghui He  \\
Shanghai AI Laboratory \\
}

\iclrfinalcopy 
\begin{document}

\maketitle

\begin{figure}[h]
    \centering
    \vspace{-25pt}
    \includegraphics[width=0.97\linewidth]{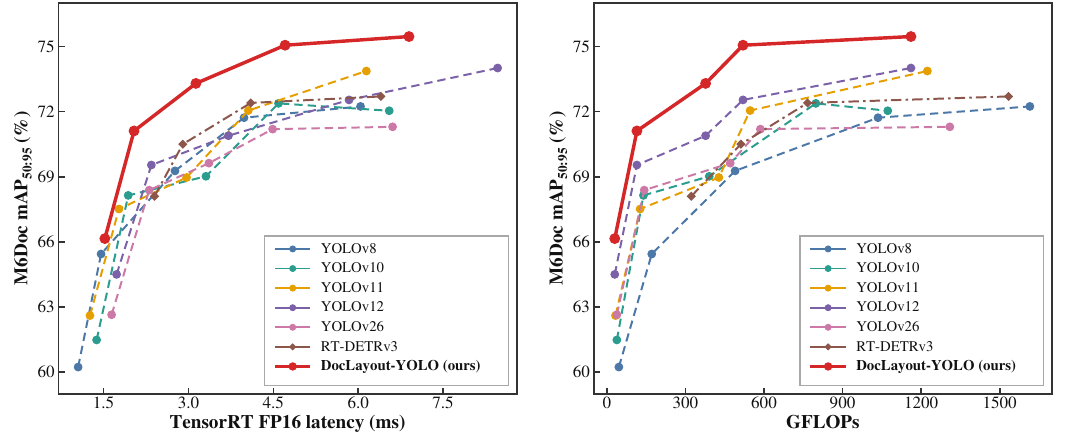}
    \vspace{-8pt}
    \caption{Comparisons with popular real-time detectors in terms of latency-accuracy (left) and FLOPs-accuracy (right) trade-offs on M\textsuperscript{6}Doc.}
    \label{fig:fig1}
\end{figure}

\begin{abstract}
Accurate and efficient document layout analysis is critical for modern document parsing systems. However, although the YOLO framework represents a leading paradigm for the real-time detector, its adaptation to heterogeneous documents remains underexplored. To address this gap, we propose DocLayout-YOLO, a document-specialized real-time detector that integrates diverse synthetic document pretraining and page-axis context modeling. From the data perspective, we introduce {MeshLayout-1M}, a synthetic pretraining corpus generated by the proposed Mesh-candidate BestFit algorithm, which provides large-scale and diverse document layouts for pretraining and improves adaptation to heterogeneous document scenarios. From the model perspective, we propose {Page-Axis Layout Attention (PALA)}, which captures long-range dependencies along the horizontal and vertical page axes while remaining computationally efficient. Extensive experiments demonstrate that DocLayout-YOLO achieves state-of-the-art performance on four heterogeneous document benchmarks, obtaining 72.4\%/75.5\%/96.0\%/73.1\% mAP$_{50:95}$ on D\textsuperscript{4}LA, M\textsuperscript{6}Doc, HJDataset, and PRIMA-LAD, respectively. Compared with existing real-time detectors, DocLayout-YOLO establishes a superior accuracy-efficiency trade-off in document layout analysis. Code, model, and data are available at \url{https://github.com/opendatalab/DocLayout-YOLO/tree/release-v2}.
\end{abstract}

\section{Introduction}

Document Layout Analysis (DLA)~\citep{DBLP:conf/icdar/ZhongTJ19,DBLP:conf/kdd/PfitzmannADNS22,DBLP:conf/cvpr/ChengZWZZXLDJ23} aims to locate and classify semantic regions such as text, tables, and figures, providing essential structural information for document parsing and understanding~\citep{wang2026mineru25pro,cui2026paddleocrvl,li2026realworld,yu2026mact}. With the rapid development of large language models (LLMs)~\citep{DBLP:journals/corr/abs-2303-08774,yang2025qwen3} and retrieval-augmented generation (RAG)~\citep{yu2024visrag,cho2024m3docrag,chen2025documenthaystacks}, the demand for high-quality structured document data has increased substantially, requiring real-time document detectors to achieve both higher accuracy and efficiency.

Among existing real-time object detectors, the YOLO series~\citep{jocher2026ultralyticsyolo26,tian2026yolov12,Jocher_Ultralytics_YOLO_2023} achieves a strong accuracy-efficiency trade-off, yet its potential for document layout analysis remains largely unexplored. However, adapting YOLO detectors to document scenarios is non-trivial due to two critical challenges. At the data level, adapting detectors to document scenarios requires effective document-domain pretraining, which relies on large-scale and structurally diverse corpora covering rich layout variations. However, existing datasets face a trade-off among dataset scale, layout diversity, annotation quality, and so on~(as shown in Tab.~\ref{tab:main1}), while constructing large-scale annotated document corpora with fine-grained layouts remains costly~\citep{DBLP:conf/icdar/ZhongTJ19,DBLP:conf/coling/LiXCHWLZ20}. At the model level, accurately localizing document elements requires effective structural context modeling, as document layouts contain complex long-range dependencies among text lines, paragraph columns, table grids, and figure-caption pairs~\citep{wang2024dlaformer,DBLP:conf/iccv/DaLZY23}. 
However, conventional convolutional YOLO architectures lack explicit mechanisms to capture such dependencies, while the Area Attention recently introduced in YOLOv12~\citep{tian2026yolov12} is primarily designed for efficient generic visual modeling and does not explicitly exploit the horizontal and vertical organization of document pages.

To overcome these challenges, we propose \textbf{DocLayout-YOLO}, a document-specialized real-time detector that improves detection performance through diverse synthetic document pretraining and page-axis context modeling. At the data level, we introduce \textbf{MeshLayout-1M}, a large-scale synthetic pretraining dataset generated by our proposed Mesh-candidate BestFit algorithm based on the principles of 2D Bin-packing~\citep{lodi2002recent}. By synthesizing diverse document layouts with varying elements, scales, and layout structure, MeshLayout-1M provides a more diverse pretraining corpus, allowing better adaptation to heterogeneous document scenarios.

At the model level, we propose \textbf{Page-Axis Layout Attention (PALA)}, which explicitly captures long-range dependencies along the horizontal and vertical axes of document pages. By leveraging inherent page structures, PALA enables efficient context aggregation among layout elements while avoiding the computational overhead of global attention. By combining structure-diverse synthetic pretraining with page-axis context modeling, DocLayout-YOLO achieves a favorable accuracy-efficiency trade-off, as shown in Fig.~\ref{fig:fig1}. The contributions of this paper are summarized as follows:
\begin{itemize}[leftmargin=*,noitemsep,topsep=0pt]
\item We introduce {MeshLayout-1M}, a structure-diverse synthetic pretraining dataset generated by the proposed Mesh-candidate BestFit algorithm, which enables large-scale document-domain pretraining with diverse element and layout organizations.
\item We propose {Page-Axis Layout Attention (PALA)}, an efficient context modeling mechanism that captures long-range dependencies along horizontal and vertical page axes, explicitly exploiting the inherent structural organization of document layouts while maintaining efficient computation.
\item We present {DocLayout-YOLO}, a document-specialized real-time detector integrating synthetic layout pretraining and page-axis context modeling. It achieves state-of-the-art performance on four heterogeneous benchmarks while delivering superior accuracy-efficiency trade-offs over existing real-time detectors.

\end{itemize}

\section{Related Work}

\subsection{Document Layout Analysis Datasets}


Current document datasets are summarized in Tab.~\ref{tab:main1}. Existing datasets have advanced document layout analysis by providing large-scale annotations, such as PubLayNet~\citep{DBLP:conf/icdar/ZhongTJ19} and DocBank~\citep{DBLP:conf/coling/LiXCHWLZ20}, or diverse document types and fine-grained categories, such as DocLayNet~\citep{DBLP:conf/kdd/PfitzmannADNS22}, D\textsuperscript{4}LA~\citep{DBLP:conf/iccv/DaLZY23}, and M\textsuperscript{6}Doc~\citep{DBLP:conf/cvpr/ChengZWZZXLDJ23}. However, existing datasets still face limitations in scale, layout diversity, element coverage, or annotation cost. Meanwhile, existing document generation approaches~\citep{DBLP:conf/iccv/0004GSL023,DBLP:conf/cvpr/InoueKSOY23,DBLP:conf/cvpr/Hui0ZXWL23,DBLP:conf/cvpr/JiangGSDWMYL023,DBLP:conf/eccv/KongJCZHGE22,DBLP:conf/iccv/GuptaLA0MS21} mainly focus on academic-style layouts, limiting their ability to support diverse document-domain pretraining.

\subsection{Document Layout Analysis Approaches}

Existing DLA methods can be broadly categorized into multimodal and unimodal approaches. Multimodal methods, such as LayoutLMv3~\citep{DBLP:conf/mm/HuangL0LW22}, VGT~\citep{DBLP:conf/iccv/DaLZY23}, recently M2Doc~\citep{zhang2024m2doc}, and DoPTA-HR~\citep{DBLP:journals/corr/abs-2412-12902}, improve performance by leveraging text-visual interactions. In contrast, unimodal methods~\citep{shen2021layoutparser,banerjee2023swindocsegmenter,DBLP:conf/icdar/BanerjeeBLP24,biswas2022docsegtr,DBLP:conf/mm/LiXL0ZW22,chen2024rodla} rely primarily on visual features and generic detection architectures. However, these methods either introduce additional multimodal complexity or lack explicit modeling of document-specific structural dependencies. As for real-time detectors, recent YOLO-based real-time detectors~\citep{Jocher_Ultralytics_YOLO_2023,DBLP:journals/corr/abs-2405-14458,tian2026yolov12,jocher2026ultralyticsyolo26} have achieved strong accuracy-efficiency trade-offs through advances in detection architectures and attention mechanisms. Nevertheless, they are primarily optimized for natural images and do not explicitly exploit the structural characteristics of document layouts.

\begin{table*}[t]
\centering
\caption{
Comparison of existing document pretraining datasets.
}
\vspace{-8pt}
\resizebox{0.95\linewidth}{!}{
\begin{tabular}{l|ccccc}
\toprule
\textbf{Dataset}
& \textbf{\#Pages}
& \textbf{\#Elements}
& \textbf{Element Types}
& \textbf{Layout Diversity}
& \textbf{Annotation Type} \\
\midrule

IIT-CDIP~\citep{DBLP:conf/sigir/LewisAAFGH06}
& 42M
& \textcolor{lightred}{-~(None)}
& \textcolor{lightred}{-~(None)}
& 11~(Diverse)
& \textcolor{lightred}{Unannotated} \\

PRIMA-LAD~\citep{DBLP:conf/icdar/AntonacopoulosBPP09}
& \textcolor{lightred}{478}
& \textcolor{lightred}{7,453}
& \textcolor{lightred}{5}
& 5~(Diverse)
& Automatic \\

PubLayNet~\citep{DBLP:conf/icdar/ZhongTJ19}
& 360K
& 3,311,660
& \textcolor{lightred}{5}
& \textcolor{lightred}{1~(Academic)}
& Automatic \\

HJDataset~\citep{shen2020large}
& \textcolor{lightred}{2,271}
& \textcolor{lightred}{250,000+}
& \textcolor{lightred}{7}
& \textcolor{lightred}{4~(Historical)}
& Semi-automatic \\

DocBank~\citep{DBLP:conf/coling/LiXCHWLZ20}
& 500K
& 6,715,907
& 13
& \textcolor{lightred}{1~(Academic)}
& Automatic \\

DocLayNet~\citep{DBLP:conf/kdd/PfitzmannADNS22}
& \textcolor{lightred}{80,863}
& 1,107,470
& {11}
& 6~(Diverse)
& \textcolor{lightred}{Manual} \\

M\textsuperscript{6}Doc~\citep{DBLP:conf/cvpr/ChengZWZZXLDJ23}
& \textcolor{lightred}{9,080}
& \textcolor{lightred}{237,116}
& 74
& 7~(Diverse)
& \textcolor{lightred}{Manual} \\

D\textsuperscript{4}LA~\citep{DBLP:conf/iccv/DaLZY23}
& \textcolor{lightred}{11,092}
& \textcolor{lightred}{293,692}
& 27
& 12~(Diverse)
& \textcolor{lightred}{Manual} \\

IndicDLP~\citep{nath2025indicdlp}
& \textcolor{lightred}{121,198}
& 1,856,241
& 42
& 12~(Diverse)
& \textcolor{lightred}{Manual} \\

\midrule

\textbf{MeshLayout-1M~(Ours)}
& \textbf{\textcolor{resultgreen}{1M}}
& \textbf{\textcolor{resultgreen}{14,549,846}}
& \textbf{\textcolor{resultgreen}{11}}
& \textbf{\textcolor{resultgreen}{Highly Diverse}}
& \textbf{\textcolor{resultgreen}{Automatic}} \\

\bottomrule
\end{tabular}
}
\label{tab:main1}
\end{table*}
\begin{figure*}[t]
    \centering
    \vspace{-10pt}
    \includegraphics[width=0.93\textwidth]{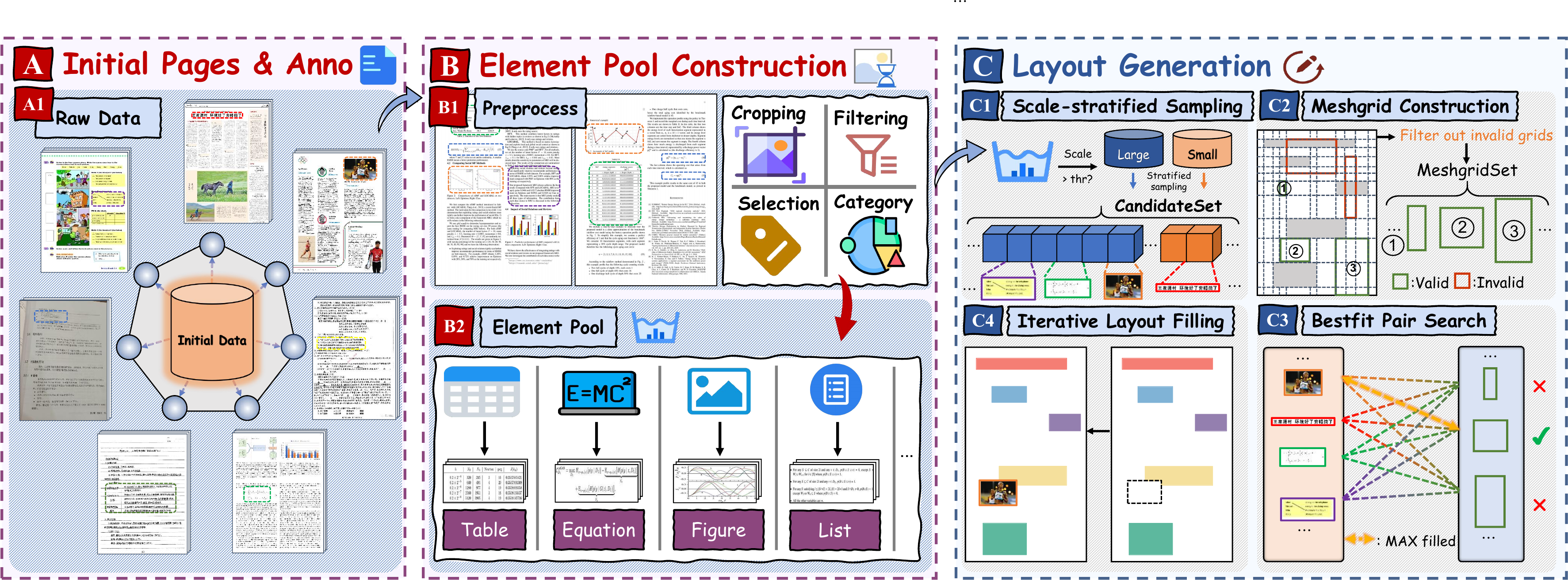}
    \vspace{-5pt}
    \caption{Illustration of Mesh-candidate BestFit. Given initial pages and annotations (\textbf{A}), a category-wise element pool is constructed~(\textbf{B}). Consequently, our method generates diverse and coherent layouts via iterative candidate-grid matching and synthetic pages are formed (\textbf{C}).}
    \label{fig:BestFit}
\end{figure*}

\section{Diverse pretraining Documents Synthesis}

The diversity of pretraining documents arises mainly from two aspects: (1) element diversity, which covers different element categories and scales; and (2) layout diversity, which captures various page organizations such as single-column, multi-column, and magazine-style layouts. To construct a diverse and scalable document pretraining corpus, we propose \textbf{Mesh-candidate BestFit}, a synthesis framework that jointly considers element diversity and layout diversity~(as shown in Fig.~\ref{fig:BestFit}). Unlike random placement or existing generative approaches, our method explicitly controls element composition, scale distribution, and spatial arrangement to generate realistic and diverse document layouts.

\subsection{Preprocessing: Ensuring Element Diversity}


In the preprocessing phase, to ensure the inclusion of a diverse range of document elements, we utilize an initial data corpus that includes document pages and annotations as element source. In the selection of initial data, we use DocBank~\citep{DBLP:conf/coling/LiXCHWLZ20} as the element source due to its diverse element-level annotations, covering various document components, appearances, and scales. 
Consequently, we first construct an element pool from the original annotations. Regions are extracted and noisy excessively small elements are filtered out. Finally, we organize previously retained crops into category-specific pools for subsequent sampling.


\subsection{Layout Generation: Ensuring Layout Diversity}



Given an empty page, Mesh-candidate BestFit iteratively places elements through scale-stratified sampling and mesh-based bestfit matching. Unlike random placement or existing generative approaches, our method balances layout diversity and structural plausibility.

\begin{enumerate}[leftmargin=*,noitemsep,topsep=0pt]
\item \textbf{\textit{Scale-stratified Candidate Sampling}} Directly sampling elements from the entire pool may overpopulate layouts with small elements, preventing the formation of stable global structures. To address this issue, we partition elements into large- and small-scale groups based on their normalized sizes and sample candidates with a predefined ratio. This strategy preserves fine-grained elements while ensuring sufficient large regions for coherent page structures and establishing major page regions.
  \item \textbf{\textit{Meshgrid Construction}} Construct the meshgrid based on the existing layout and filter out the invalid grids that overlap with inserted elements. Only the remaining grids will be able to participate in matching with the candidate in subsequent steps.
  \item \textbf{\textit{BestFit Pair Search}} \ For each candidate, traverse all grids that meet the size requirement and search for the candidate–grid pair with the highest fill rate. Subsequently, remove the optimal candidate from the candidate set and update the layout.
  \item \textbf{\textit{Iterative Layout Filling}} \ Finally, the selected candidate is resized and center-placed to the assigned grid and composited onto the page. Repeat steps 2$\sim$4 until no valid candidate--grid pair satisfies the size requirement or the maximum number of elements is reached.
  
\end{enumerate}


\begin{wrapfigure}{r}{0.55\linewidth}
    \vspace{-10pt}
    \centering
    \includegraphics[width=1.0\linewidth]{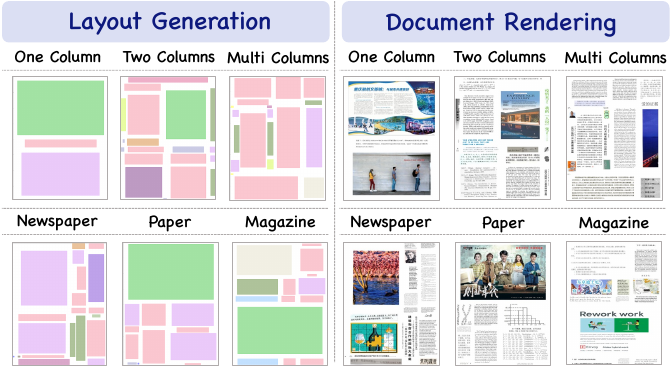}
    \caption{Examples of synthetic document pages.}
    \label{fig:vis0930}
    \vspace{-1.0\baselineskip}
\end{wrapfigure}

Through the process above, elements are progressively placed into suitable regions, producing more regular and compact document layouts, as shown in Fig.~\ref{fig:vis0930}. Meanwhile, the generated layouts also exhibit improved alignment and higher page occupancy. The generated pages from MeshLayout-1M, a large-scale synthetic pretraining corpus containing diverse element and layout organizations, enables pretrained models to adapt to a variety of real-world document types effectively. More detailed analysis and explanations are available in Appendix~\ref{sec:appendix_bestfit},~\ref{sec:appendix_bestfit_vis}.

\section{Page-Axis Layout Attention}



Document layouts exhibit strong structural regularities along the horizontal and vertical page axes, including text lines, paragraph columns, table structures, and cross-region correspondences. Capturing such long-range axis-aware context is important for accurate document element detection. Although global attention~\citep{DBLP:conf/cvpr/0004GGH18} can model these dependencies, its computational cost becomes prohibitive for high-resolution document features. Area Attention in YOLOv12~\citep{tian2026yolov12} improves efficiency by restricting interactions within local regions, but such partitioning may interrupt long-range dependencies across region boundaries. Motivated by the characteristics of document layout, we propose \textbf{Page-Axis Layout Attention (PALA)}, which models long-range document dependencies along the page axes more effectively.

\begin{figure}[t]
    \centering
    \includegraphics[width=1.0\linewidth]{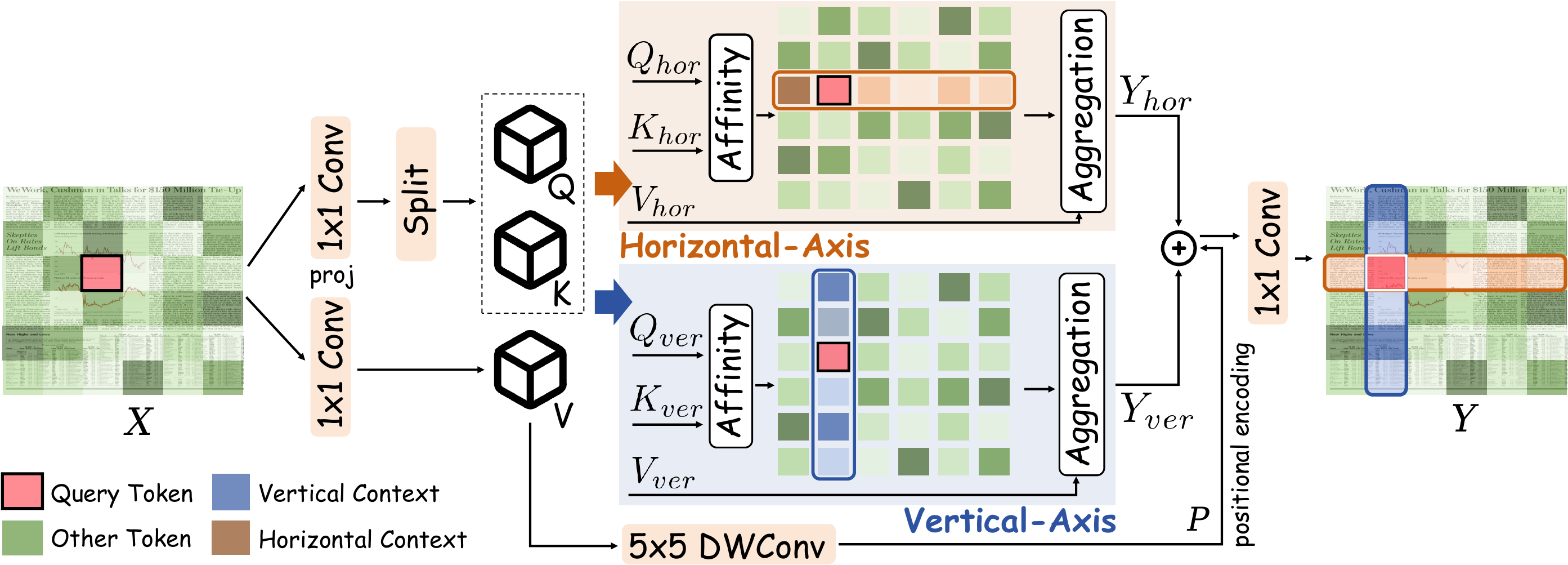}
    \caption{Illustration of Page-Axis Layout Attention~(PALA).}
    \label{fig:g2l}
\end{figure}

Specifically, given the projected query, key, and value features
$Q,K,V\in\mathbb{R}^{B\times C\times H\times W}$,
PALA reorganizes them along the two page axes and models horizontal and vertical long-range context through two parallel attention branches. For the vertical branch, features are reorganized along the vertical axis as
\(
Q_{ver},K_{ver},V_{ver}\in\mathbb{R}^{(BW)\times M\times H\times d},
\)
where $M$ is the number of attention heads and $d=C/M$. Each sequence contains the $H$ spatial positions sharing the same horizontal coordinate, allowing every token to attend to all positions along the vertical extent. The vertical attention and feature aggregation are computed as:
\begin{equation}
Y_{ver}
=
\mathrm{Softmax}
\left(
\frac{Q_{ver}K_{ver}^{T}}{\sqrt{d}}
\right)V_{ver}.
\end{equation}

Similarly, for the horizontal branch, the features are reorganized as:
\(
Q_{hor},K_{hor},V_{hor}\in\mathbb{R}^{(BH)\times M\times W\times d},
\)
where each sequence corresponds to one feature-map row, enabling direct context aggregation across the full horizontal extent. The horizontal attention and feature aggregation are computed as:
\begin{equation}
Y_{hor}
=
\mathrm{Softmax}
\left(
\frac{Q_{hor}K_{hor}^{T}}{\sqrt{d}}
\right)V_{hor}.
\end{equation}
Following the projection and positional encoding design of Area Attention, $Q$ and $K$ are obtained by splitting
$QK=\phi_{qk}(X)$ along the channel dimension, while
$V=\phi_v(X)$ and local spatial cues are encoded as
$P=\mathrm{DWConv}_{5\times5}(V)$.
Finally, the two axis-aware features are reshaped and fused as:
\begin{equation}
Y=\mathrm{Proj}(Y_{ver}+Y_{hor}+P).
\end{equation}
Unlike Area Attention, which restricts interactions within predefined regions, PALA preserves long-range context along the full horizontal and vertical axes. Meanwhile, its parallel branches and independent normalization design~(as shown in Fig.~\ref{fig:g2l}) avoid directional-order dependency and cross-axis competition, allowing complementary structural context to be effectively modeled.

We further analyze the attention complexity of PALA: omitting same linear projections, the attention-aggregation complexities of global attention, Area Attention with \(A\) equal-sized regions, and PALA are, respectively:
\begin{equation}
\begin{aligned}
\mathcal{O}_{\mathrm{Global}} &= O(BH^2W^2C),\quad \mathcal{O}_{\mathrm{AA}} = O(B\frac{H^2W^2}{A}C),\\
\mathcal{O}_{\mathrm{PALA}} &= O(BWH^2C+BHW^2C)
=O(BHW(H+W)C).
\end{aligned}
\end{equation}
For square feature maps ($H=W=S$), PALA reduces the quadratic complexity of global attention from $O(BS^4C)$ to $O(BS^3C)$ while preserving full horizontal and vertical interactions. Compared with Area Attention, PALA avoids restricting context aggregation within predefined regions and provides a more efficient axis-aware attention mechanism for document page.

\begin{table*}[t]
    \centering
    \caption{
        \textbf{Comparison with efficient real-time detectors on four document layout analysis benchmarks.}
        Accuracy is reported as mAP$_{50:95}$ (\%) and
        FLOPs are measured at an input resolution of $1600\times1600$.
        TensorRT FP16 latency is measured on an NVIDIA H200 GPU with batch size 1.
    }
    \vspace{-10pt}
    \resizebox{0.93\textwidth}{!}{
        \begin{tabular}{l|ccc|cccc}
            \toprule
            \textbf{Method}
            & \textbf{\makecell{FLOPs \\ (G)}}
            & \textbf{\makecell{\#Param. \\ (M)}}
            & \textbf{\makecell{Latency \\ (ms)}}
            & \textbf{D\textsuperscript{4}LA}
            & \textbf{M\textsuperscript{6}Doc}
            & \textbf{HJDataset}
            & \textbf{PRIMA-LAD} \\
            \midrule

            YOLOv8-N~\citep{Jocher_Ultralytics_YOLO_2023}
            & 45.9 & 2.8 & 1.1
            & 64.6 & 60.2 & 87.8 & 61.3 \\
            YOLOv10-N~\citep{DBLP:journals/corr/abs-2405-14458}
            & 38.2 & 2.1 & 1.4
            & 66.9 & 61.5 & 92.9 & 58.7 \\
            YOLOv11-N~\citep{khanam2024yolov11}
            & 33.1 & 2.3 & 1.3
            & 65.3 & 62.6 & 88.3 & 60.5 \\
            YOLOv12-N~\citep{tian2026yolov12}
            & 30.2 & 2.2 & 1.7
            & 66.5 & 64.5 & 89.6 & 60.4 \\
            YOLOv26-N~\citep{jocher2026ultralyticsyolo26}
            & 38.2 & 2.6 & 1.6
            & 65.2 & 62.6 & 91.5 & 59.5 \\
            \rowcolor{palerowgreen}
            \textbf{DocLayout-YOLO-N}
            & \textbf{30.2} & \textbf{2.2} & \textbf{1.5}
            & \textbf{67.2} & \textbf{66.2}
            & \textbf{93.6} & \textbf{67.7} \\

            \midrule

            YOLOv8-S~\citep{Jocher_Ultralytics_YOLO_2023}
            & 171.5 & 10.8 & 1.5
            & 65.7 & 65.4 & 88.1 & 63.2 \\
            YOLOv10-S~\citep{DBLP:journals/corr/abs-2405-14458}
            & 139.8 & 7.4 & 1.9
            & 69.1 & 68.1 & 95.9 & 61.2 \\
            YOLOv11-S~\citep{khanam2024yolov11}
            & 127.1 & 9.1 & 1.8
            & 67.8 & 67.5 & 91.4 & 62.6 \\
            YOLOv12-S~\citep{tian2026yolov12}
            & 114.8 & 8.8 & 2.3
            & 69.2 & 69.5 & 91.7 & 64.1 \\
            YOLOv26-S~\citep{jocher2026ultralyticsyolo26}
            & 142.7 & 10.0 & 2.3
            & 67.7 & 68.4 & 92.2 & 63.7 \\
            \rowcolor{palerowgreen}
            \textbf{DocLayout-YOLO-S}
            & \textbf{114.8} & \textbf{8.8} & \textbf{2.0}
            & \textbf{70.6} & \textbf{71.1}
            & \textbf{94.8} & \textbf{70.3} \\

            \midrule

            YOLOv8-M~\citep{Jocher_Ultralytics_YOLO_2023}
            & 489.8 & 25.7 & 2.8
            & 69.3 & 69.3 & 94.7 & 67.0 \\
            RT-DETRv3-R18~\citep{DBLP:conf/wacv/WangXLS25}
            & 322.5 & 21.1 & 2.4 & 69.6 & 68.1 & 94.1 & 65.7 \\
            RT-DETRv3-R34~\citep{DBLP:conf/wacv/WangXLS25}
            & 511.7 & 32.4 & 2.9 & 70.5 & 70.5 & 94.9 & 68.8 \\
            YOLOv10-M~\citep{DBLP:journals/corr/abs-2405-14458}
            & 391.0 & 16.1 & 3.3
            & 70.6 & 69.0 & 94.1 & 63.8 \\
            YOLOv11-M~\citep{khanam2024yolov11}
            & 427.7 & 20.1 & 3.0
            & 69.0 & 69.0 & 94.9 & 66.0 \\
            YOLOv12-M~\citep{tian2026yolov12}
            & 377.1 & 19.7 & 3.7
            & 70.2 & 70.9 & 93.4 & 66.3 \\
            YOLOv26-M~\citep{jocher2026ultralyticsyolo26}
            & 471.2 & 21.9 & 3.4
            & 68.7 & 69.6 & 94.2 & 65.3 \\
            \rowcolor{palerowgreen}
            \textbf{DocLayout-YOLO-M}
            & \textbf{377.1} & \textbf{19.7} & \textbf{3.1}
            & \textbf{70.7} & \textbf{73.3}
            & \textbf{95.3} & \textbf{71.3} \\

            \midrule

            YOLOv8-L~\citep{Jocher_Ultralytics_YOLO_2023}
            & 1035.3 & 43.7 & 4.0
            & 70.9 & 71.7 & 95.7 & 68.6 \\
            RT-DETRv3-R50~\citep{DBLP:conf/wacv/WangXLS25}
            & 766.5 & 43.7 & 4.1 & 72.1 & 72.4 & 94.3 & 70.4 \\
            YOLOv10-L~\citep{DBLP:journals/corr/abs-2405-14458}
            & 798.1 & 25.9 & 4.6
            & 71.6 & 72.4 & 95.6 & 65.7 \\
            YOLOv11-L~\citep{khanam2024yolov11}
            & 547.0 & 25.4 & 4.1
            & 71.4 & 72.1 & 94.9 & 67.9 \\
            YOLOv12-L~\citep{tian2026yolov12}
            & 519.9 & 26.5 & 5.8
            & 71.1 & 72.5 & 94.9 & 66.9 \\
            YOLOv26-L~\citep{jocher2026ultralyticsyolo26}
            & 586.2 & 26.3 & 4.5
            & 69.8 & 70.9 & 95.1 & 64.6 \\
            \rowcolor{palerowgreen}
            \textbf{DocLayout-YOLO-L}
            & \textbf{519.9} & \textbf{26.5} & \textbf{4.7}
            & \textbf{72.0} & \textbf{75.1}
            & \textbf{95.7} & \textbf{73.4} \\

            \midrule

            YOLOv8-X~\citep{Jocher_Ultralytics_YOLO_2023}
            & 1615.2 & 68.2 & 6.0
            & 71.0 & 72.2 & 95.4 & 69.6 \\
            RT-DETRv3-R101~\citep{DBLP:conf/wacv/WangXLS25}
            & 1533.8 & 84.6 & 6.4 & 72.2 & 72.7 & 93.0 & 70.3 \\
            YOLOv10-X~\citep{DBLP:journals/corr/abs-2405-14458}
            & 1072.6 & 31.8 & 6.5
            & 72.1 & 72.0 & 94.0 & 61.1 \\
            YOLOv11-X~\citep{khanam2024yolov11}
            & 1223.9 & 56.9 & 6.1
            & 71.7 & 73.9 & 95.4 & 65.5 \\
            YOLOv12-X~\citep{tian2026yolov12}
            & 1161.1 & 59.4 & 8.5
            & 70.8 & 74.0 & 94.0 & 67.0 \\
            YOLOv26-X~\citep{jocher2026ultralyticsyolo26}
            & 1309.6 & 59.0 & 6.6
            & 70.8 & 71.3 & 95.7 & 65.8 \\
            \rowcolor{palerowgreen}
            \textbf{DocLayout-YOLO-X}
            & \textbf{1161.1} & \textbf{59.4} & \textbf{6.9}
            & \textbf{72.4} & \textbf{75.5}
            & \textbf{96.0} & \textbf{73.1} \\

            \bottomrule
        \end{tabular}
    }
    \label{tab:yolo_efficiency}
\end{table*}

\section{Experiments}

\subsection{Experimental Setup}

\paragraph{Benchmarks and comparison methods} We evaluate DocLayout-YOLO on four heterogeneous DLA benchmarks: D\textsuperscript{4}LA~\citep{DBLP:conf/iccv/DaLZY23}, M\textsuperscript{6}Doc~\citep{DBLP:conf/cvpr/ChengZWZZXLDJ23}, HJDataset~\citep{shen2020large}, and PRIMA-LAD~\citep{DBLP:conf/icdar/AntonacopoulosBPP09}. These datasets cover diverse document scenarios, spanning modern and historical sources, multiple languages, and diverse page structures. DocLayout-YOLO is compared with two types of methods: (1) real-time detectors, including YOLO-based models~\citep{Jocher_Ultralytics_YOLO_2023,DBLP:journals/corr/abs-2405-14458,khanam2024yolov11,tian2026yolov12,jocher2026ultralyticsyolo26} and RT-DETRv3~\citep{DBLP:conf/wacv/WangXLS25}; and (2) specialized DLA approaches, including representative unimodal and multimodal methods. Details of the comparison methods and benchmarks are provided in the Appendix~\ref{sec:bench_method_detail}.


\paragraph{Implementation Details}

DocLayout-YOLO is built on YOLOv12~\citep{tian2026yolov12}, with FlashAttention~\citep{dao2022flashattention,dao2024flashattention} integrated into PALA. Each model is pretrained on MeshLayout-1M and subsequently finetuned on the downstream benchmarks using an input resolution of \(1600\times1600\) on 8$\times$~H200 GPUs. All models are trained
for 500 epochs using the SGD optimizer with an initial learning rate of 0.02, consistent with YOLOv12. A linear learning rate decay schedule is adopted, with a linear warm-up for the first three epochs. The latencies of all models are tested on an H200 GPU using TensorRT FP16. See Appendix~\ref{sec:appendix_imp} for full implementation details.

\subsection{Comparisons with State-of-the-Art Real-time Detectors}
\label{sec:sota_comparison}

We compare DocLayout-YOLO with existing real-time detectors in Tab.~\ref{tab:yolo_efficiency}. 
\textbf{N-scale:} DocLayout-YOLO-N improves over the baseline YOLOv12-N by 0.7\%/1.7\%/4.0\%/7.3\% mAP on four benchmarks, while reducing latency by 11.8\%. It further exceeds YOLOv8-N, YOLOv10-N, and YOLOv26-N by up to 9.0\% at comparable inference speed. 
\textbf{S-scale:} DocLayout-YOLO-S delivers gains of 1.4\%/1.6\%/3.1\%/6.2\% over YOLOv12-S with 13.0\% lower latency. Compared with YOLOv26-S and RT-DETRv3-R18, it achieves improvements of up to 6.6\% and 4.6\%, respectively. 
\textbf{M-scale:} With a latency of 3.1\,ms, DocLayout-YOLO-M surpasses YOLOv12-M by 0.5\%/2.4\%/1.9\%/5.0\% and provides a 16.2\% speedup. It also outperforms RT-DETRv3-R34 by up to 2.8\% while using 26.3\% fewer FLOPs. 
\textbf{L-scale:} DocLayout-YOLO-L improves upon YOLOv12-L by 0.9\%/2.6\%/0.8\%/6.5\% while reducing latency from 5.8\,ms to 4.7\,ms. 
\textbf{X-scale:} DocLayout-YOLO-X achieves the best overall results of 72.4\%/75.5\%/96.0\%/73.1\% mAP on four benchmarks. Compared with YOLOv12-X, it yields gains of 1.6\%/1.5\%/2.0\%/6.1\% together with an 18.8\% latency reduction, and surpasses RT-DETRv3-R101 by up to 3.0\% with comparable inference latency. In addition, PALA retains the same feature projections and positional encoding as Area Attention in YOLOv12, while standard FLOPs profiling does not account for attention aggregation, resulting in identical parameter and GFLOP counts; PALA nevertheless achieves consistently lower latency through more efficient axis-wise aggregation.

\subsection{Comparisons with State-of-the-Art Specialized DLA Methods}

We compare DocLayout-YOLO with specialized document layout analysis methods in Tab.~\ref{tab:comp1}. DocLayout-YOLO-X establishes state-of-the-art performance on all four benchmarks, achieving 72.4\%/75.5\%/96.0\%/73.1\% mAP on D\textsuperscript{4}LA, M\textsuperscript{6}Doc, HJDataset, and PRIMA-LAD, respectively. Notably, DocLayout-YOLO relies solely on visual inputs and consistently surpasses both visual-only unimodal and visual-language multimodal approaches. Compared with the previous best-performing unimodal methods, DocLayout-YOLO improves performance on D\textsuperscript{4}LA and M\textsuperscript{6}Doc by 1.5\% and 2.6\% while using only 59.4M parameters, substantially fewer than most competitive DLA models. Furthermore, unlike multimodal methods such as LayoutLMv3, VGT, and DoPTA-HR, DocLayout-YOLO does not require OCR annotations or semantic language features during inference, demonstrating that strong document layout perception can be achieved through efficient visual-only modeling. Moreover, many approaches rely on large backbones, DocLayout-YOLO maintains a lightweight YOLO-based architecture and achieves superior performance with significantly lower model complexity.

\begin{table*}[t]
\centering
\caption{Performance comparison with existing state-of-the-art~(SOTA) methods on public benchmarks. \textbf{Best} and \underline{Second Best} results are highlighted. $\ssymbol{2}$ indicates that ground truth text-level bounding boxes and OCR annotations are used during prediction.}
\vspace{-10pt}
\resizebox{1.0\textwidth}{!}{
    \begin{tabular}{l|l|l|cc|c|cccc}
    \toprule
    \multirow{2}{*}{\textbf{Method}}
    & \multirow{2}{*}{\textbf{Venue}}
    & \multirow{2}{*}{\textbf{Backbone}}
    & \multicolumn{2}{c|}{\textbf{Modality}}
    & \multirow{2}{*}{\textbf{Params}}
    & \multirow{2}{*}{\textbf{D\textsuperscript{4}LA}}
    & \multirow{2}{*}{\textbf{M\textsuperscript{6}Doc}}
    & \multirow{2}{*}{\textbf{HJDataset}}
    & \multirow{2}{*}{\textbf{PRIMA-LAD}} \\
    & & & \textbf{V} & \textbf{L} & & & & & \\
    \midrule

    LayoutParser~\citep{shen2021layoutparser}
    & ICDAR 2021
    & ResNext~\citep{DBLP:conf/cvpr/XieGDTH17}
    & \Checkmark & \XSolidBrush & 6.6M
    & - & - & 81.6 & \underline{64.7} \\

    DocSegTr~\citep{biswas2022docsegtr}
    & arXiv 2022
    & ResNet~\citep{DBLP:conf/cvpr/HeZRS16}
    & \Checkmark & \XSolidBrush & 84.2M
    & - & 60.3 & 83.1 & 42.5 \\

    DiT-L~\citep{DBLP:conf/mm/LiXL0ZW22}
    & ACM MM 2022
    & ViT-L~\citep{DBLP:conf/iclr/DosovitskiyB0WZ21}
    & \Checkmark & \XSolidBrush & 361.6M
    & 68.2 & 70.2 & - & - \\

    SelfDocSeg~\citep{DBLP:conf/icdar/MaityBMBLBP23}
    & ICDAR 2023
    & ResNet-50~\citep{DBLP:conf/cvpr/HeZRS16}
    & \Checkmark & \XSolidBrush & 63.7M
    & - & - & 78.8 & 52.1 \\

    SwinDocSegmenter~\citep{banerjee2023swindocsegmenter}
    & ICDAR 2023
    & Swin~\citep{DBLP:conf/iccv/LiuLCHWYZLG21}
    & \Checkmark & \XSolidBrush & 220.9M
    & - & 47.1 & 84.6 & 54.4 \\

    GraphKD~\citep{DBLP:conf/icdar/BanerjeeBLP24}
    & ICDAR 2024
    & ResNet~\citep{DBLP:conf/cvpr/HeZRS16}
    & \Checkmark & \XSolidBrush & 136.6M
    & - & - & 79.7 & 41.9 \\

    SemiDocSeg~\citep{banerjee2024semidocseg}
    & IJDAR 2024
    & Swin-L~\citep{DBLP:conf/iccv/LiuLCHWYZLG21}
    & \Checkmark & \XSolidBrush & 220.9M
    & - & - & 81.1 & 45.9 \\

    RoDLA~\citep{chen2024rodla}
    & CVPR 2024
    & InternImage~\citep{wang2023internimage}
    & \Checkmark & \XSolidBrush & 338.9M
    & - & 70.0 & - & - \\

    PARL~\citep{liu2026parl}
    & arXiv 2026
    & HRNetv2~\citep{sun2019high}
    & \Checkmark & \XSolidBrush & 65M
    & \underline{70.9} & \underline{72.9} & - & - \\

    HybriDLA~\citep{chen2026hybridla}
    & AAAI 2026
    & InternImage~\citep{wang2023internimage}
    & \Checkmark & \XSolidBrush & 392M
    & - & 71.4 & - & - \\

    \midrule

    LayoutLMv3~\citep{DBLP:conf/mm/HuangL0LW22}
    & ACM MM 2022
    & RoBERTa-B~\citep{DBLP:journals/corr/abs-1907-11692}
    & \Checkmark & \Checkmark & 138.4M
    & 60.0 & 64.3 & 82.7 & 40.3 \\

    VGT~\citep{DBLP:conf/iccv/DaLZY23}
    & ICCV 2023
    & ViT-B~\citep{DBLP:conf/iclr/DosovitskiyB0WZ21}
    & \Checkmark & \Checkmark & 266.8M
    & 68.8 & - & - & - \\

    M2Doc$\ssymbol{2}$~\citep{zhang2024m2doc}
    & AAAI 2024
    & ResNet~\citep{DBLP:conf/cvpr/HeZRS16}
    & \Checkmark & \Checkmark & 266.5M
    & - & 69.9 & - & - \\

    DoPTA-HR~\citep{DBLP:journals/corr/abs-2412-12902}
    & arXiv 2024
    & ViT-B~\citep{DBLP:conf/iclr/DosovitskiyB0WZ21}
    & \Checkmark & \Checkmark & 85M
    & 70.7 & 69.5 & - & - \\

    DocSAM~\citep{li2025docsam}
    & CVPR 2025
    & Swin-L~\citep{DBLP:conf/iccv/LiuLCHWYZLG21}
    & \Checkmark & \Checkmark & 317M
    & 55.5 & 50.0 & \underline{90.5} & - \\

    PromptDLA~\citep{zhang2026promptdla}
    & TMM 2026
    & ViT-B~\citep{DBLP:conf/iclr/DosovitskiyB0WZ21}
    & \Checkmark & \Checkmark & -
    & 69.2 & 69.1 & - & - \\

    \midrule
    \rowcolor{palerowgreen}
    \textbf{DocLayout-YOLO-X}
    & \textbf{-}
    & \textbf{v12x w/ PALA}
    & \textbf{\Checkmark}
    & \textbf{\XSolidBrush}
    & \textbf{59.4M}
    & \textbf{72.4}
    & \textbf{75.5}
    & \textbf{96.0}
    & \textbf{73.1} \\

    \bottomrule
    \end{tabular}
}
\label{tab:comp1}
\end{table*}

\begin{table}[t]
\centering
\vspace{-10pt}
\caption{Ablation study of MeshLayout-1M pretraining and PALA across different model sizes.}
\label{tab:ablation_improvements}
\vspace{-10pt}
\resizebox{0.892\linewidth}{!}{
\begin{tabular}{c|cc|cccc}
\toprule
\multirow{2}{*}{\textbf{Size}}
& \multicolumn{2}{c|}{\textbf{Components}}
& \multicolumn{4}{c}{\textbf{Benchmarks}} \\
& \textbf{MeshLayout-1M}
& \textbf{PALA}
& \textbf{D\textsuperscript{4}LA}
& \textbf{M\textsuperscript{6}Doc}
& \textbf{HJDataset}
& \textbf{PRIMA-LAD} \\
\midrule

\multirow{4}{*}{\textbf{N}}
& & & 66.5 & 64.5 & 89.6 & 60.4 \\
& & \cmark
& 67.0 & 65.3 & 91.0 & 62.6 \\
& \cmark &
& 66.9 & 65.4 & 92.1 & 65.8 \\
& \cmark & \cmark
& \textbf{67.2}
& \textbf{66.2}
& \textbf{93.6}
& \textbf{67.7} \\

\midrule

\multirow{4}{*}{\textbf{S}}
& & & 69.2 & 69.5 & 91.7 & 64.1 \\
& & \cmark
& 70.2 & 70.1 & 92.9 & 66.8 \\
& \cmark &
& 69.8 & 70.4 & 93.9 & 68.6 \\
& \cmark & \cmark
& \textbf{70.6}
& \textbf{71.1}
& \textbf{94.8}
& \textbf{70.3} \\

\midrule

\multirow{4}{*}{\textbf{M}}
& & & 70.2 & 70.9 & 93.4 & 66.3 \\
& & \cmark
& 70.6 & 71.5 & 95.2 & 69.2 \\
& \cmark &
& 70.6 & 72.6 & 95.0 & 70.0 \\
& \cmark & \cmark
& \textbf{70.7}
& \textbf{73.3}
& \textbf{95.3}
& \textbf{71.3} \\

\midrule

\multirow{4}{*}{\textbf{L}}
& & & 71.1 & 72.5 & 94.9 & 66.9 \\
& & \cmark
& 71.6 & 73.7 & 95.3 & 69.1 \\
& \cmark &
& 71.7 & 74.5 & 95.4 & 71.9 \\
& \cmark & \cmark
& \textbf{72.0}
& \textbf{75.1}
& \textbf{95.7}
& \textbf{73.4} \\

\midrule

\multirow{4}{*}{\textbf{X}}
& & & 70.8 & 74.0 & 94.0 & 67.0 \\
& & \cmark
& 71.8 & 74.5 & 95.2 & 67.8 \\
& \cmark &
& 71.5 & 75.0 & 95.8 & 71.8 \\
& \cmark & \cmark
& \textbf{72.4}
& \textbf{75.5}
& \textbf{96.0}
& \textbf{73.1} \\

\bottomrule
\end{tabular}
}
\end{table}

\subsection{Ablation Studies}
\label{sec:ablation}



\paragraph{Component Effectiveness Analysis.}
Tab.~\ref{tab:ablation_improvements} further disentangles the contributions of MeshLayout-1M pretraining and PALA. Specifically, the MeshLayout-only setting corresponds to the original YOLOv12 pretrained on MeshLayout-1M, while the PALA-only replaces Area Attention with PALA without pretraining. Averaged across five model sizes, PALA alone improves mAP by 0.7\%/0.7\%/1.2\%/2.2\% on D\textsuperscript{4}LA, M\textsuperscript{6}Doc, HJDataset, and PRIMA-LAD, respectively, while MeshLayout-1M pretraining alone yields gains of 0.5\%/1.3\%/1.7\%/4.7\%. Combining both components consistently achieves the best results, with average improvements of 1.0\%/2.0\%/2.4\%/6.2\% over the YOLOv12 baseline. These results demonstrate that the gains of DocLayout-YOLO arise from both the architectural design of PALA and document-domain pretraining with MeshLayout-1M, with the two components providing complementary benefits.



\begin{figure}[t]
    \centering
    \includegraphics[width=0.80\linewidth]{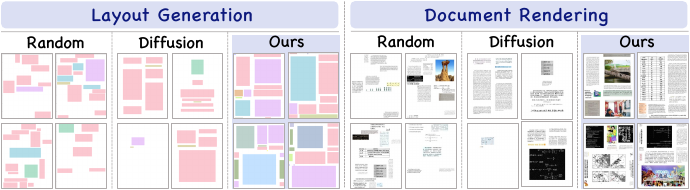}
    \vspace{-8pt}
    \caption{Qualitative comparison of random placement, diffusion, and Mesh-candidate BestFit.}
    \label{fig:comp6}
\end{figure}

\paragraph{Comparison of Pretraining Datasets.}
Tab.~\ref{tab:synthetic_pretrain_comparison_m6doc} compares MeshLayout-1M with representative real-world and synthetic document pretraining datasets. Real-world datasets include PubLayNet~\citep{DBLP:conf/icdar/ZhongTJ19} and DocBank~\citep{DBLP:conf/coling/LiXCHWLZ20}, which provide annotated real-world scientific pages. Synthetic alternatives include DocSynth~\citep{DBLP:conf/icdar/BiswasR0021} which synthesizes pages using layouts derived from PubLayNet, while RanLayNet~\citep{anand2023ranlaynet} adopts random placement. For diffusion-based methods, we use LayoutDM~\citep{DBLP:conf/cvpr/InoueKSOY23} and LACE~\citep{chen2024towards} trained on DocBank to generate layouts and then render with DocBank elements. Fig.~\ref{fig:comp6} qualitatively compares synthetic layouts and pages generated by random placement, diffusion-based generation, and Mesh-candidate BestFit. For a fair comparison, we pretrain DocLayout-YOLO-S on different datasets under identical settings and evaluate downstream performance. Existing datasets show inconsistent transfer across document types, whereas MeshLayout yields increasingly consistent gains with larger pretraining scales. With 1M pages, it improves all seven document types by 1.1\%$\sim$3.1\% mAP, demonstrating scalable and robust transfer to heterogeneous layouts.

\definecolor{poslight}{RGB}{225,245,225}
\definecolor{posmid}{RGB}{190,235,190}
\definecolor{posstrong}{RGB}{150,220,150}

\definecolor{neglight}{RGB}{255,248,210}
\definecolor{negmid}{RGB}{255,238,160}
\definecolor{negstrong}{RGB}{255,222,100}

\newcommand{\poslightbox}[1]{\colorbox{poslight}{\scriptsize #1}}
\newcommand{\posmidbox}[1]{\colorbox{posmid}{\scriptsize #1}}
\newcommand{\posstrongbox}[1]{\colorbox{posstrong}{\scriptsize #1}}

\newcommand{\neglightbox}[1]{\colorbox{neglight}{\scriptsize #1}}
\newcommand{\negmidbox}[1]{\colorbox{negmid}{\scriptsize #1}}
\newcommand{\negstrongbox}[1]{\colorbox{negstrong}{\scriptsize #1}}

\begin{table}[t]
\centering
\caption{
Comparison of different document pretraining datasets across
M\textsuperscript{6}Doc document types.
}
\vspace{-10pt}
\label{tab:synthetic_pretrain_comparison_m6doc}

\resizebox{0.99\textwidth}{!}{
\begin{tabular}{lc|ccccccc}
\toprule
\multirow{2}{*}{\textbf{Pretraining Data}}
& \multirow{2}{*}{\textbf{Volume}}
& \multicolumn{7}{c}{\textbf{M\textsuperscript{6}Doc}} \\
\cmidrule(lr){3-9}
&
& \textbf{Book}
& \textbf{Magazine}
& \textbf{Newspaper}
& \textbf{Note}
& \textbf{Paper}
& \textbf{Test Paper}
& \textbf{Textbook} \\
\midrule

None
& --
& 53.5
& 74.2
& 74.4
& 64.9
& 71.5
& 71.5
& 65.7 \\

\midrule

DocBank~\citep{DBLP:conf/coling/LiXCHWLZ20}
& 400K
& 53.9 \poslightbox{(+0.4)}
& 74.7 \poslightbox{(+0.5)}
& 74.8 \poslightbox{(+0.4)}
& 65.5 \posmidbox{(+0.6)}
& 72.7 \posstrongbox{(+1.2)}
& 71.1 \neglightbox{(-0.4)}
& 66.1 \poslightbox{(+0.4)} \\

PubLayNet~\citep{DBLP:conf/icdar/ZhongTJ19}
& 340K
& 53.7 \poslightbox{(+0.2)}
& 74.0 \neglightbox{(-0.2)}
& 74.5 \poslightbox{(+0.1)}
& 64.0 \negmidbox{(-0.9)}
& 70.5 \negmidbox{(-1.0)}
& 71.6 \poslightbox{(+0.1)}
& 65.9 \poslightbox{(+0.2)} \\

\midrule

DocSynth~\citep{DBLP:conf/icdar/BiswasR0021}
& 300K
& 53.2 \neglightbox{(-0.3)}
& 73.7 \neglightbox{(-0.5)}
& 73.5 \negmidbox{(-0.9)}
& 62.7 \negstrongbox{(-2.2)}
& 69.9 \negstrongbox{(-1.6)}
& 69.4 \negstrongbox{(-2.1)}
& 65.0 \negmidbox{(-0.7)} \\

RanLayNet~\citep{anand2023ranlaynet}
& 7K
& 53.9 \poslightbox{(+0.4)}
& 73.2 \negmidbox{(-1.0)}
& 72.8 \negstrongbox{(-1.6)}
& 62.3 \negstrongbox{(-2.6)}
& 70.7 \negmidbox{(-0.8)}
& 70.3 \negstrongbox{(-1.3)}
& 65.2 \neglightbox{(-0.5)} \\

LayoutDM~\citep{DBLP:conf/cvpr/InoueKSOY23}
& 300K
& 53.3 \neglightbox{(-0.2)}
& 74.7 \poslightbox{(+0.5)}
& 74.5 \poslightbox{(+0.1)}
& 62.3 \negstrongbox{(-2.6)}
& 69.7 \negstrongbox{(-1.8)}
& 69.9 \negstrongbox{(-1.6)}
& 64.5 \negstrongbox{(-1.3)} \\

LACE~\citep{chen2024towards}
& 300K
& 54.0 \poslightbox{(+0.5)}
& 74.0 \neglightbox{(-0.2)}
& 74.3 \neglightbox{(-0.1)}
& 64.2 \negmidbox{(-0.7)}
& 71.8 \poslightbox{(+0.3)}
& 71.3 \neglightbox{(-0.2)}
& 65.7 {\scriptsize(0.0)} \\

\midrule

MeshLayout
& 300K
& 53.9 \poslightbox{(+0.4)}
& 74.9 \posmidbox{(+0.7)}
& 75.1 \posmidbox{(+0.7)}
& 65.7 \posmidbox{(+0.8)}
& 72.2 \posmidbox{(+0.7)}
& 72.0 \poslightbox{(+0.5)}
& 65.5 \neglightbox{(-0.2)} \\

MeshLayout
& 500K
& 54.7 \posstrongbox{(+1.2)}
& 75.2 \posmidbox{(+1.0)}
& 75.3 \posmidbox{(+0.9)}
& 65.6 \posmidbox{(+0.7)}
& 72.9 \posstrongbox{(+1.4)}
& 72.4 \posmidbox{(+0.9)}
& 66.4 \posmidbox{(+0.7)} \\

\textbf{MeshLayout}
& \textbf{1M}
& \textbf{55.6} \posstrongbox{\textbf{(+2.1)}}
& \textbf{76.0} \posstrongbox{\textbf{(+1.8)}}
& \textbf{75.9} \posstrongbox{\textbf{(+1.5)}}
& \textbf{66.7} \posstrongbox{\textbf{(+1.8)}}
& \textbf{74.6} \posstrongbox{\textbf{(+3.1)}}
& \textbf{72.6} \posstrongbox{\textbf{(+1.1)}}
& \textbf{66.9} \posstrongbox{\textbf{(+1.2)}} \\

\bottomrule
\end{tabular}
}
\end{table}

\begin{table}[t]
    \centering
    \caption{Ablation study on the core components of Mesh-candidate BestFit.}
    \vspace{-10pt}
    \resizebox{0.99\linewidth}{!}{
        \begin{tabular}{lccc|ccc}
            \toprule
            \textbf{Method}
            & \textbf{\makecell{Meshgrid \\ Construction}}
            & \textbf{\makecell{Scale-stratified \\ Sampling}}
            & \textbf{\makecell{BestFit \\ Matching}}
            & \textbf{M6Doc}
            & \textbf{HJDataset}
            & \textbf{PRIMA-LAD} \\
            \midrule

            Baseline (w/o MeshLayout pretraining)
            & -- & -- & --
            & 70.1
            & 92.9
            & 66.8 \\
            \midrule

            w/o Meshgrid Construction
            & \redx & \cmark & --
            & 69.9
            & 92.2
            & 67.2 \\

            w/o Scale-stratified Sampling
            & \cmark & \redx & \cmark
            & 70.8
            & 94.4
            & 69.2 \\

            w/o BestFit Matching
            & \cmark & \cmark & \redx
            & 70.4 & 94.3 & 68.5 \\

            \midrule

            \rowcolor{palerowgreen}
            \textbf{Full MeshLayout}
            & \cmark & \cmark & \cmark
            & \textbf{71.1}
            & \textbf{94.8}
            & \textbf{70.3} \\

            \bottomrule
        \end{tabular}
    }
    \label{tab:meshlayout_component_ablation}
\end{table}

\paragraph{Ablations on MeshLayout Components.}
Tab.~\ref{tab:meshlayout_component_ablation} ablates each component of Mesh-candidate BestFit using DocLayout-YOLO-S. For every variant, we generate the same number of pretraining pages (1M) and compare downstream fine-tuning performance. 
1) Removing meshgrid construction reduces synthesis to unconstrained random placement, producing disordered layouts (Fig.~\ref{fig:comp6}) and degrading performance by 1.2\%/2.6\%/3.1\% on M\textsuperscript{6}Doc, HJDataset, and PRIMA-LAD, respectively. 
2) Without scale-stratified sampling, small elements become overrepresented, weakening layout structure and reducing performance by 0.3\%/0.4\%/1.1\%. 
3) Without BestFit matching, random candidate-region assignment causes size mismatches and excessive blank areas, leading to drops of 0.7\%/0.5\%/1.8\%. 
These results demonstrate that the three components jointly ensure structural plausibility, scale consistency, and effective downstream transfer.

\begin{table}[t]
\centering
\vspace{-10pt}
\begin{minipage}[t]{0.493\linewidth}
\vspace{0pt}
\centering
\caption{Local attentions comparison.}
\label{tab:pala_axis_replacement_ablation}
\vspace{-10pt}

\resizebox{\linewidth}{!}{
    \begin{tabular}{c|l|c|cc|c}
    \toprule
    \textbf{Size}
    & \textbf{\makecell{Attention\\Type}}
    & \textbf{\makecell{Mem.\\(G)}}
    & \textbf{\makecell{CUDA\\FP32}}
    & \textbf{\makecell{CUDA\\FP16}}
    & \textbf{M\textsuperscript{6}Doc} \\
    \midrule

    \multirow{9}{*}{\textbf{N}}
    & Global~\citep{DBLP:conf/cvpr/0004GGH18}
    & 0.295 & 9.498 & 2.123 & 64.5 \\

    & Area~\citep{tian2026yolov12}
    & 0.295 & 5.250 & 1.656 & 64.5 \\

    & Criss-Cross~\citep{DBLP:conf/iccv/HuangWHHWL19}
    & 0.301 & 3.187 & 1.847 & 64.7 \\

    & Swin~\citep{DBLP:conf/iccv/LiuLCHWYZLG21}
    & 0.295 & 2.988 & 1.656 & 62.3 \\

    & CSWin~\citep{DBLP:conf/cvpr/DongBCZYYCG22}
    & 0.295 & 3.587 & 1.755 & 63.6 \\

    & Axial~\citep{DBLP:conf/eccv/WangZGAYC20}
    & 0.295 & 3.690 & 1.643 & 64.2 \\

    \cmidrule(lr){2-6}

    & Horizontal-only
    & 0.295 & 2.867 & 1.378 & 64.1 \\

    & Vertical-only
    & 0.295 & 2.901 & 1.398 & 63.9 \\

    & \cellcolor{palerowgreen}\textbf{PALA (ours)}
    & \cellcolor{palerowgreen}\textbf{0.295}
    & \cellcolor{palerowgreen}\textbf{3.596}
    & \cellcolor{palerowgreen}\textbf{1.500}
    & \cellcolor{palerowgreen}\textbf{65.3} \\

    \midrule

    \multirow{9}{*}{\textbf{L}}
    & Global~\citep{DBLP:conf/cvpr/0004GGH18}
    & 0.514 & 61.524 & 9.184 & 73.0 \\

    & Area~\citep{tian2026yolov12}
    & 0.512 & 30.956 & 5.637 & 71.1 \\

    & Criss-Cross~\citep{DBLP:conf/iccv/HuangWHHWL19}
    & 0.516 & 19.399 & 6.588 & 73.2 \\

    & Swin~\citep{DBLP:conf/iccv/LiuLCHWYZLG21}
    & 0.512 & 18.847 & 5.485 & 71.7 \\

    & CSWin~\citep{DBLP:conf/cvpr/DongBCZYYCG22}
    & 0.514 & 21.559 & 5.905 & 72.6 \\

    & Axial~\citep{DBLP:conf/eccv/WangZGAYC20}
    & 0.512 & 22.266 & 5.223 & 72.4 \\

    \cmidrule(lr){2-6}

    & Horizontal-only
    & 0.512 & 17.862 & 4.325 & 72.7 \\

    & Vertical-only
    & 0.512 & 17.912 & 4.378 & 71.4 \\

    & \cellcolor{palerowgreen}\textbf{PALA (ours)}
    & \cellcolor{palerowgreen}\textbf{0.512}
    & \cellcolor{palerowgreen}\textbf{20.709}
    & \cellcolor{palerowgreen}\textbf{4.694}
    & \cellcolor{palerowgreen}\textbf{73.7} \\

    \bottomrule
    \end{tabular}
}
\end{minipage}
\hfill
\begin{minipage}[t]{0.495\linewidth}
\vspace{0pt}
\centering
\caption{
Speed, memory, and efficiency.
}
\label{tab:efficiency_comparison}
\vspace{-10pt}
\resizebox{\linewidth}{!}{
    \begin{tabular}{l|cc|cc|c}
    \toprule
    \multirow{2}{*}{\textbf{Model}}
    & \multirow{2}{*}{\makecell{\textbf{FLOPs}\\\textbf{(G)}}}
    & \multirow{2}{*}{\makecell{\textbf{Mem.}\\\textbf{(GiB)}}}
    & \multicolumn{2}{c|}{\textbf{CUDA Latency (ms)}}
    & \multirow{2}{*}{\textbf{M\textsuperscript{6}Doc}} \\
    & & & \textbf{FP32} & \textbf{FP16} & \\
    \midrule

    YOLOv8-N~\citep{Jocher_Ultralytics_YOLO_2023}
    & 45.9 & 0.29 & 2.4 & 1.1 & 60.2 \\
    YOLOv10-N~\citep{DBLP:journals/corr/abs-2405-14458}
    & 38.2 & 0.31 & 2.8 & 1.4 & 61.5 \\
    YOLOv12-N~\citep{tian2026yolov12}
    & 30.2 & 0.30 & 5.3 & 1.7 & 64.5 \\
    \rowcolor{palerowgreen}
    \textbf{DocLayout-YOLO-N}
    & \textbf{30.2}
    & \textbf{0.30}
    & \textbf{3.6}
    & \textbf{1.5}
    & \textbf{66.2} \\

    \midrule

    YOLOv8-S~\citep{Jocher_Ultralytics_YOLO_2023}
    & 171.5 & 0.34 & 5.5 & 1.5 & 65.4 \\
    YOLOv10-S~\citep{DBLP:journals/corr/abs-2405-14458}
    & 139.8 & 0.48 & 5.6 & 1.9 & 68.1 \\
    YOLOv12-S~\citep{tian2026yolov12}
    & 114.8 & 0.35 & 9.9 & 2.3 & 69.5 \\
    \rowcolor{palerowgreen}
    \textbf{DocLayout-YOLO-S}
    & \textbf{114.8}
    & \textbf{0.35}
    & \textbf{6.7}
    & \textbf{2.0}
    & \textbf{71.1} \\

    \midrule

    YOLOv8-M~\citep{Jocher_Ultralytics_YOLO_2023}
    & 489.8 & 0.44 & 13.2 & 2.8 & 69.3 \\
    YOLOv10-M~\citep{DBLP:journals/corr/abs-2405-14458}
    & 391.0 & 0.53 & 12.0 & 3.3 & 69.0 \\
    YOLOv12-M~\citep{tian2026yolov12}
    & 377.1 & 0.49 & 19.4 & 3.7 & 70.9 \\
    \rowcolor{palerowgreen}
    \textbf{DocLayout-YOLO-M}
    & \textbf{377.1}
    & \textbf{0.49}
    & \textbf{14.0}
    & \textbf{3.1}
    & \textbf{73.3} \\

    \midrule

    YOLOv8-L~\citep{Jocher_Ultralytics_YOLO_2023}
    & 1035.3 & 0.57 & 22.3 & 4.0 & 71.7 \\
    YOLOv10-L~\citep{DBLP:journals/corr/abs-2405-14458}
    & 798.1 & 0.61 & 19.0 & 4.6 & 72.4 \\
    YOLOv12-L~\citep{tian2026yolov12}
    & 519.9 & 0.51 & 31.2 & 5.8 & 72.5 \\
    \rowcolor{palerowgreen}
    \textbf{DocLayout-YOLO-L}
    & \textbf{519.9}
    & \textbf{0.51}
    & \textbf{20.7}
    & \textbf{4.7}
    & \textbf{75.1} \\

    \midrule

    YOLOv8-X~\citep{Jocher_Ultralytics_YOLO_2023}
    & 1615.2 & 0.69 & 36.3 & 6.0 & 72.2 \\
    YOLOv10-X~\citep{DBLP:journals/corr/abs-2405-14458}
    & 1072.6 & 0.71 & 27.4 & 6.5 & 72.0 \\
    YOLOv12-X~\citep{tian2026yolov12}
    & 1161.1 & 0.70 & 53.1 & 8.5 & 74.0 \\
    \rowcolor{palerowgreen}
    \textbf{DocLayout-YOLO-X}
    & \textbf{1161.1}
    & \textbf{0.70}
    & \textbf{37.9}
    & \textbf{6.9}
    & \textbf{75.5} \\

    \bottomrule
    \end{tabular}
}
\end{minipage}
\end{table}

\paragraph{Page-Axis Layout Attention.}
Tab.~\ref{tab:pala_axis_replacement_ablation} compares different attention mechanisms on M\textsuperscript{6}Doc. Latencies are obtained on identical H200 to ensure fairness, while Memory usage is measured with the TensorRT model (bs = 1). Following YOLOv12~\citep{tian2026yolov12}, we adjust the number of channels to ensure that all models have similar parameters and FLOPs and disable Flash-Attention. 
Compared with Global Attention, PALA achieves higher accuracy with substantially lower latency, suggesting that unrestricted global interactions may introduce cross-region interference, while axis-wise constraints better match document structures. Swin performs worse because its local windows limit direct interactions across long document structures, while Criss-Cross and CSWin extend attention along horizontal and vertical directions to better capture such dependencies. PALA further combines both page axes and achieves the best 65.3\%/73.7\% mAP at N/L scales. Horizontal-only consistently outperforms Vertical-only, while combining both yields further gains, indicating the importance and complementarity of horizontal and vertical dependencies in document layouts. More explanations and analysis are available in Appendix~\ref{sec:appendix_pala}.

\paragraph{Speed, memory, and efficiency.} 
Tab.~\ref{tab:efficiency_comparison} evaluates the speed, memory usage, and accuracy-efficiency trade-off of DocLayout-YOLO across five model scales under identical H200 GPU hardware settings. Despite identical parameter counts to YOLOv12, DocLayout-YOLO consistently achieves lower latency, highlighting the practical efficiency of PALA's attention aggregation. Compared with YOLOv12, DocLayout-YOLO reduces FP32 latency by 1.7/3.2/5.4/10.5/15.2\,ms and FP16 latency by 0.2/0.3/0.6/1.1/1.6\,ms from N to X scales, while maintaining comparable peak memory usage. Meanwhile, compared with YOLOv8 and YOLOv10, DocLayout-YOLO further provides a favorable balance between inference speed, memory usage, and detection accuracy.

\section{Conclusion}

This work presents DocLayout-YOLO, a real-time detector specifically designed for document layout analysis. DocLayout-YOLO adapts the YOLO paradigm to heterogeneous document layouts by combining large-scale synthetic document pretraining with MeshLayout-1M and efficient long-range page-axis modeling with PALA. Extensive experiments show that DocLayout-YOLO achieves state-of-the-art results across multiple benchmarks and consistently delivers a superior accuracy--efficiency trade-off over existing real-time detectors.

\clearpage

\clearpage

\bibliography{main}
\bibliographystyle{iclr2027_conference}


\clearpage


\appendix
\renewcommand{\theHsection}{appendix.\Alph{section}}

\setcounter{table}{0}
\setcounter{figure}{0}

\renewcommand{\thetable}{S\arabic{table}}
\renewcommand{\thefigure}{S\arabic{figure}}

\newcommand\DoToC{%
  \startcontents
  \printcontents{}{1}{\noindent \textbf{\Large{Table of Contents}}\vskip3pt\vskip5pt}
  \vskip3pt\vskip5pt
}

\DoToC
\clearpage

\section{More Implementation Details}
\label{sec:appendix_imp}

\paragraph{Training Details.}
All DocLayout-YOLO variants are first pretrained on MeshLayout-1M and subsequently finetuned on each downstream benchmark. Both stages use an input resolution of \(1600\times1600\), a global batch size of 64 distributed across 8$\times$ H200 GPUs. We adopt SGD with a momentum of 0.9 and a weight decay of \(5\times10^{-4}\). For downstream finetuning, the initial learning rate is set to 0.02 and linearly decayed to \(2\times10^{-4}\). Pretraining is performed for 50 epochs, and downstream finetuning is performed for 500 epochs. Following the scale-dependent augmentation strategy of YOLOv12, Mixup ratios of 0/0/0/0.15/0.20 and copy-paste ratios of 0/0/0/0.50/0.60 are used for the N/S/M/L/X variants during downstream fine-tuning. Detailed hyperparameters are summarized in Tab.~\ref{tab:training_configuration}. For comparisons with real-time detectors, YOLOv8, YOLOv10, YOLOv11, baseline YOLOv12, are trained from scratch on each downstream dataset using the exact same and identical input resolution, optimizer and learning rate with minor adaptation to ensure a fair comparison. For specialized DLA methods, we report the results provided in their original publications.

\begin{table}[h]
    \centering
    \caption{
        Hyperparameters for MeshLayout-1M pretraining and downstream
        fine-tuning of the DocLayout-YOLO.
    }
    \resizebox{0.99\linewidth}{!}{
        \begin{tabular}{lcc}
            \toprule
            \textbf{Hyperparameters}
            & \textbf{MeshLayout-1M Pretraining}
            & \textbf{Downstream Fine-tuning} \\
            \midrule

            \multicolumn{3}{c}{\textit{Training Configuration}} \\
            \midrule
            Model scales
            & N/S/M/L/X
            & N/S/M/L/X \\

            Input resolution
            & $1600\times1600$
            & $1600\times1600$ \\

            Maximum epochs
            & 50
            & 500 \\

            Optimizer
            & SGD
            & SGD \\

            Momentum
            & 0.9
            & 0.9 \\

            Global batch size
            & 64
            & 64 \\

            Number of GPUs
            & 8
            & 8 \\

            Weight decay
            & $5\times10^{-4}$
            & $5\times10^{-4}$ \\

            Warm-up epochs
            & 3
            & 3 \\

            Warm-up momentum
            & 0.8
            & 0.8 \\

            Warm-up bias learning rate
            & 0.0
            & 0.0 \\

            Initial learning rate
            & 0.02
            & 0.02 \\

            Final learning rate
            & $2\times10^{-4}$
            & $2\times10^{-4}$ \\

            Learning-rate schedule
            & Linear decay
            & Linear decay \\

            \midrule
            \multicolumn{3}{c}{\textit{Loss Parameters}} \\
            \midrule

            Box loss gain
            & 7.5
            & 7.5 \\

            Classification loss gain
            & 0.5
            & 0.5 \\

            DFL loss gain
            & 1.5
            & 1.5 \\

            \midrule
            \multicolumn{3}{c}{\textit{Augmentation Parameters}} \\
            \midrule

            HSV hue augmentation
            & 0.015
            & 0.015 \\

            HSV saturation augmentation
            & 0.7
            & 0.7 \\

            HSV value augmentation
            & 0.4
            & 0.4 \\

            Translation augmentation
            & 0.1
            & 0.1 \\

            Scale augmentation
            & 0.5
            & 0.5 \\

            Mosaic augmentation
            & 1.0
            & 1.0 \\

            Mixup augmentation
            & 0
            & 0/0/0/0.15/0.20 \\

            Copy-paste augmentation
            & 0
            & 0/0/0/0.50/0.60 \\

            Close-mosaic epochs
            & 10
            & 10 \\

            \bottomrule
        \end{tabular}
    }
    \label{tab:training_configuration}
\end{table}

\clearpage

\paragraph{Architecture Details.}
DocLayout-YOLO preserves the macro-architecture and scaling configurations of the baseline YOLOv12, such as the R-ELAN feature aggregation. The number and placement of attention blocks also remain unchanged. Within each block, we retain the original query, key, and value projections, output projection, positional encoding branch, MLP, normalization layers, and residual connections. The only architectural modification is the replacement of area attention with PALA.

\paragraph{YOLOv26 Implementation.}
We carefully adjust YOLOv26 for optimal performance and a fair comparison. Specifically, each model is trained for 500 epochs at an input resolution of $1600\times1600$ using 8$\times$ NVIDIA H200 GPUs and a global batch size of 64. Following YOLOv26, we use MuSGD with an initial learning rate of 0.02, momentum of 0.937, and weight decay of $5\times10^{-4}$. The learning rate is linearly decayed to $2\times10^{-4}$, with a three-epoch warm-up. AMP is enabled during training. We use the same data-augmentation settings as YOLOv12.

\paragraph{RT-DETRv3 Implementation.}
We also carefully adjust RT-DETR settings for optimal performance and a fair comparison. Following deault settings, all models are trained on 8$\times$ NVIDIA H200 GPUs with eight images per GPU, giving a global batch size of 64, and initialized with their corresponding ImageNet-pretrained weights. The evaluation resolution is fixed at $1600\times1600$, while training uses multi-scale inputs from 1216 to 1984 pixels, obtained by proportionally scaling the official RT-DETRv3 augmentation range. The training pipeline includes random photometric distortion, expansion, cropping, horizontal flipping, and multi-scale resizing. We adopt AdamW with a base learning rate of $4\times10^{-4}$, a weight decay of $1\times10^{-4}$, gradient clipping with a maximum norm of 0.1, and EMA with a decay of 0.9999. The backbone learning-rate multiplier is 0.1 for R18/R34/R50 and 0.01 for R101. AMP is enabled, and the number of object queries is fixed to 300. To account for the substantially different dataset sizes, D\textsuperscript{4}LA, M\textsuperscript{6}Doc, HJDataset, and PRIMA-LAD are trained for 96/120/240/500 epochs with 700/450/120/50 warm-up iterations, respectively.

\paragraph{TensorRT Details.} For TensorRT details, GPU latency is measured on a single NVIDIA H200 GPU using TensorRT~10.16.1.11 with batch size one and a static input shape of $1\times3\times1600\times1600$. Models are exported as simplified FP32 ONNX graphs with opset~17 and compiled into separate FP32 and FP16 engines. After 30 warm-up iterations, synchronized engine execution is measured for at least 5,000 iterations and 60 seconds, excluding data transfer, preprocessing, decoding, and NMS. We report mean latency, whose 95\% confidence-interval half-width is below 0.10\,ms for all reported measurements.

\clearpage
\clearpage

\section{Benchmark \& Comparison Methods Details}
\label{sec:bench_method_detail}

\subsection{Public Benchmark Details}

To evaluate the generalization ability of DocLayout-YOLO across diverse document scenarios, we conduct experiments on four main DLA benchmarks, including M\textsuperscript{6}Doc~\citep{DBLP:conf/cvpr/ChengZWZZXLDJ23}, D\textsuperscript{4}LA~\citep{DBLP:conf/iccv/DaLZY23}, HJDataset~\citep{shen2020large}, and PRIMA-LAD~\footnote{For PRIMA-LAD, we use the 11-class annotation configuration adopted in our experiments, containing 8,942 valid regions over 478 pages.}~\citep{DBLP:conf/icdar/AntonacopoulosBPP09}. These benchmarks cover a wide range of document types, including modern and historical documents, multilingual content, and diverse layout structures.

In addition, we provide supplementary evaluations on other three widely used document layout datasets, including PubLayNet~\citep{DBLP:conf/icdar/ZhongTJ19}, DocLayNet~\citep{DBLP:conf/kdd/PfitzmannADNS22}, and IndicDLP~\citep{nath2025indicdlp}, to further examine the robustness of our method across different document domains. Detailed statistics of all benchmarks, including data volume, document types, categories, and example images, are summarized in Tab.~\ref{tab:benchmark_detail}.

\begin{table*}[h]
    \centering
    \caption{Details of public benchmarks used for evaluation.}
    \resizebox{1.0\textwidth}{!}{
        \begin{tabular}{cccccc}
        \toprule
        \textbf{Benchmark}
        & \textbf{\# Training}
        & \textbf{\# Testing}
        & \textbf{Document Types}
        & \textbf{Categories}
        & \textbf{Example Documents} \\
        \cmidrule(r){1-1}
        \cmidrule(lr){2-2}
        \cmidrule(lr){3-3}
        \cmidrule(lr){4-4}
        \cmidrule(lr){5-5}
        \cmidrule(l){6-6}

        M\textsuperscript{6}Doc
        & 5,448
        & 2,724
        & \makecell{
            Book, Magazine, Newspaper,\\
            Note, Paper, Test Paper, Textbook
        }
        & 74 fine-grained categories
        & \begin{minipage}{.3\textwidth}
            \centering
            \includegraphics[
                width=\linewidth,
                height=25mm
            ]{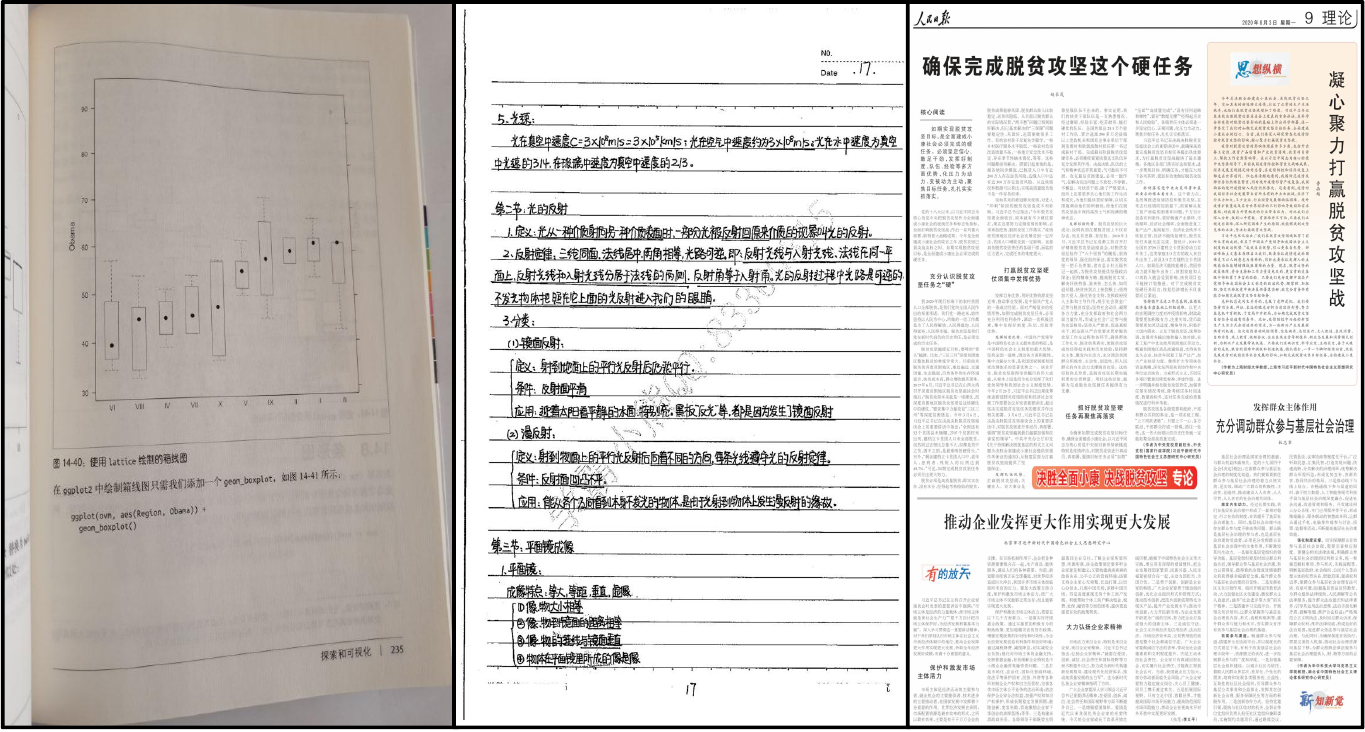}
        \end{minipage} \\

        \cmidrule(r){1-1}
        \cmidrule(lr){2-2}
        \cmidrule(lr){3-3}
        \cmidrule(lr){4-4}
        \cmidrule(lr){5-5}
        \cmidrule(l){6-6}

        D\textsuperscript{4}LA
        & 8,868
        & 2,224
        & \makecell{
            Budget, Email, Form,\\
            Invoice, Letter, Memo,\\
            News, Presentation, Resume
        }
        & \makecell{
            DocTitle, ListText, LetterHead, Question,\\
            RegionList, TableName, FigureName, Footer,\\
            Number, ParaTitle, RegionTitle, LetterDear,\\
            OtherText, Abstract, Table, Equation,\\
            PageHeader, Catalog, ParaText, Date,\\
            LetterSign, RegionKV, Author, Figure,\\
            Reference, PageFooter, PageNumber
        }
        & \begin{minipage}{.3\textwidth}
            \centering
            \includegraphics[
                width=\linewidth,
                height=25mm
            ]{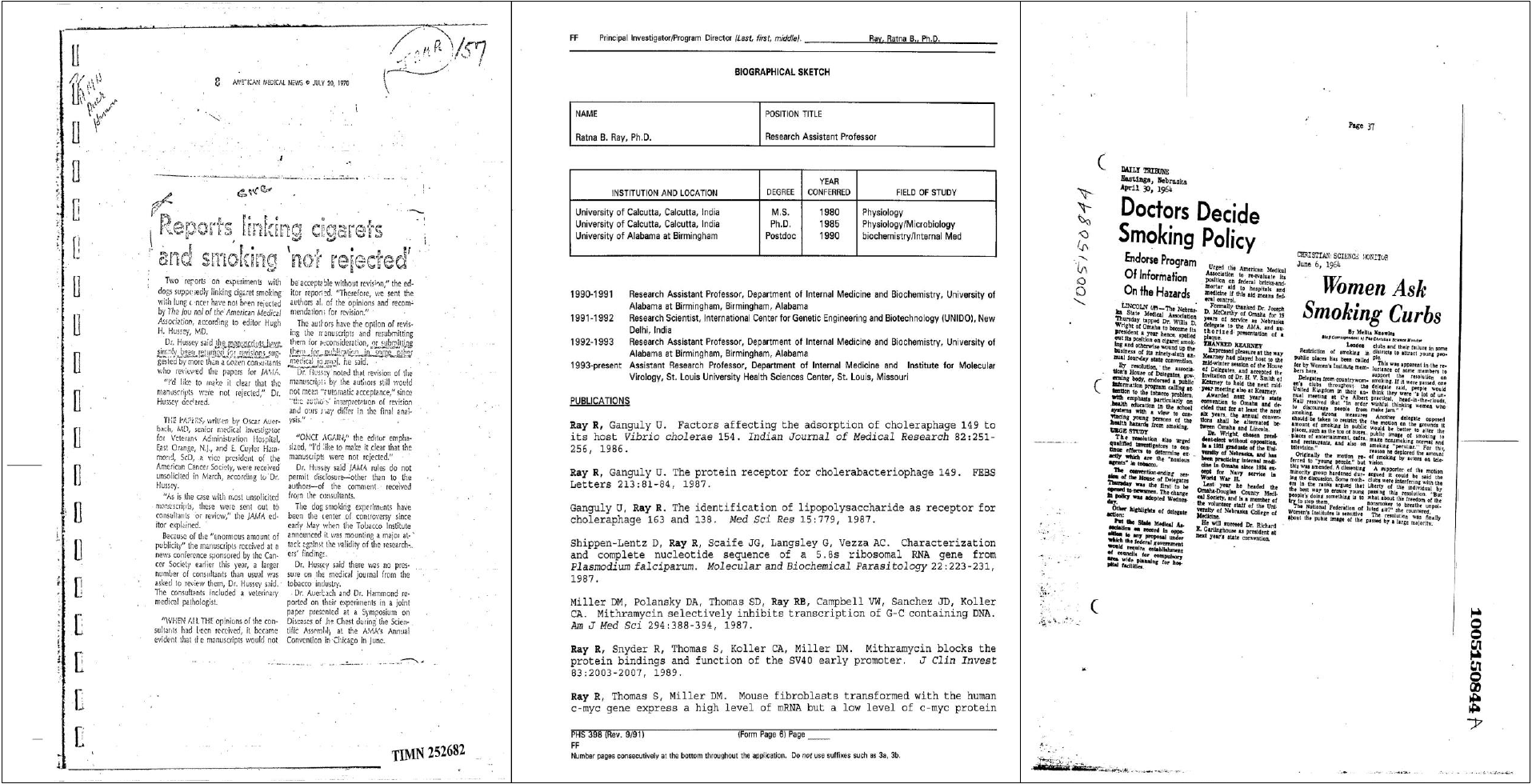}
        \end{minipage} \\

        \cmidrule(r){1-1}
        \cmidrule(lr){2-2}
        \cmidrule(lr){3-3}
        \cmidrule(lr){4-4}
        \cmidrule(lr){5-5}
        \cmidrule(l){6-6}

        HJDataset
        & 1,589
        & 341
        & \makecell{
            Main, Advertisement,\\
            Index, Other
        }
        & \makecell{
            Body, Row, Title, Bio,\\
            Name, Position, Other
        }
        & \begin{minipage}{.3\textwidth}
            \centering
            \includegraphics[
                width=\linewidth,
                height=25mm
            ]{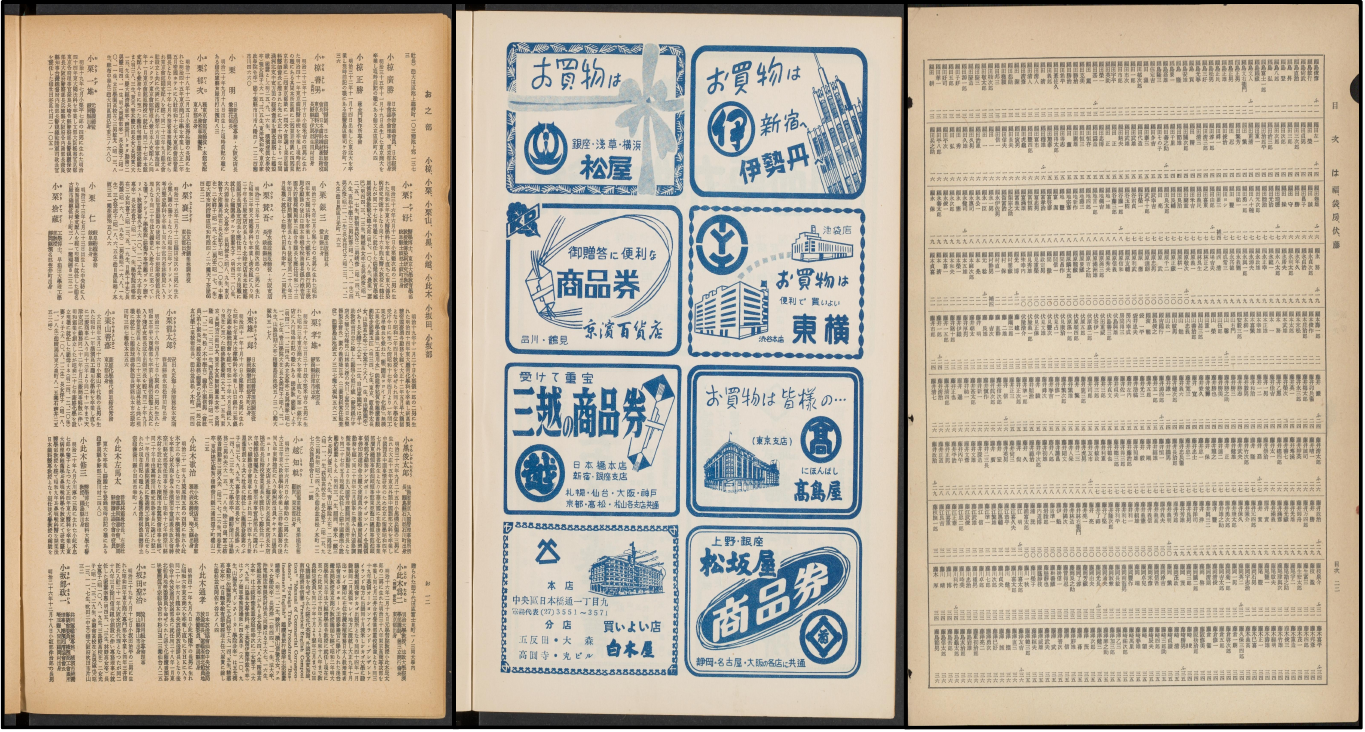}
        \end{minipage} \\

        \cmidrule(r){1-1}
        \cmidrule(lr){2-2}
        \cmidrule(lr){3-3}
        \cmidrule(lr){4-4}
        \cmidrule(lr){5-5}
        \cmidrule(l){6-6}

        PRIMA-LAD
        & 382
        & 96
        & \makecell{
            Magazine,\\
            Technical Article
        }
        & \makecell{
            Figure, Table, Paragraph, Header,\\
            Page-number, Floating, Caption, Credit,\\
            Drop-capital, Heading, Footer
        }
        & \begin{minipage}{.3\textwidth}
            \centering
            \includegraphics[
                width=\linewidth,
                height=25mm
            ]{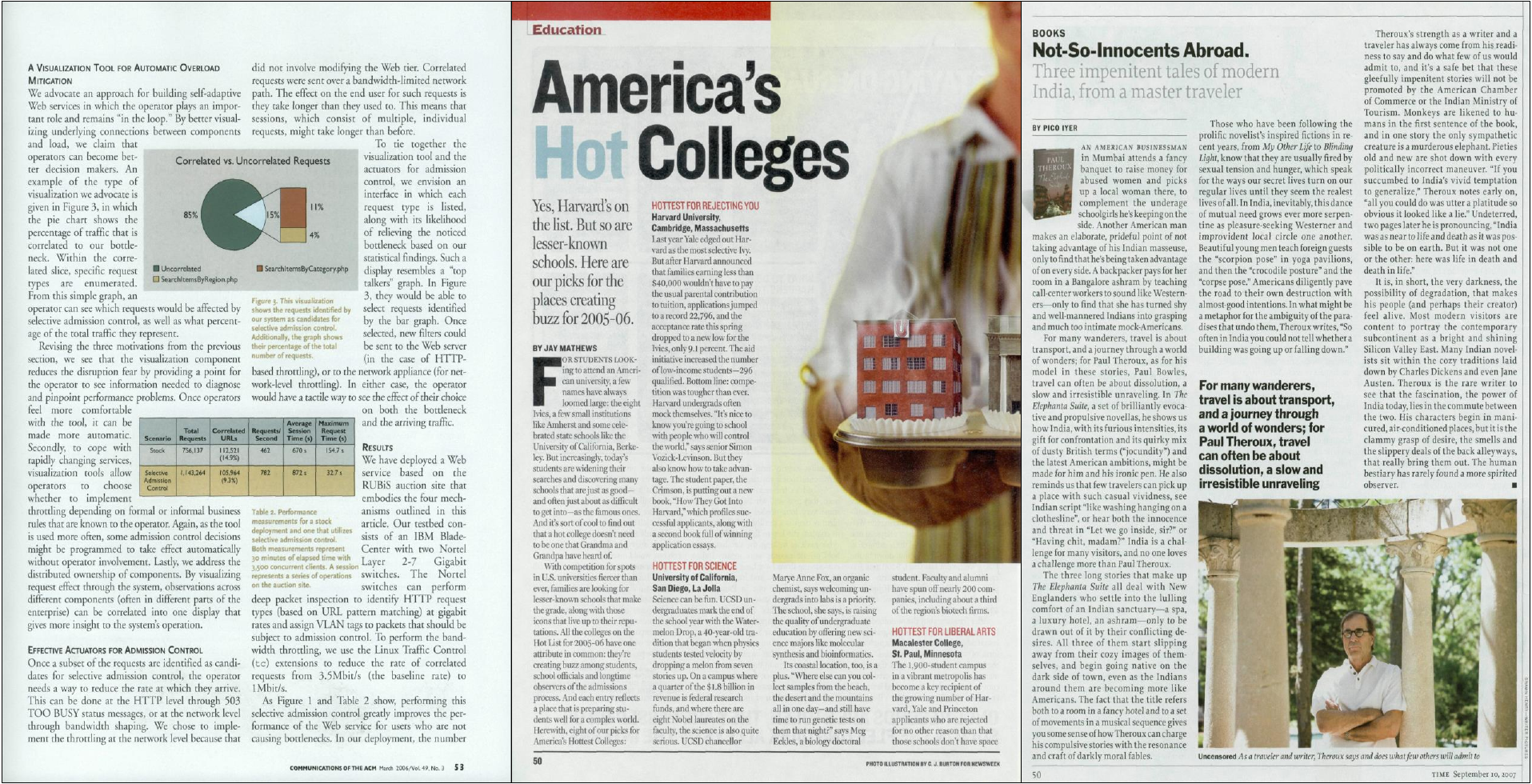}
        \end{minipage} \\

        \cmidrule(r){1-1}
        \cmidrule(lr){2-2}
        \cmidrule(lr){3-3}
        \cmidrule(lr){4-4}
        \cmidrule(lr){5-5}
        \cmidrule(l){6-6}

        PubLayNet
        & 335,703
        & 11,245
        & Academic
        & \makecell{
            Text, Title, List,\\
            Table, Figure
        }
        & \begin{minipage}{.3\textwidth}
            \centering
            \includegraphics[
                width=\linewidth,
                height=25mm
            ]{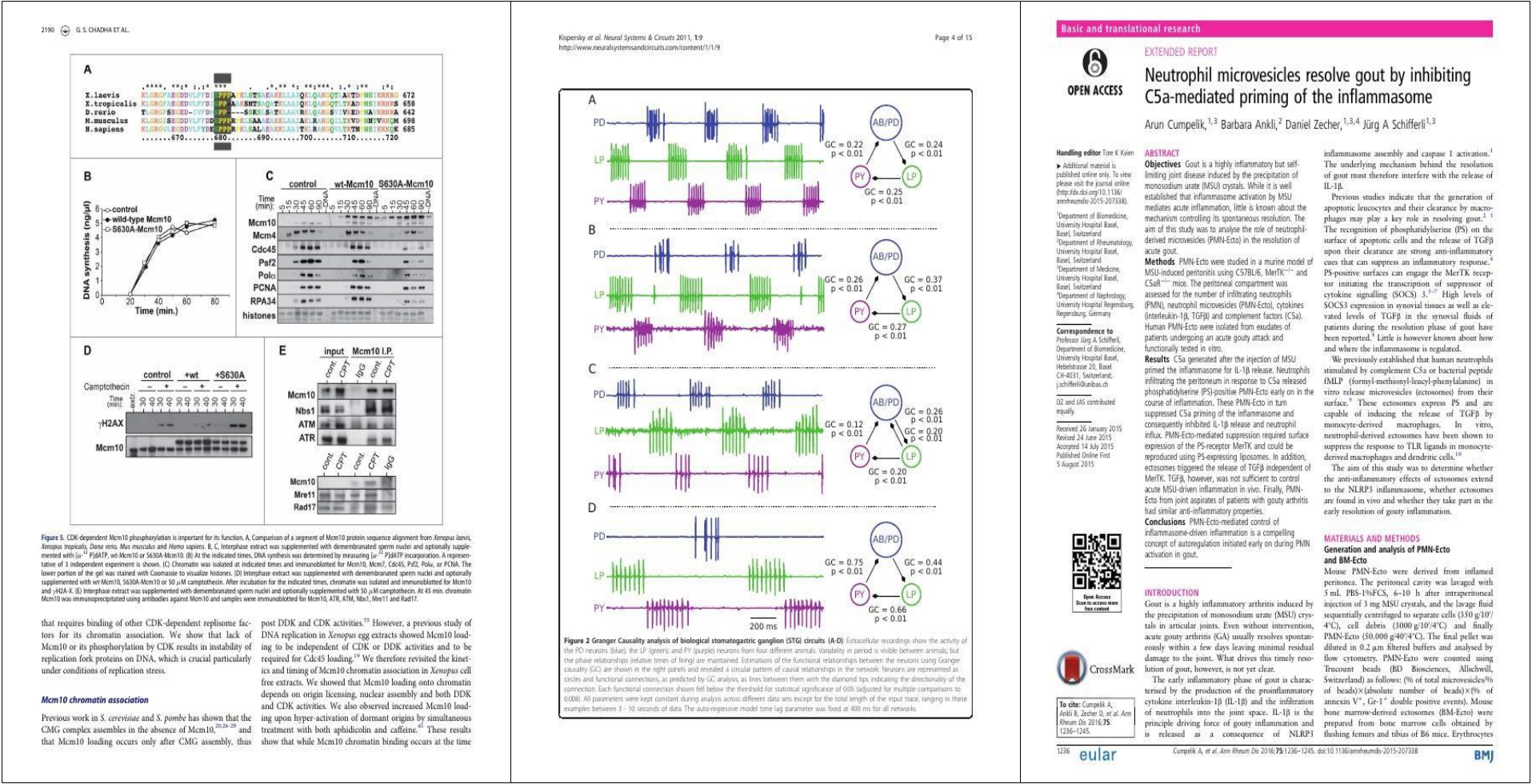}
        \end{minipage} \\

        \cmidrule(r){1-1}
        \cmidrule(lr){2-2}
        \cmidrule(lr){3-3}
        \cmidrule(lr){4-4}
        \cmidrule(lr){5-5}
        \cmidrule(l){6-6}

        DocLayNet
        & 69,103
        & 6,480
        & \makecell{
            Financial, Scientific,\\
            Patents, Manuals,\\
            Laws, Tenders
        }
        & \makecell{
            Caption, Footnote, Formula,\\
            List-item, Page-footer, Page-header,\\
            Picture, Section-header, Table,\\
            Text, Title
        }
        & \begin{minipage}{.3\textwidth}
            \centering
            \includegraphics[
                width=\linewidth,
                height=25mm
            ]{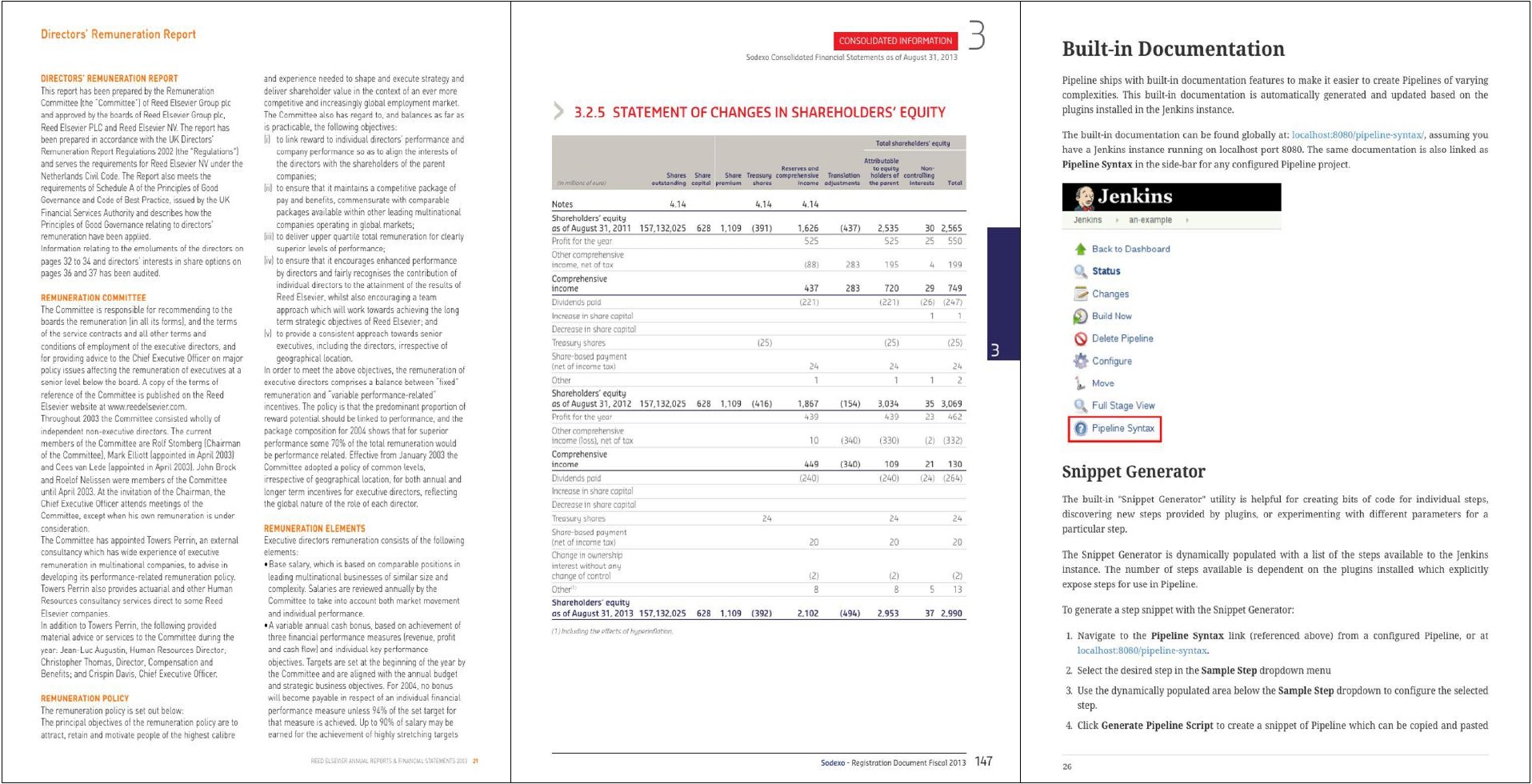}
        \end{minipage} \\

        \cmidrule(r){1-1}
        \cmidrule(lr){2-2}
        \cmidrule(lr){3-3}
        \cmidrule(lr){4-4}
        \cmidrule(lr){5-5}
        \cmidrule(l){6-6}

        IndicDLP
        & 95,824
        & 11,994
        & \makecell{
            Acts \& Rules, Brochures, Forms,\\
            Magazines, Manuals, Newspapers,\\
            Notices, Novels, Question Papers,\\
            Research Papers, Syllabi, Textbooks
        }
        & 42 fine-grained categories
        & \begin{minipage}{.3\textwidth}
            \centering
            \includegraphics[
                width=\linewidth,
                height=25mm
            ]{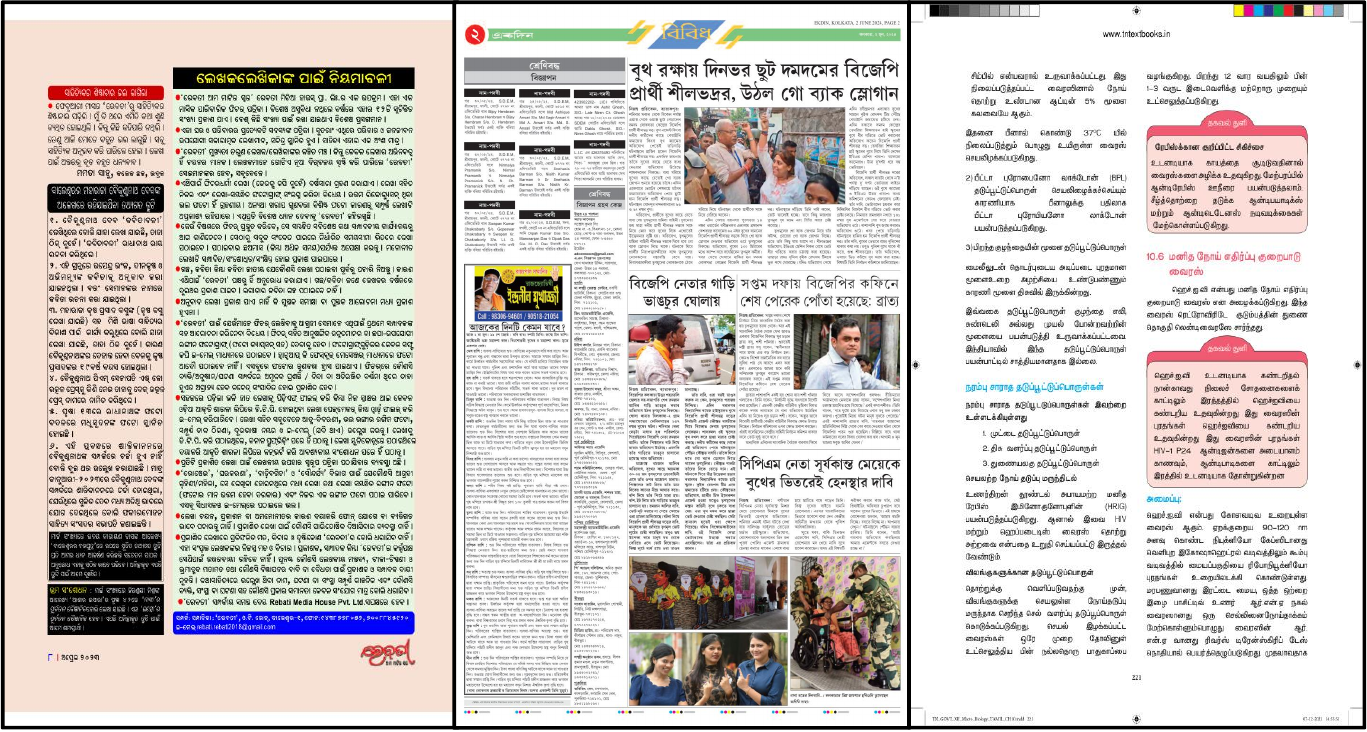}
        \end{minipage} \\

        \bottomrule
        \end{tabular}
    }
    \label{tab:benchmark_detail}
\end{table*}
\clearpage
\subsection{Comparison Methods}

We compare DocLayout-YOLO with representative methods from three categories:
(1) real-time object detectors, which focus on achieving favorable accuracy--efficiency trade-offs;
(2) specialized unimodal DLA methods, which rely primarily on visual information for document layout understanding;
and (3) specialized multimodal DLA methods, which leverage additional OCR annotation and language semantic representation through vision-language modeling.
The complete list of comparison methods is summarized in Tab.~\ref{tab:comparison_methods}.

\begin{table}[h]
\centering
\caption{Summary of comparison methods.}
\label{tab:comparison_methods}

\resizebox{1.0\textwidth}{!}{
\begin{tabular}{l|l}
\toprule
\textbf{Category} & \textbf{Methods} \\
\midrule

\textbf{Real-time Detectors}
&
YOLOv8~\citep{Jocher_Ultralytics_YOLO_2023}, 
YOLOv10~\citep{DBLP:journals/corr/abs-2405-14458},
\\
&
YOLOv11~\citep{khanam2024yolov11},
YOLOv12~\citep{tian2026yolov12},
\\
&
YOLOv26~\citep{jocher2026ultralyticsyolo26},
RT-DETRv3~\citep{DBLP:conf/wacv/WangXLS25}
\\

\midrule

\textbf{Specialized DLA}
&
\multicolumn{1}{l}{\textit{Unimodal}}\\
&
LayoutParser~\citep{shen2021layoutparser},
DocSegTr~\citep{biswas2022docsegtr},
\\
&
DiT-L~\citep{DBLP:conf/mm/LiXL0ZW22},
SelfDocSeg~\citep{DBLP:conf/icdar/MaityBMBLBP23},
SwinDocSegmenter~\citep{banerjee2023swindocsegmenter},
\\
&
GraphKD~\citep{DBLP:conf/icdar/BanerjeeBLP24},
SemiDocSeg~\citep{banerjee2024semidocseg},
RoDLA~\citep{chen2024rodla},
\\
&
PARL~\citep{liu2026parl},
HybriDLA~\citep{chen2026hybridla},
DocSAM~\citep{li2025docsam}
\\

\midrule

\textbf{Specialized DLA}
&
\multicolumn{1}{l}{\textit{Multimodal}}\\
&
LayoutLMv3~\citep{DBLP:conf/mm/HuangL0LW22},
VGT~\citep{DBLP:conf/iccv/DaLZY23},
\\
&
M2Doc~\citep{zhang2024m2doc},
DoPTA-HR~\citep{DBLP:journals/corr/abs-2412-12902},
\\
&
PromptDLA~\citep{zhang2026promptdla}
\\

\bottomrule
\end{tabular}
}

\end{table}

\clearpage

\section{More Experiments}

\subsection{Results on More Comparison Datasets}

As shown in Tab.~\ref{tab:large_scale_benchmarks}, we further evaluate DocLayout-YOLO on three additional document layout benchmarks, including PubLayNet, DocLayNet, and IndicDLP. Compared to YOLOv12-X, DocLayout-YOLO-X consistently improves mAP by 0.6\%/1.3\%/0.6\%, respectively. The consistent gains across datasets with different document domains and layout characteristics further demonstrate the generalization ability of our proposed method.

\begin{table}[h]
    \centering
    \caption{
        Comparison results on PubLayNet, DocLayNet, and IndicDLP.
    }
    \label{tab:large_scale_benchmarks}
    \resizebox{0.6\linewidth}{!}{
        \begin{tabular}{l|ccc}
            \toprule
            \textbf{Method}
            & \textbf{PubLayNet}
            & \textbf{DocLayNet}
            & \textbf{IndicDLP} \\
            \midrule

            YOLOv12-X
            & 96.9 & 81.0 & 56.3 \\

            \rowcolor{palerowgreen}
            \textbf{DocLayout-YOLO-X}
            & \textbf{97.5}
            & \textbf{82.3}
            & \textbf{56.9} \\

            \bottomrule
        \end{tabular}
    }
\end{table}
\subsection{Robustness Evaluation}

Tabs.~\ref{tab:publaynet_p_robustness} and~\ref{tab:m6doc_p_robustness} evaluate robustness under 12 document perturbations at three severity levels on PubLayNet-P and M\textsuperscript{6}Doc-P, respectively~\citep{chen2024rodla}. P-Avg denotes the average accuracy over all perturbations, while mRD measures normalized performance degradation, with lower mRD and higher P-Avg indicating stronger robustness. On PubLayNet-P, DocLayout-YOLO-X achieves 97.5\% clean mAP, 85.8\% P-Avg, and 48.7 mRD, improving over baseline YOLOv12-X by 0.6\%/9.1\% in clean mAP/P-Avg and reducing mRD by 30.7. It also surpasses the previous best RoDLA by 15.8\% on P-Avg while lowering mRD by 67.3. On M\textsuperscript{6}Doc-P, DocLayout-YOLO-X achieves 75.5\% clean mAP, 68.0\% P-Avg, and 118.8 mRD, outperforming the baseline YOLOv12-X by 1.5\%/0.3\% in clean mAP/P-Avg with a 2.6 reduction in mRD. Compared with the previous best results, it further improves clean mAP/P-Avg by 5.3\%/6.3\% and reduces mRD by 28.8. These results suggest that DocLayout-YOLO provides improved robustness to diverse document degradations without noticeably compromising clean-document performance.

\begin{table*}[h]
\centering
\caption{
{Robustness evaluation on the PubLayNet-P dataset.}
V, L, and T denote visual, layout, and textual modalities, respectively.
\textbf{Best} and \underline{second-best} results are highlighted.
}
\label{tab:publaynet_p_robustness}

\scriptsize
\setlength{\tabcolsep}{2.3pt}
\renewcommand{\arraystretch}{1.05}

\resizebox{\textwidth}{!}{
\begin{tabular}{l|l|ccc|c|ccc|cc|ccc|cc|cc|cc}
\toprule
\multirow{2}{*}{\textbf{Method}}
& \multirow{2}{*}{\textbf{Backbone}}
& \multicolumn{3}{c|}{\textbf{Modality}}
& \multirow{2}{*}{\textbf{Clean}}
& \multicolumn{3}{c|}{\textbf{Spatial}}
& \multicolumn{2}{c|}{\textbf{Content}}
& \multicolumn{3}{c|}{\textbf{Inconsistency}}
& \multicolumn{2}{c|}{\textbf{Blur}}
& \multicolumn{2}{c|}{\textbf{Noise}}
& \multirow{2}{*}{\textbf{P-Avg}$\uparrow$}
& \multirow{2}{*}{\textbf{mRD}$\downarrow$} \\
& & \textbf{V} & \textbf{L} & \textbf{T}
&
& \textbf{P1} & \textbf{P2} & \textbf{P3}
& \textbf{P4} & \textbf{P5}
& \textbf{P6} & \textbf{P7} & \textbf{P8}
& \textbf{P9} & \textbf{P10}
& \textbf{P11} & \textbf{P12}
& & \\
\midrule

LayoutParser~\citep{shen2021layoutparser}
& ResNeXt~\citep{DBLP:conf/cvpr/XieGDTH17}
& \Checkmark & \XSolidBrush & \XSolidBrush
& 89.0
& 35.8 & 78.1 & 68.0
& 84.8 & 45.9
& 79.0 & 85.8 & 82.6
& 51.4 & 31.1
& 53.6 & 0.3
& 58.0 & 212.7 \\

Faster R-CNN~\citep{ren2015faster}
& ResNet~\citep{DBLP:conf/cvpr/HeZRS16}
& \Checkmark & \XSolidBrush & \XSolidBrush
& 90.2
& \textbf{44.2} & 75.3 & 74.3
& 78.9 & 53.2
& 81.4 & 82.7 & 80.2
& 80.6 & 63.2
& 55.7 & 24.3
& 66.2 & 175.5 \\

DocSegTr~\citep{biswas2022docsegtr}
& ResNet~\citep{DBLP:conf/cvpr/HeZRS16}
& \Checkmark & \XSolidBrush & \XSolidBrush
& 90.4
& 28.3 & 76.0 & 71.2
& 85.2 & 46.2
& 69.3 & 86.2 & 68.9
& 50.8 & 19.8
& 60.2 & 0.4
& 55.2 & 233.0 \\

Mask R-CNN~\citep{DBLP:conf/iccv/HeGDG17}
& ResNet~\citep{DBLP:conf/cvpr/HeZRS16}
& \Checkmark & \XSolidBrush & \XSolidBrush
& 91.0
& \underline{40.0} & 74.0 & 71.8
& 76.9 & 48.7
& 79.0 & 80.6 & 77.4
& 78.2 & 54.2
& 55.1 & 31.9
& 64.0 & 192.7 \\

SwinDocSegmenter~\citep{banerjee2023swindocsegmenter}
& Swin~\citep{DBLP:conf/iccv/LiuLCHWYZLG21}
& \Checkmark & \XSolidBrush & \XSolidBrush
& 93.7
& 39.0 & 83.3 & 61.3
& 88.9 & 46.6
& 87.9 & 88.1 & 86.6
& 37.3 & 29.2
& 36.1 & 0.5
& 57.1 & 214.4 \\

Cascade R-CNN~\citep{DBLP:conf/cvpr/CaiV18}
& DiT~\citep{DBLP:conf/mm/LiXL0ZW22}
& \Checkmark & \XSolidBrush & \XSolidBrush
& 94.5
& 31.9 & \textbf{86.5} & \textbf{82.2}
& 92.1 & 79.6
& 87.2 & 92.0 & 91.6
& 93.8 & 71.3
& 69.9 & 41.8
& 76.7 & 95.8 \\

Cascade R-CNN~\citep{DBLP:conf/cvpr/CaiV18}
& LayoutLMv3~\citep{DBLP:conf/mm/HuangL0LW22}
& \Checkmark & \Checkmark & \Checkmark
& 95.1
& 32.7 & \underline{85.9} & 79.8
& 92.3 & 68.5
& 86.5 & 93.1 & 86.7
& 82.9 & 47.0
& \underline{82.1} & \underline{45.1}
& 73.6 & 116.2 \\

\midrule

Co-DINO~\citep{DBLP:conf/iccv/ZongS023}
& Swin~\citep{DBLP:conf/iccv/LiuLCHWYZLG21}
& \Checkmark & \XSolidBrush & \XSolidBrush
& 94.3
& 22.4 & 43.0 & 24.7
& 92.6 & 56.8
& 86.4 & 72.6 & 75.8
& 35.0 & 20.2
& 55.4 & 14.9
& 50.0 & 254.1 \\

Cascade R-CNN~\citep{DBLP:conf/cvpr/CaiV18}
& InternImage~\citep{wang2023internimage}
& \Checkmark & \XSolidBrush & \XSolidBrush
& 94.1
& 27.7 & 80.2 & 79.7
& 89.6 & 56.6
& 83.8 & 91.6 & 89.6
& 84.9 & 54.0
& 55.2 & 0.2
& 66.1 & 141.9 \\

DINO~\citep{DBLP:conf/iclr/0097LL000NS23}
& InternImage~\citep{wang2023internimage}
& \Checkmark & \XSolidBrush & \XSolidBrush
& 95.4
& 34.4 & 82.2 & \textbf{82.2}
& 92.3 & 57.7
& 91.7 & 92.8 & 92.1
& 90.9 & 63.4
& 47.3 & 1.3
& 69.0 & 120.7 \\

Co-DINO~\citep{DBLP:conf/iccv/ZongS023}
& InternImage~\citep{wang2023internimage}
& \Checkmark & \XSolidBrush & \XSolidBrush
& 94.2
& 19.1 & 48.3 & 35.7
& 91.3 & 60.6
& 87.9 & 81.4 & 87.6
& 37.9 & 27.9
& 49.7 & 11.2
& 53.2 & 230.3 \\

RoDLA~\citep{chen2024rodla}
& InternImage~\citep{wang2023internimage}
& \Checkmark & \XSolidBrush & \XSolidBrush
& 96.0
& 31.6 & 79.3 & 80.6
& 92.9 & 61.6
& 91.6 & 92.6 & 91.6
& 91.3 & 67.7
& 58.8 & 0.3
& 70.0 & 116.0 \\

\midrule

YOLOv12-X~\citep{tian2026yolov12}
& v12-X
& \Checkmark & \XSolidBrush & \XSolidBrush
& \underline{96.9}
& 29.4 & 79.5 & \underline{81.7}
& \underline{95.9} & \underline{87.1}
& \underline{95.0} & \underline{96.7} & \underline{95.9}
& \underline{96.2} & \underline{72.9}
& 76.9 & 13.2
& \underline{76.7} & \underline{79.4} \\

\rowcolor{palerowgreen}
\textbf{DocLayout-YOLO-X}
& \textbf{v12-X + PALA}
& \textbf{\Checkmark}
& \textbf{\XSolidBrush}
& \textbf{\XSolidBrush}
& \textbf{97.5}
& 21.2 & 85.7 & 80.3
& \textbf{97.4} & \textbf{96.4}
& \textbf{97.5} & \textbf{97.2} & \textbf{96.3}
& \textbf{96.4} & \textbf{76.3}
& \textbf{89.3} & \textbf{95.9}
& \textbf{85.8} & \textbf{48.7} \\

\bottomrule
\end{tabular}
}
\end{table*}

\begin{table*}[h]
\centering
\caption{
{Robustness evaluation on the M\textsuperscript{6}Doc-P dataset.}
}
\label{tab:m6doc_p_robustness}

\resizebox{0.82\textwidth}{!}{
\begin{tabular}{l|l|ccc|ccc}
\toprule
\multirow{2}{*}{\textbf{Method}}
& \multirow{2}{*}{\textbf{Backbone}}
& \multicolumn{3}{c|}{\textbf{Modality}}
& \multirow{2}{*}{\textbf{Clean}}
& \multirow{2}{*}{\textbf{P-Avg}$\uparrow$}
& \multirow{2}{*}{\textbf{mRD}$\downarrow$} \\
& & \textbf{V} & \textbf{L} & \textbf{T}
& & & \\
\midrule

DocSegTr~\citep{biswas2022docsegtr}
& ResNet~\citep{DBLP:conf/cvpr/HeZRS16}
& \Checkmark & \XSolidBrush & \XSolidBrush
& 60.3 & 43.2 & 212.6 \\

Mask R-CNN~\citep{DBLP:conf/iccv/HeGDG17}
& ResNet~\citep{DBLP:conf/cvpr/HeZRS16}
& \Checkmark & \XSolidBrush & \XSolidBrush
& 61.9 & 48.9 & 216.7 \\

Faster R-CNN~\citep{ren2015faster}
& ResNet~\citep{DBLP:conf/cvpr/HeZRS16}
& \Checkmark & \XSolidBrush & \XSolidBrush
& 62.0 & 49.6 & 192.2 \\

SwinDocSegmenter~\citep{banerjee2023swindocsegmenter}
& Swin~\citep{DBLP:conf/iccv/LiuLCHWYZLG21}
& \Checkmark & \XSolidBrush & \XSolidBrush
& 47.1 & 39.7 & 239.2 \\

Cascade R-CNN~\citep{DBLP:conf/cvpr/CaiV18}
& DiT~\citep{DBLP:conf/mm/LiXL0ZW22}
& \Checkmark & \XSolidBrush & \XSolidBrush
& 70.2 & 60.6 & 164.8 \\

Cascade R-CNN~\citep{DBLP:conf/cvpr/CaiV18}
& LayoutLMv3~\citep{DBLP:conf/mm/HuangL0LW22}
& \Checkmark & \Checkmark & \Checkmark
& 64.3 & 57.0 & 176.1 \\

RoDLA~\citep{chen2024rodla}
& InternImage~\citep{wang2023internimage}
& \Checkmark & \XSolidBrush & \XSolidBrush
& 70.0 & 61.7 & 147.6 \\

\midrule

YOLOv12-X~\citep{tian2026yolov12}
& v12-X
& \Checkmark & \XSolidBrush & \XSolidBrush
& \underline{74.0}
& \underline{67.7}
& \underline{121.4} \\

\rowcolor{palerowgreen}
\textbf{DocLayout-YOLO-X}
& \textbf{v12-X + PALA}
& \textbf{\Checkmark}
& \textbf{\XSolidBrush}
& \textbf{\XSolidBrush}
& \textbf{75.5}
& \textbf{68.0}
& \textbf{118.8} \\

\bottomrule
\end{tabular}
}
\end{table*}
\subsection{Heatmap Visualization}

\begin{figure}[h]
    \centering
    \includegraphics[width=1.0\linewidth]{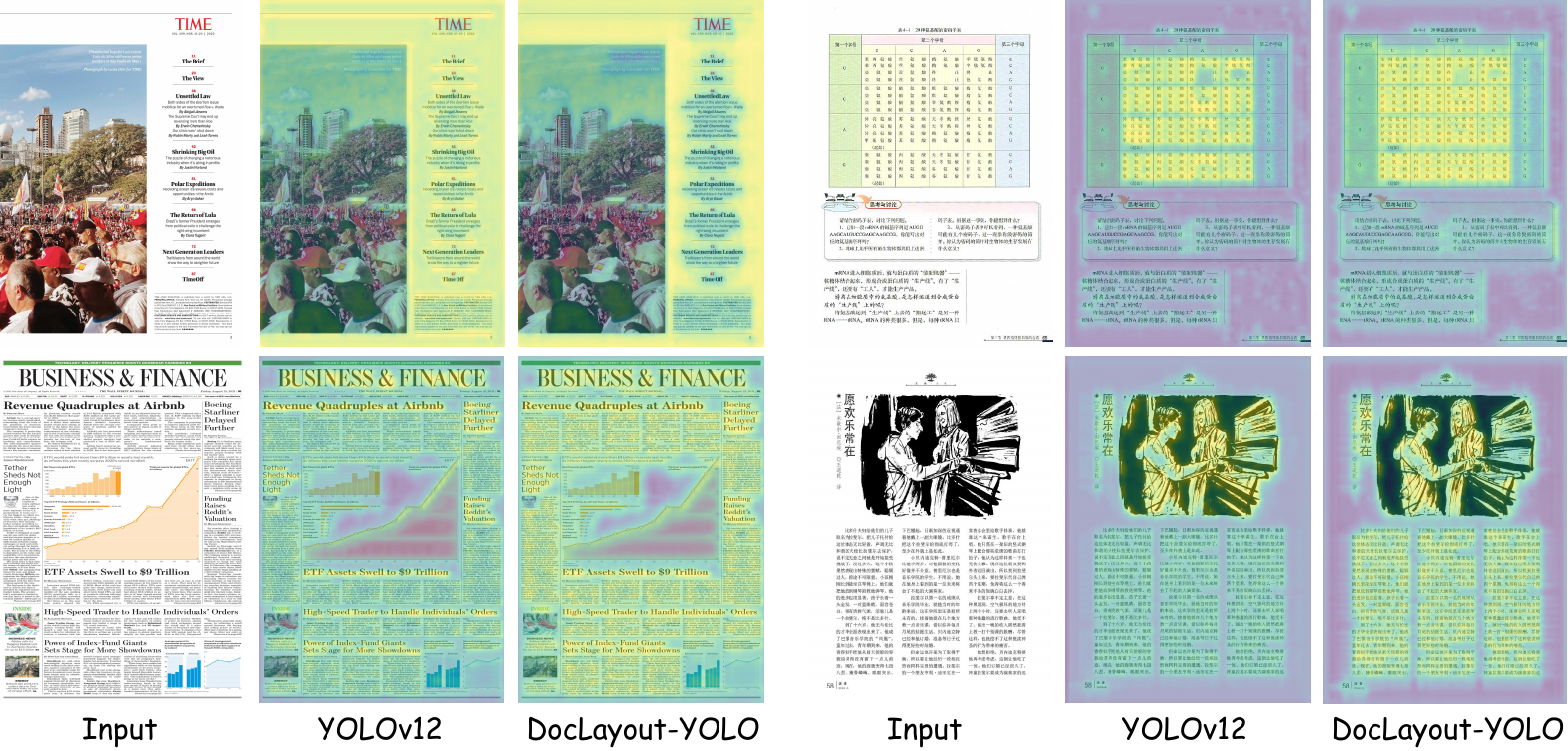}
    \caption{Comparison of heatmaps between YOLOv12 and the proposed DocLayout-YOLO. DocLayout-YOLO exhibits clearer object perception than YOLOv12.}
    \label{fig:heatmap}
\end{figure}

To further examine the learned spatial representations, we visualize the third-stage (S3) feature maps of the X-scale  YOLOv12 baseline and DocLayout-YOLO. As shown in Fig.~\ref{fig:heatmap}, the baseline YOLOv12 produces relatively diffuse responses, particularly on pages with dense text, tabular structures, and multi-column layouts. In contrast, DocLayout-YOLO generates more structured activations that better align with semantically meaningful regions, including text blocks, table cells, figures, and their surrounding boundaries. This improvement is especially evident in layouts with strong horizontal and vertical dependencies, where PALA maintains coherent responses across related elements while suppressing irrelevant background activation. These observations qualitatively support the effectiveness of PALA in enhancing long-range page-context modeling and element-level representation for document layout analysis.

\subsection{Design-heuristic Metric Evaluation Evaluation on MeshLayout-1M}
\label{sec:align}

In this section, we quantitatively evaluate whether the synthetic document page produced by Mesh-Candidate BestFit aligns with the human design principle. The evaluation employs the \textit{Align} and \textit{Density} metrics, which respectively measure the aesthetic quality of layouts in terms of document alignment and density. For \textit{Align}, we utilize the LayoutGAN++~\citep{DBLP:conf/mm/KikuchiSOY21,DBLP:journals/tvcg/LiY0LWX21} metric which measures the alignment of elements in the document:
\begin{equation}
    L_{alg} = \sum_{i=1}^N \min \left(
    \begin{array}{rl}
    g(\Delta x_{i}^{L}),g(\Delta x_{i}^{C}),g(\Delta x_{i}^{R}) \\
    g(\Delta y_{i}^{T}),g(\Delta y_{i}^{C}),g(\Delta y_{i}^{B})
    \end{array}
    \right).
\end{equation}

where $x_{i}^{*}(*=L,C,R),y_{i}^{*}(*=T,C,B)$ denotes the x-axis left/center/right and y-axis top/center/bottom of $i$-th elements in document, $g(x)=-\log(1-x)$, and $\Delta x_{i}^{*}(*=L,C,R)$ is computed as:
\begin{equation}
    \Delta x_{i}^{*} = \displaystyle \min_{\forall j \neq i} | x_{i}^{*} - x_{j}^{*} |
\end{equation}

$\Delta y_{i}^{*}(*=T,C,B)$ can be computed similarly. For \textit{Density}, we calculate the ratio of filled area in the layout:
\begin{equation}
    L_{dst} = \frac{\sum_{i=1}^N |e_i|}{|L|}
\end{equation}

\begin{wraptable}{r}{0.50\linewidth}
    \centering
    \caption{Quantitative comparison between different layout generation methods.}
    \resizebox{1.0\linewidth}{!}{
    \begin{tabular}{c|cc}
    \toprule
    \textbf{Layout Generation} & \textbf{Align}$\downarrow$ & \textbf{Density}$\uparrow$ \\
    \midrule
    Random & 0.0171 & 0.259 \\
    Diffusion~(LACE) & 0.0032 & 0.476 \\
    \midrule
    \textbf{Mesh-candidate BestFit (ours)} 
    & \textbf{0.0009} & \textbf{0.645} \\
    \bottomrule
    \end{tabular}
    }
    \label{tab:tab_metric}
\end{wraptable}

where $|e_i|$ denotes area of element $e_i$ in $L$, and $|L|$ denotes area of the whole layout. For \textit{Align}, a lower value denotes a more aligned document. For \textit{Density}, a larger value denotes a more compact and dense layout. The experimental results, as shown in Tab.~\ref{tab:tab_metric}. Results indicate that the Mesh-candidate BestFit method significantly outperforms diffusion and random methods in both alignment and density. Visual results further confirm that the layouts produced by Mesh-candidate BestFit better conform to the standards of human aesthetics and design principles.

\subsection{Ablation on Pretraining Data Volume}
\label{sec:volume}

\begin{figure}[h]
  \centering
  \includegraphics[width=0.9\linewidth]{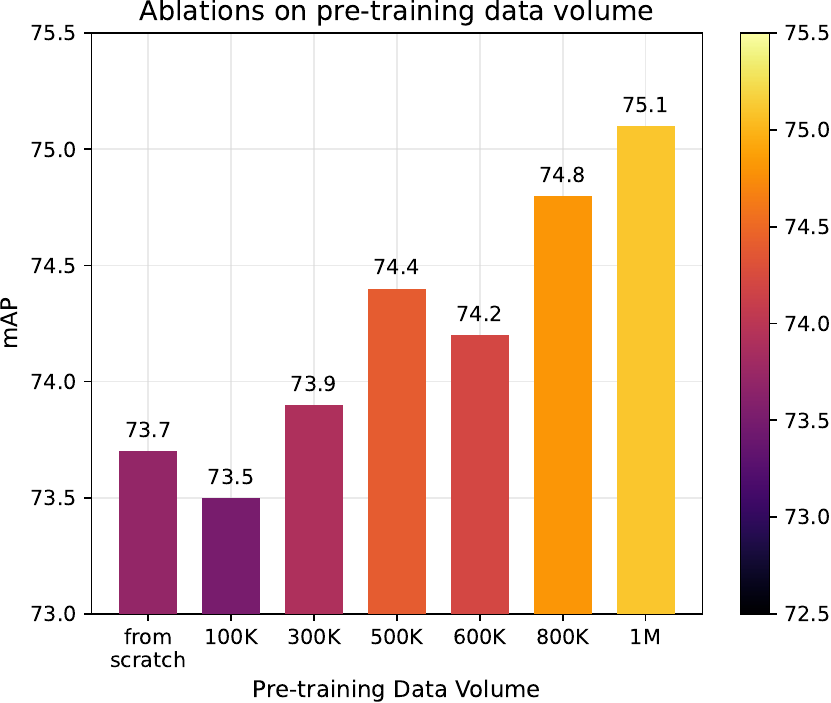}
  \caption{Ablations on pretraining data volume.}
  \label{fig:data}
\end{figure}


We conduct ablation experiments on the pretraining data volume of MeshLayout-1M. Specifically, we pretrain DocLayout-YOLO-L using different amounts of Mesh-candidate generated data, ranging from 0 to 1M pages, and subsequently fine-tune it on M\textsuperscript{6}Doc. Results are shown in Fig.~\ref{fig:data}. Overall, model performance improves as the pretraining data volume increases, despite minor fluctuations at intermediate scales. The mAP rises from 73.7\% without pretraining to 74.4\% with 500K pages, reaches 74.8\% at 800K, and further improves to 75.1\% with 1M pages. These results indicate a clear benefit from scaling the synthetic pretraining corpus, while the marginal gains become smaller at larger data volumes.
\subsection{Effects of MeshLayout-1M Pretraining on Long-tail Distribution}

\begin{figure*}[h]
    \centering
    \begin{subfigure}{1.0\textwidth}
        \includegraphics[width=\linewidth]{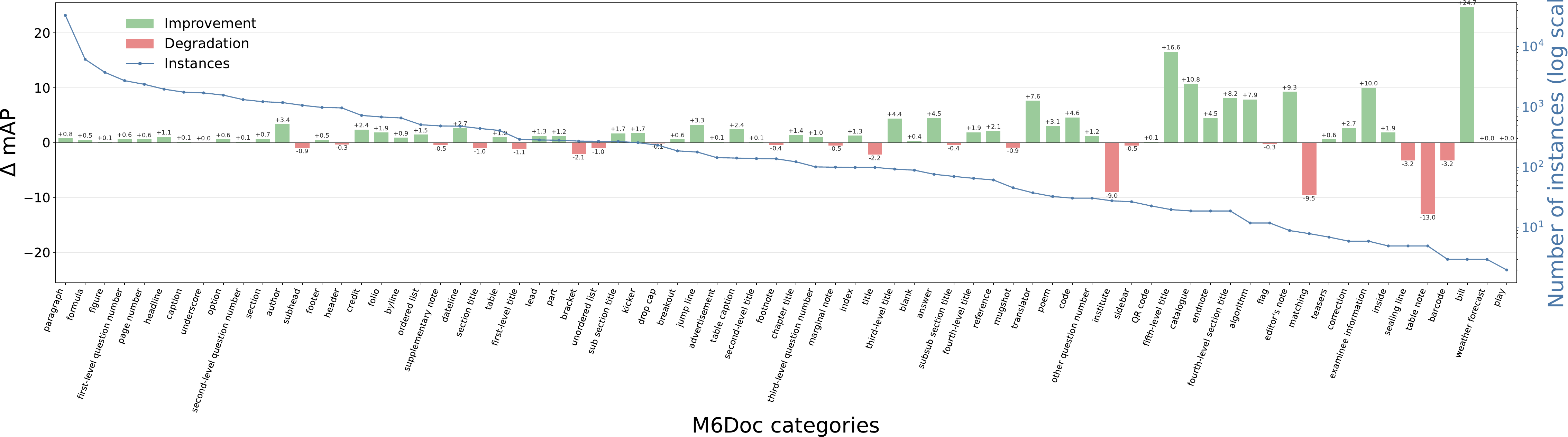}
        \caption{M\textsuperscript{6}Doc.}
        \label{fig:fig_classwise_m6doc}
    \end{subfigure}
    \begin{subfigure}{1.0\textwidth}
        \includegraphics[width=\linewidth]{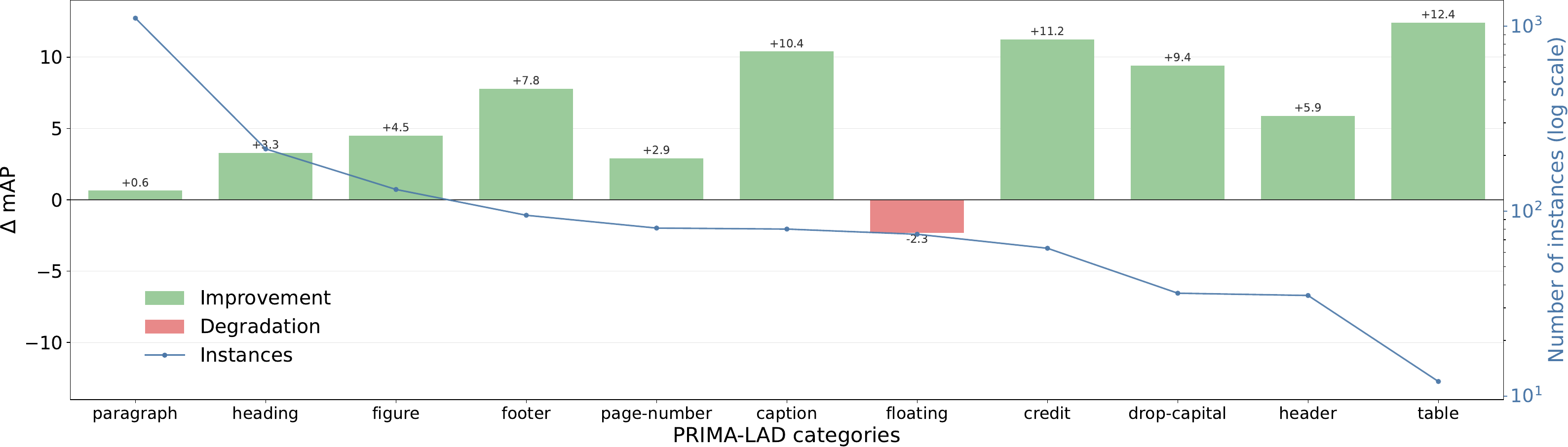}
        \caption{PRIMA-LAD.}
        \label{fig:fig_classwise_prima_lad}
    \end{subfigure}
    \caption{Class-wise metric analysis. Left Y-Axis indicate class-wise AP and right Y-Axis indicate number of instances for each category.}
\end{figure*}

We analyze the effect of MeshLayout-1M pretraining under long-tailed category distributions. The class-wise results on M\textsuperscript{6}Doc and PRIMA-LAD are shown in Fig.~\ref{fig:fig_classwise_m6doc} and Fig.~\ref{fig:fig_classwise_prima_lad}, respectively. The left y-axis reports the AP change for each category, while the right y-axis shows the number of instances. $\Delta$ mAP denotes the performance difference between DocLayout-YOLO-L with and without MeshLayout-1M pretraining.

Overall, MeshLayout-1M yields improvements across a broad range of downstream categories, although a small number of classes exhibit performance degradation. The gains are particularly pronounced for several low-frequency categories, suggesting that diverse synthetic pretraining can alleviate long-tail effects to some extent. For example, on M\textsuperscript{6}Doc, the improvements reach 7.6\% for translator, 10.8\% for catalog, and 9.3\% for editor's note. On PRIMA-LAD, caption, credit, and table improve by 10.4\%, 11.2\%, and 12.4\%, respectively.

\clearpage

\section{Additional Analysis of Page-Axis Layout Attention~(PALA)}
\label{sec:appendix_pala}

\subsection{More Analysis \& Comparisons of Different Local Attentions}
\label{sec:appendix_attention_analysis}

\begin{figure}[h]
    \centering
    \includegraphics[width=1.0\linewidth]{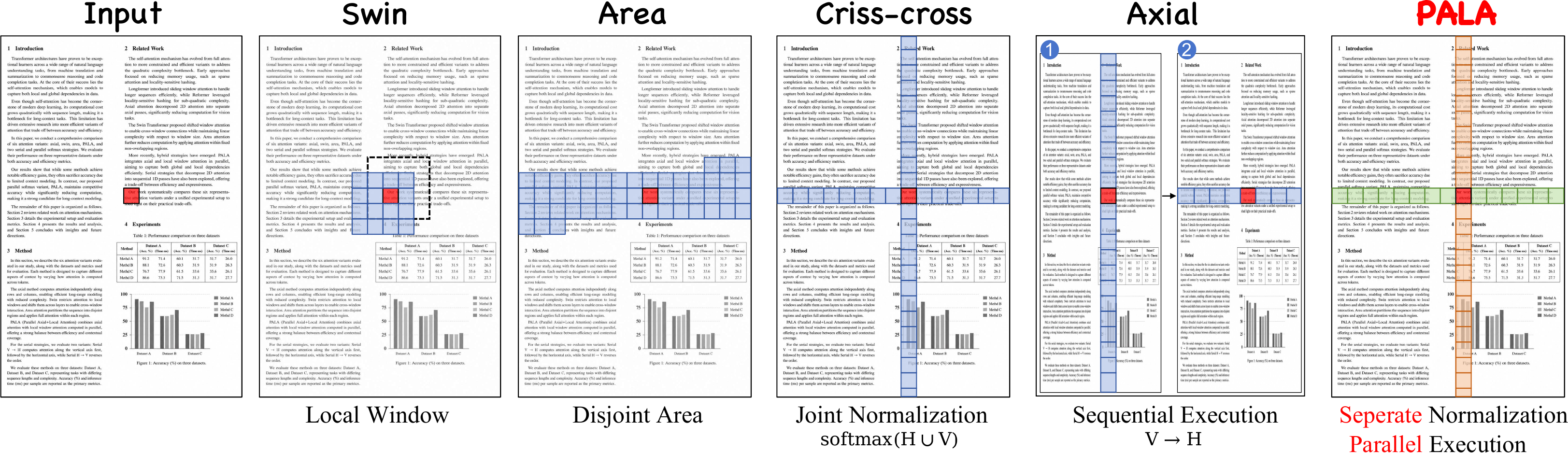}
    \caption{Comparison of representative local attention mechanisms with our proposed PALA.}
    \label{fig:attn_comp}
\end{figure}

\paragraph{Comparison with Different Attention Mechanisms.}
Fig.~\ref{fig:attn_comp} illustrates the structural differences between PALA and representative local attention mechanisms. Swin Attention restricts interactions within local windows, while Area Attention partitions the feature map into disjoint regions, preventing direct context aggregation across region boundaries. Criss-Cross Attention extends interactions along both spatial axes in parallel, but jointly normalizes horizontal and vertical attention weights. Axial Attention instead performs separately normalized axis-wise attention sequentially, making the second direction dependent on the output of the first. In contrast, PALA performs horizontal and vertical attention in parallel with separate normalization, allowing the two directional contexts to be modeled independently while preserving long-range dependencies along both page axes. This design preserves complementary horizontal and vertical structural cues without local partitioning, cross-axis competition, or directional-order dependency.

\paragraph{Horizontal vs. Vertical Context Modeling.}
Following Tab.~\ref{tab:pala_axis_replacement_ablation}, we further isolate the horizontal and vertical branches of PALA in Tab.~\ref{tab:single_axis_analysis}. Horizontal-only consistently outperforms Vertical-only across both datasets and model scales, with gains of 0.2\%/1.3\% mAP on M\textsuperscript{6}Doc and 0.5\%/2.1\% mAP on PRIMA-LAD for the N/L models, respectively. We notice that this asymmetry is consistent with the geometry of document elements. As shown in Tab.~\ref{tab:document_box_geometry}, 89.0\% and 83.8\% of elements in M\textsuperscript{6}Doc and PRIMA-LAD are wider than tall, while 63.9\% and 58.2\% have aspect ratios of at least $2{:}1$, respectively. Horizontal attention therefore better covers elongated structures such as text lines, headings, captions, and table rows. Row-wise context may also be more semantically coherent, as tokens along the same row often belong to the same layout element. Nevertheless, vertical attention remains complementary for modeling paragraph columns, table columns, vertically stacked regions, and cross-region correspondences.

\begin{table}[h]
    \centering
    \caption{
    Comparison of Horizontal-only, Vertical-only, and complete PALA.
    }
    \label{tab:single_axis_analysis}
    \small
    \setlength{\tabcolsep}{4pt}
    \renewcommand{\arraystretch}{1.05}
    \begin{tabular}{llcccc}
        \toprule
        \textbf{Dataset}
        & \textbf{Size}
        & \textbf{Horizontal-only}
        & \textbf{Vertical-only}
        & \textbf{PALA}
        & \textbf{$\Delta (H-V)$} \\
        \midrule

        M\textsuperscript{6}Doc
        & N
        & 64.1
        & 63.9
        & 65.3
        & +0.2 \\

        M\textsuperscript{6}Doc
        & L
        & 72.7
        & 71.4
        & 73.7
        & +1.3 \\

        PRIMA-LAD
        & N
        & 60.6
        & 60.1
        & 62.6
        & +0.5 \\

        PRIMA-LAD
        & L
        & 67.0
        & 64.9
        & 69.1
        & +2.1 \\

        \bottomrule
    \end{tabular}
\end{table}

\begin{table*}[h]
\centering
\caption{
Geometric statistics of valid document elements in the official dataset splits.
Aspect ratios are computed from pixel-space axis-aligned bounding boxes.
}
\label{tab:document_box_geometry}

\resizebox{1.0\textwidth}{!}{
\begin{tabular}{lrrrrrr}
    \toprule
    \textbf{Dataset}
    & \textbf{\# Boxes}
    & \textbf{\shortstack{Wider than\\tall}}
    & \textbf{\shortstack{Aspect ratio\\$\geq 2{:}1$}}
    & \textbf{\shortstack{Aspect ratio\\$\geq 4{:}1$}}
    & \textbf{\shortstack{Mean\\aspect ratio}}
    & \textbf{\shortstack{Geometric-mean\\aspect ratio}} \\
    \midrule

    M\textsuperscript{6}Doc
    & 237,116
    & 89.0\%
    & 63.9\%
    & 41.5\%
    & 5.22
    & 3.19 \\

    PRIMA-LAD
    & 8,942
    & 83.8\%
    & 58.2\%
    & 40.1\%
    & 6.05
    & 3.02 \\

    \bottomrule
\end{tabular}
}

\end{table*}

\paragraph{Why Global Attention Underperforms Criss-Cross and PALA.}
Following Tab.~\ref{tab:pala_axis_replacement_ablation}, we conduct a diagnostic analysis of Global Attention as shown in Tab.~\ref{tab:global_attention_diagnostics}. Global Attention achieves the highest recall (76.8\%) among the three variants, but its precision is substantially lower than Criss-Cross and PALA (83.0\% vs.\ 86.7\%/88.1\%), resulting in the lowest mAP (64.5\% vs.\ 64.7\%/65.3\%). The higher recall but lower precision indicates that the performance degradation of Global Attention primarily stems from additional false positives rather than missed detections. This suggests that unrestricted global interactions may introduce interference between visually similar but spatially unrelated document regions. In contrast, Criss-Cross and PALA constrain long-range interactions along the page axes, reducing irrelevant cross-region context while preserving structurally relevant dependencies.

\begin{table}[h]
    \centering
    \caption{
    Detection diagnostics for Global, Criss-Cross, and PALA.
    }
    \label{tab:global_attention_diagnostics}
    \small
    \setlength{\tabcolsep}{4pt}
    \renewcommand{\arraystretch}{1.05}
    \begin{tabular}{llccc}
        \toprule
        \textbf{Dataset}
        & \textbf{Attention}
        & \textbf{Precision}
        & \textbf{Recall}
        & \textbf{mAP$_{50:95}$} \\
        \midrule

        M\textsuperscript{6}Doc
        & Global
        & 83.0
        & 76.8
        & 64.5 \\

        M\textsuperscript{6}Doc
        & Criss-Cross
        & 86.7
        & 75.5
        & 64.7 \\

        M\textsuperscript{6}Doc
        & PALA
        & 88.1
        & 76.3
        & 65.3 \\

        \bottomrule
    \end{tabular}
\end{table}


\subsection{Distinguishing PALA from Criss-Cross and Axial Attention}
\label{sec:appendix_pala_relation}

To clarify the relationship between PALA, Criss-Cross Attention, and Axial Attention, we compare their formulations along two key dimensions. Although all three attentions model long-range context along the horizontal and vertical axes, they differ in \textit{axis execution} (parallel vs.\ sequential) and \textit{attention normalization} (joint vs.\ separate), as summarized in Tab.~\ref{tab:pala_core_difference}.

\begin{table}[h]
    \centering
    \caption{Structural differences between Criss-Cross Attention, Axial Attention, and PALA.}
    \label{tab:pala_core_difference}
    \small
    \setlength{\tabcolsep}{7pt}
    \renewcommand{\arraystretch}{1.08}
    \begin{tabular}{lcc}
        \toprule
        \textbf{Method}
        & \textbf{Axis Execution}
        & \textbf{Attention Normalization} \\
        \midrule
        Criss-Cross Attention
        & Parallel
        & Joint \\
        Axial Attention
        & Sequential
        & Separate \\
        PALA
        & Parallel
        & Separate \\
        \bottomrule
    \end{tabular}
\end{table}

\paragraph{PALA vs.\ Criss-Cross Attention.}
Both PALA and Criss-Cross Attention aggregate horizontal and vertical context in parallel. Their main difference lies in how attention weights along the two axes are normalized. For a given query, let $\mathbf{s}_{hor}$ and $\mathbf{s}_{ver}$ denote the attention logits over its horizontal and vertical context, respectively. Criss-Cross Attention concatenates the two sets of logits and applies a joint softmax:
\begin{equation}
    \left[
        A_{hor}^{\mathrm{CC}},
        A_{ver}^{\mathrm{CC}}
    \right]
    =
    \operatorname{Softmax}
    \left(
        [\mathbf{s}_{hor},\mathbf{s}_{ver}]
    \right).
\end{equation}
Thus, horizontal and vertical context compete within a single attention distribution, such that strong responses along one axis can suppress attention along the other.

In contrast, PALA normalizes the two axes independently:
\begin{equation}
    A_{hor}^{\mathrm{PALA}}
    =
    \operatorname{Softmax}(\mathbf{s}_{hor}),
    \qquad
    A_{ver}^{\mathrm{PALA}}
    =
    \operatorname{Softmax}(\mathbf{s}_{ver}).
\end{equation}
This design treats horizontal and vertical context as two independent sources of structural information. Such separation is well suited to document layouts, where the two axes often represent distinct relationships, e.g., text lines and table rows horizontally, versus paragraph columns and vertically stacked regions vertically.

\paragraph{PALA vs.\ Axial Attention.}
Both PALA and Axial Attention normalize horizontal and vertical attention independently, but differ in how the two axes are executed. Axial Attention applies them sequentially:
\begin{equation}
    Y^{\mathrm{Axial}}
    =
    \mathcal{A}_{hor}
    \left(
        \mathcal{A}_{ver}(X)
    \right),
\end{equation}
where the second attention operation receives features already transformed by the first. PALA instead computes both directions in parallel from the same projected features:
\begin{equation}
    Y_{hor}=\mathcal{A}_{hor}(X),
    \qquad
    Y_{ver}=\mathcal{A}_{ver}(X).
\end{equation}
Neither branch depends on the output of the other, avoiding sequential dependency and directional-order sensitivity while preserving independently normalized context along both page axes.

\paragraph{Experimental Results}

As shown in Tab.~\ref{tab:pala_execution_normalization_ablation}, PALA achieves the best performance in three out of four settings. Compared with Sequential + Separate, parallel execution consistently improves mAP by 0.8\%--3.2\%, indicating the benefit of avoiding sequential dependency between the two axes. Compared with Parallel + Joint, separate normalization yields comparable performance on M\textsuperscript{6}Doc and improves PRIMA-LAD results by 0.2\%--2.8\%, suggesting that avoiding cross-axis competition generally provides more robust structural context modeling.

\begin{table}[h]
    \centering
    \caption{
        Ablation study of the execution and normalization strategies in PALA.
    }
    \label{tab:pala_execution_normalization_ablation}
    \small
    \setlength{\tabcolsep}{6pt}
    \renewcommand{\arraystretch}{1.08}
    \begin{tabular}{lc|ccc}
        \toprule
        \textbf{Dataset}
        & \textbf{Size}
        & \textbf{\makecell{Parallel + Separate\\(PALA)}}
        & \textbf{\makecell{Parallel\\+ Joint}}
        & \textbf{\makecell{Sequential\\+ Separate}} \\
        \midrule

        \multirow{2}{*}{M\textsuperscript{6}Doc}
        & N & 65.3 & \textbf{65.5} & 63.4 \\
        & L & \textbf{73.7} & 73.1 & 72.9 \\

        \midrule

        \multirow{2}{*}{PRIMA-LAD}
        & N & \textbf{62.6} & 59.8 & 59.6 \\
        & L & \textbf{69.1} & 67.6 & 65.9 \\

        \bottomrule
    \end{tabular}
\end{table}

\paragraph{Summary.}
PALA can therefore be distinguished from the two related designs by combining \textit{parallel axis execution} with \textit{separate axis normalization}. Compared with Criss-Cross Attention, PALA removes competition between horizontal and vertical attention distributions; compared with Axial Attention, PALA removes the sequential dependency between the two directional operations.

\subsection{Area Attention vs.\ PALA: Is PALA Always More Efficient?}
\label{sec:appendix_area_efficiency}

A natural question is whether PALA is always more efficient than Area Attention, or whether Area Attention can achieve lower latency by using finer area partitioning. We investigate this question from both theoretical complexity and practical inference performance.

\paragraph{Theoretical Analysis.}
Ignoring linear projections, the attention aggregation complexities of Global Attention, Area Attention with $A$ equal-sized areas, and PALA are $O(BH^{2}W^{2}C)$, $O(BH^{2}W^{2}C/A)$, and $O(BHW(H+W)C)$, respectively. Equating the latter two gives the complexity boundary
\begin{equation}
A^{*}=\frac{HW}{H+W}.
\end{equation}
When $A<A^{*}$, PALA has lower attention aggregation complexity than Area Attention, whereas sufficiently fine-grained partitioning ($A>A^{*}$) can make Area Attention theoretically more efficient. Therefore, PALA is not theoretically more efficient for arbitrary area configurations. We further examine their practical accuracy--efficiency trade-off under different area settings.

\begin{figure}[h]
    \centering
    \includegraphics[width=1.0\linewidth]{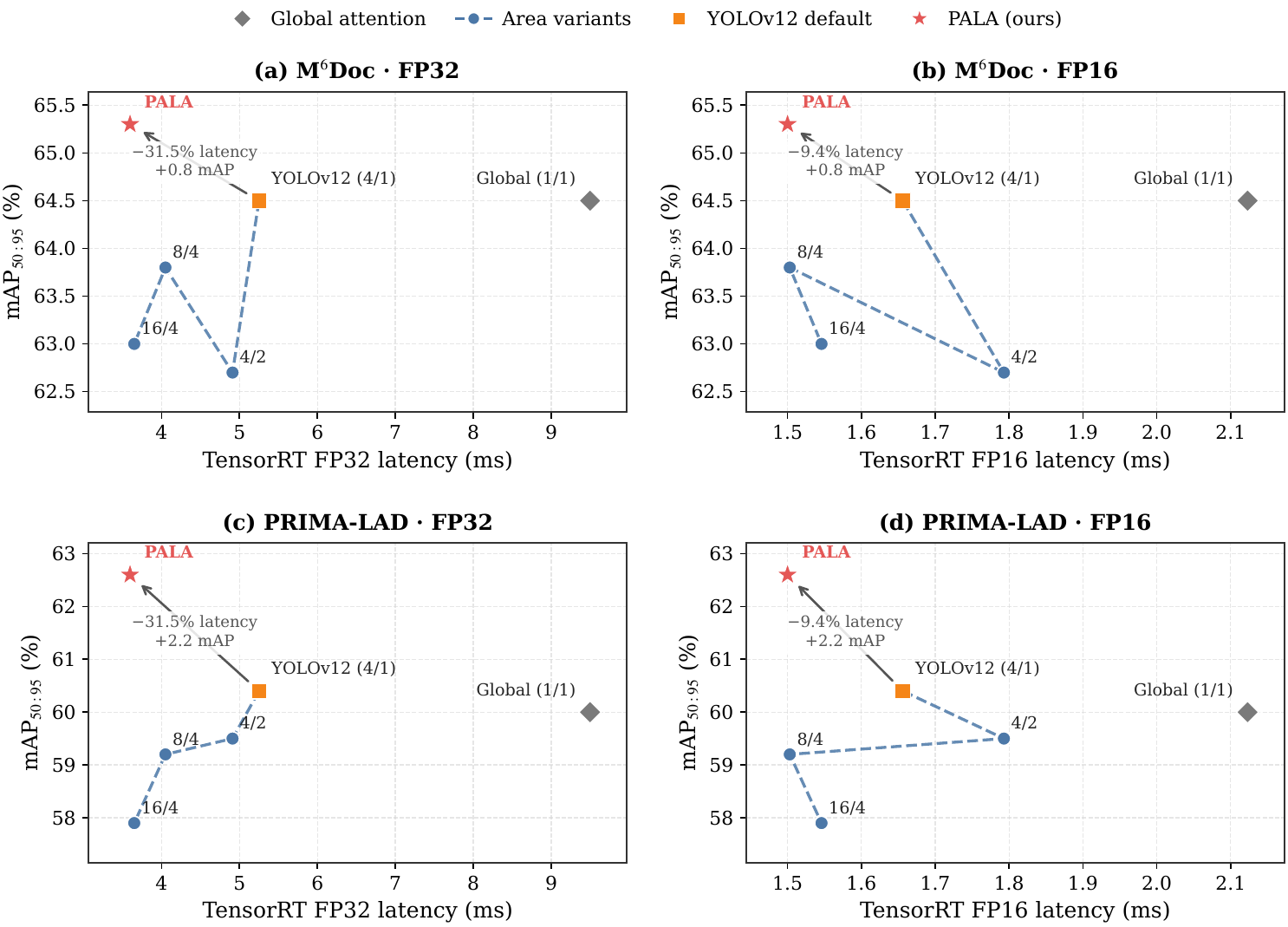}
    \caption{Accuracy–latency trade-offs of Global Attention, Area Attention with different ($A_{P4}/A_{P5}$) configurations, and PALA.}
    \label{fig:area_efficiency_analysis}
\end{figure}

\paragraph{Experimental Setting and Results.}
For a fair comparison, we train the YOLOv12-N model with different numbers of areas under the same $1600\times1600$ input resolution, training recipe, and TensorRT inference protocol. The accuracy-latency trade-off results are shown in Fig.~\ref{fig:area_efficiency_analysis}. For Area Attention, the notation $A_{P4}/A_{P5}$ denotes the numbers of areas used at the P4 and P5 feature levels, respectively; for example, $4/1$ is the default YOLOv12 configuration, while $16/4$ applies finer partitioning at both levels. At an input resolution of $1600\times1600$, the theoretical boundaries $A^{*}$ are 50 and 25 for the $A_{P4}$ and $A_{P5}$, respectively. Considering the default $4/1$ configuration in YOLOv12, we further evaluate $1/1$, $4/2$, $8/4$, and $16/4$ configurations to examine the accuracy--efficiency trade-off under different levels of area partitioning.

As shown in Fig.~\ref{fig:area_efficiency_analysis}, increasing the number of areas reduces the size of each attention region and generally leads to lower computation latency. Accordingly, the FP32 latency of Area Attention decreases from 5.250 ms at $4/1$ to 3.650 ms at $16/4$. However, finer partitioning also restricts context to increasingly smaller regions and leads to clear accuracy degradation: the $16/4$ configuration achieves only 63.0\%/57.9\% mAP on M\textsuperscript{6}Doc/PRIMA-LAD, compared with 64.5\%/60.4\% for the default $4/1$ configuration. PALA instead achieves 65.3\%/62.6\% mAP with 3.596/1.500 ms FP32/FP16 latency, preserving full horizontal and vertical interactions while remaining faster than the Area Attention configurations with comparable accuracy.

\paragraph{Summary.} In conclusion, Area Attention is not inherently less efficient than PALA: sufficiently fine partitioning can further reduce its computational cost, but at the expense of restricted context and detection accuracy. In summary, PALA provides a better speed-accuracy trade-off by combining efficient axis-wise attention with long-range document context modeling.

\clearpage

\section{Mesh-candidate BestFit Details}
\label{sec:appendix_bestfit}

\subsection{DocBank-based Element Pool Construction}

We construct the element pool exclusively from the training split of
DocBank~\citep{DBLP:conf/coling/LiXCHWLZ20}, rather than extracting elements
from any downstream evaluation benchmark. We exclude the \textit{Date} category because it is rare and visually homogeneous, being largely derived from LaTeX \texttt{\textbackslash date{...}}, \texttt{\textbackslash today}, or journal-template \texttt{Dated:} fields and thus prone to source-specific overfitting, and the \textit{Section} category because its weakly supervised annotations exhibit substantial fragmentation and label noise. For each annotated region, we crop
the corresponding visual content together with its bounding-box geometry and
retain its semantic label under the DocBank-based taxonomy. Invalid boxes are
discarded, and elements whose normalized width or height is smaller than
$\epsilon=0.01$ are removed. This filtering suppresses annotation fragments
and extremely thin crops that provide little useful visual or structural
information. The retained instances are organized into category-specific
pools, from which the page-level candidate set is subsequently constructed.

\begin{algorithm}[h]
\caption{Mesh-candidate BestFit Algorithm}
\label{alg:1}
\KwIn{
    Element pool $P$; candidate-set size $K$; maximum layout size $N$;
    filtering threshold $\epsilon$; scale threshold $\tau_s$;
    large-element ratio $\rho$; matching threshold $fr_{\mathrm{thr}}$
}
\KwOut{Generated layout $L$}

\BlankLine
\tcc{\textbf{Stage I: Element Filtering and Scale-aware Candidate Sampling}}
$P \leftarrow \operatorname{Filter}(P,\epsilon)$\;
$P_{\mathrm{large}},P_{\mathrm{small}}
    \leftarrow \operatorname{ScalePartition}(P,\tau_s)$\;

$C_{\mathrm{large}}
    \leftarrow
    \operatorname{Sample}
    (P_{\mathrm{large}},\lfloor\rho K\rfloor)$\;
$C_{\mathrm{small}}
    \leftarrow
    \operatorname{Sample}
    (P_{\mathrm{small}},K-\lfloor\rho K\rfloor)$\;
$C_{\mathrm{set}}
    \leftarrow C_{\mathrm{large}}\cup C_{\mathrm{small}}$\;

\BlankLine
\tcc{\textbf{Stage II: Layout Initialization}}
$e^* \leftarrow \operatorname{RandomSample}(C_{\mathrm{set}})$\;
$L \leftarrow \operatorname{Initialize}(e^*)$\;
$C_{\mathrm{set}}
    \leftarrow C_{\mathrm{set}}\setminus\{e^*\}$\;

\BlankLine
\tcc{\textbf{Stage III: Iterative Mesh Construction and BestFit Matching}}
\While{$|L|<N \ \land\ C_{\mathrm{set}}\neq\emptyset$}{

    \tcp{Construct candidate regions from the current layout}
    $M_{\mathrm{set}} \leftarrow \operatorname{MeshEngine}(L)$\;

    \tcp{Initialize the best element--region match}
    $fr_{\max}\leftarrow 0$\;
    $e_{\mathrm{best}}\leftarrow\varnothing$\;
    $g_{\mathrm{best}}\leftarrow\varnothing$\;

    \tcp{Search for the globally optimal BestFit match}
    \ForEach{$e_i\in C_{\mathrm{set}}$}{
        \ForEach{$g_j\in M_{\mathrm{set}}$}{
            $fr\leftarrow\operatorname{Match}(e_i,g_j)$\;
            \If{$fr>fr_{\max}$}{
                $fr_{\max}\leftarrow fr$\;
                $e_{\mathrm{best}}\leftarrow e_i$\;
                $g_{\mathrm{best}}\leftarrow g_j$\;
            }
        }
    }

    \tcp{Terminate if no valid match is available}
    \If{
        $e_{\mathrm{best}}=\varnothing
        \ \lor\ 
        fr_{\max}<fr_{\mathrm{thr}}$
    }{
        \textbf{break}\;
    }

    \tcp{Place the selected element and update the candidate set}
    $L\leftarrow L\cup
        \left\{
        \operatorname{PlaceCenter}
        (e_{\mathrm{best}},g_{\mathrm{best}})
        \right\}$\;
    $C_{\mathrm{set}}\leftarrow
        C_{\mathrm{set}}\setminus\{e_{\mathrm{best}}\}$\;
}

\BlankLine
\textbf{return} $L$\;
\end{algorithm}

\subsection{Scale-stratified Candidate Sampling}

Directly sampling from the complete element pool may overrepresent small or
narrow elements, causing the page to be filled by fragmented regions before
major content areas establish a stable structure. We therefore partition the
filtered elements into large- and small-scale group according to their
normalized scales. An element is assigned to the large-scale group when
both its normalized width and height are no smaller than $\tau_s=0.05$; all
remaining valid elements form the small-scale group. For each synthetic
page, a fixed-size candidate set of $K=500$ elements is constructed by
independently sampling the two strata with a predefined large-to-small ratio
of $\rho:(1-\rho)$, where $\rho=0.99$. The sampled subsets are then merged,
and one candidate element is randomly drawn to initialize the page. This strategy
preserves fine-grained elements while prioritizing the formation of major
layout regions. Restricting BestFit matching to a fixed-size candidate set
also bounds its search overhead and introduces page-to-page compositional
diversity.

\subsection{Mesh-candidate BestFit Algorithm}

After initialization, Mesh-candidate BestFit iteratively updates the available
page regions and searches for the most compatible element--region pair. Algorithm~\ref{alg:1} summarizes the whole process of Mesh-candidate BestFit. Given
the current layout $L$, \textsc{MeshEngine} constructs a set of non-overlapping
empty grids from the page boundaries and the horizontal and vertical
boundaries of the placed elements. For a candidate element $e_i$ and an empty
grid $g_j$, the matching score is defined as
\begin{equation}
    \operatorname{match}(e_i,g_j)=
    \begin{cases}
        \dfrac{w_i h_i}{w_j h_j},
        & w_i\leq w_j \ \text{and}\ h_i\leq h_j,\\[4pt]
        0, & \text{otherwise},
    \end{cases}
\end{equation}
where $(w_i,h_i)$ and $(w_j,h_j)$ denote the width and height of the element and
grid, respectively. The score favors candidates that occupy the available
region compactly while satisfying the size constraint. At each iteration, the
candidate--grid pair with the highest score is selected, the candidate is
centered within the matched grid, and the mesh is recomputed. Generation stops
when the layout contains $N=35$ elements or no valid pair exceeds
$fr_{\mathrm{thr}}=10^{-4}$. 

Once the layout is finalized, a blank document canvas is created by sampling its width as $W\sim\mathcal{U}_{\mathrm{int}}(1200,2000)$ and its aspect ratio as $r=H/W\sim\mathcal{U}(0.7,1.5)$, with $H$ capped at 3000 pixels. Each placed candidate box is mapped from normalized coordinates to the canvas and independently contracted about its center along the width and height using factors sampled from $\mathcal{U}(0.8,0.95)$. The corresponding element crop is then resized to fit the resulting box and composited onto the page, while its coordinates and semantic label are retained as annotations.





\clearpage

\section{Visual Diagnosis of MeshLayout-1M}
\label{sec:appendix_bestfit_vis}

In this section, we conduct a series of visual analyzes to demonstrate both the diversity of the synthesized pretraining dataset and the effectiveness of our proposed document synthesis method.

\subsection{Synthetic Documents Illustration}

\begin{figure}[h]
    \centering
    \includegraphics[width=1.0\linewidth]{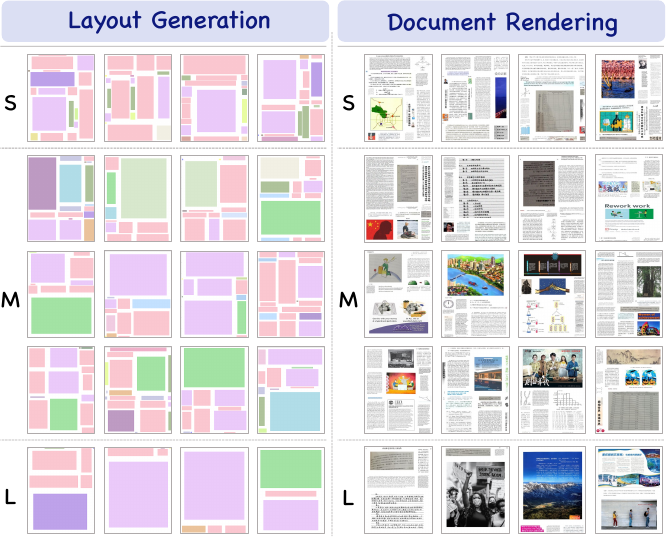}
    \caption{Visualization of generated diverse layouts and corresponding pages after rendering. \textit{S, M, L} respectively denote small, medium, and large elements, indicating the components that are relatively abundant on the page.}
    \label{fig:vis_res}
\end{figure}

Firstly, a richer visualization of the Mesh-candidate BestFit generated document is shown in Fig.~\ref{fig:vis_res}. S, M, L respectively denote small, medium, and large elements, indicating the components that are relatively abundant on the page. It is evident that the data we generate is rich in categories and possesses strong diversity. It can generate dense layouts containing many small elements and produce sparse layouts composed of a few large elements, similar to the layouts generated by diffusion-based models.

\begin{figure*}[t]
    \centering
    \begin{subfigure}{\textwidth}
        \centering
        \includegraphics[width=\textwidth]{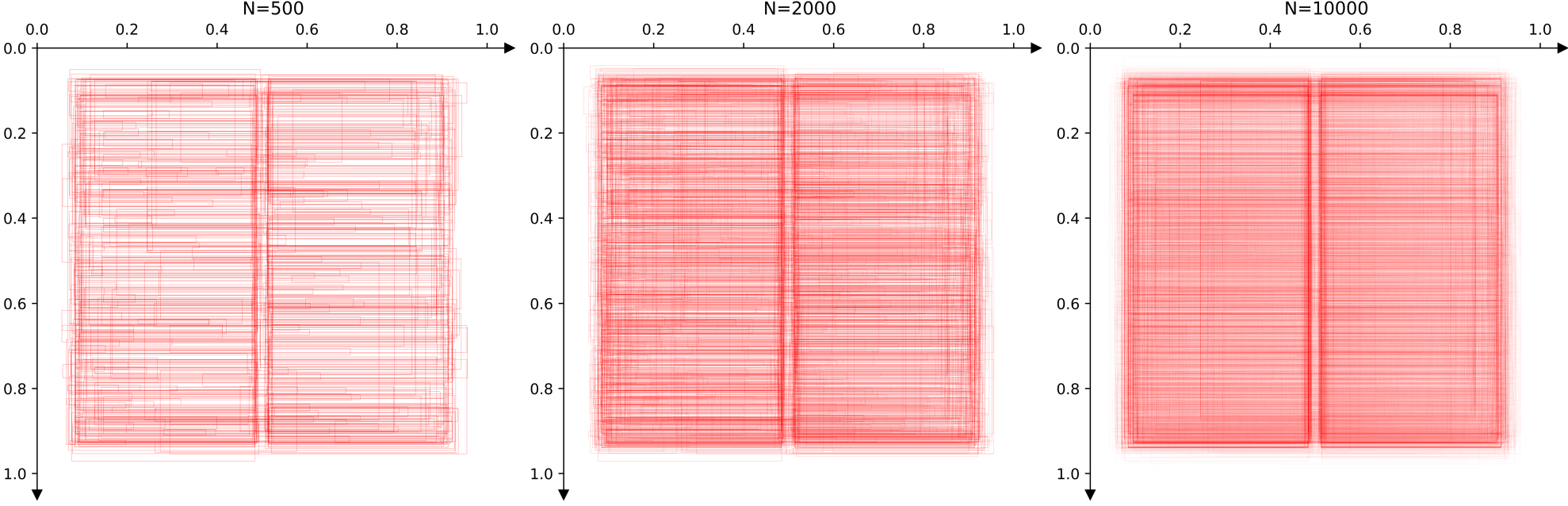}
        \caption{Document layout distribution of PubLayNet.}
        \vspace{35pt}
        \label{fig:layout_pub}
    \end{subfigure}
    \begin{subfigure}{\textwidth}
        \centering
        \includegraphics[width=\textwidth]{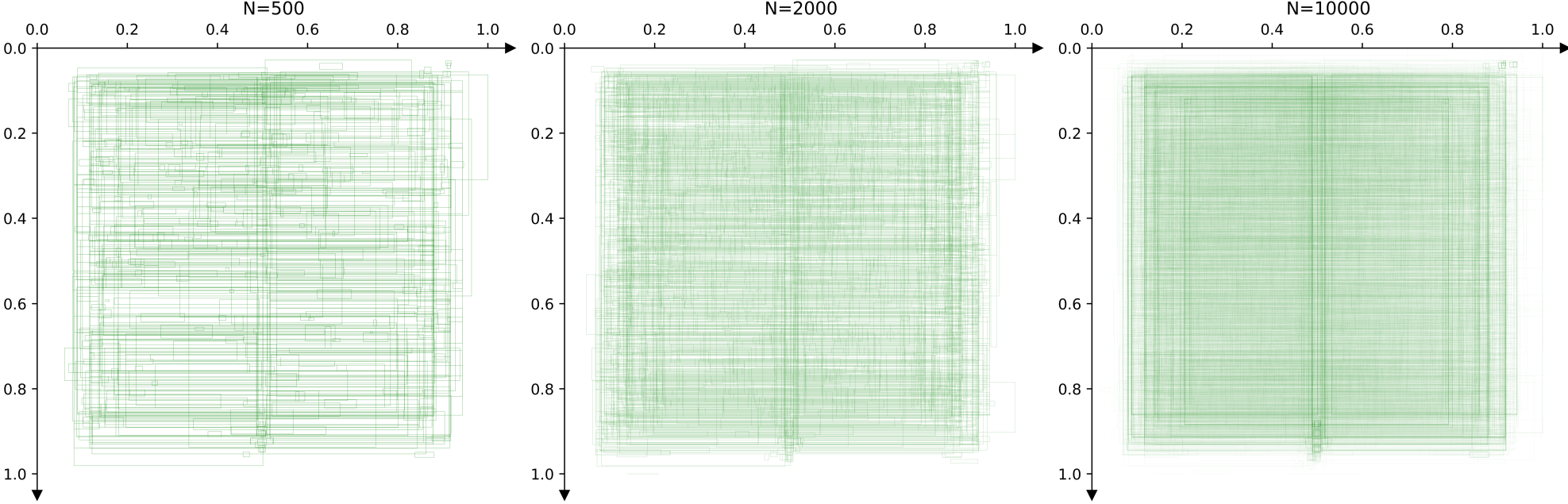}
        \caption{Document layout distribution of DocBank.}
        \vspace{35pt}
        \label{fig:layout_docbank}
    \end{subfigure}
    \begin{subfigure}{\textwidth}
        \centering
        \includegraphics[width=\textwidth]{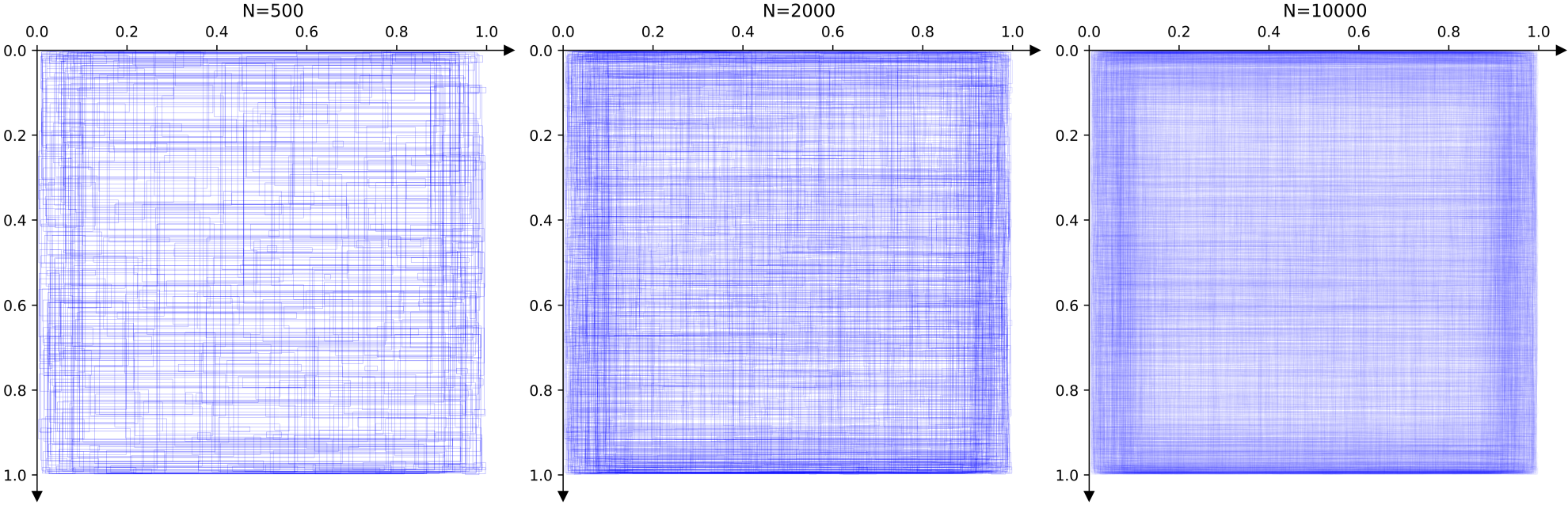}
        \caption{Document layout distribution of MeshLayout-1M.}
        \label{fig:layout_bestfit}
    \end{subfigure}
    \caption{Document layout distribution of (a) PubLayNet, (b) DocBank and (c) MeshLayout-1M. $N$ indicates number of documents used in visualization. Compared with two-column academic style layout in PubLayNet or DocBank, document layout synthesized by Mesh-candidate BestFit in MeshLayout-1M demonstrate significant variation and diversity. }
\label{fig:layout_diversity}
\end{figure*}


\subsection{Layout Diversity}
Next, we visualize and compare the document layout diversity of PubLayNet~(Fig.\ref{fig:layout_pub}), DocBank~(Fig.\ref{fig:layout_docbank}), and MeshLayout-1M~(Fig.\ref{fig:layout_bestfit}). Based on the visualization results, the document layouts in PubLayNet and DocBank clearly exhibit the two-column format and Manhattan layout typical of academic papers, lacking diversity. In contrast, the document layouts generated by Mesh-candidate BestFit demonstrate significant variation and diversity.

\clearpage
\subsection{Element Scale and Shape Distributions}

\begin{figure}[h]
    \centering
    \includegraphics[width=\linewidth]{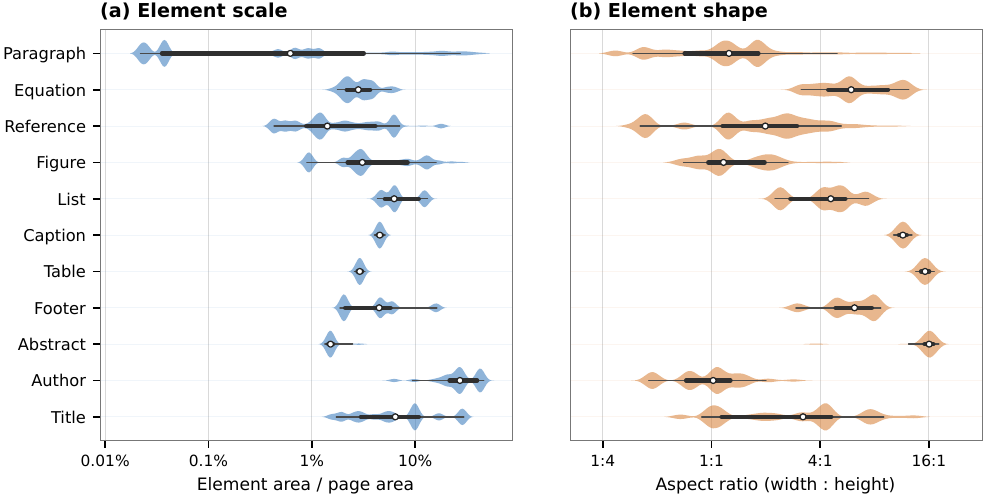}
    \caption{Class-wise element geometry in the complete MeshLayout-1M training split. (a) Bounding-box area relative to page area. (b) Width-to-height ratio.}
    \label{fig:meshlayout_scale_shape}
\end{figure}

We characterize the geometry of all 14,549,846 annotations in the complete MeshLayout-1M training split using normalized axis-aligned bounding boxes. Fig.~\ref{fig:meshlayout_scale_shape} reports the class-wise marginal distributions of relative area and width-to-height ratio. These statistics describe the geometric coverage of the pretraining corpus without assuming that any single class follows a unimodal or canonical template.

\paragraph{Scale coverage.}
Median element area ranges from 0.62\% of the page for Paragraph to 27.15\% for Author, spanning more than a factor of 40 across classes. Paragraph and Title have broad 5th--95th percentile intervals (0.022\%--27.29\% and 1.73\%--29.59\%, respectively), while Caption, Table, and Abstract occupy more concentrated scale bands. The multimodal distributions of frequent classes such as Paragraph and Reference further show that the corpus contains several recurring size regimes rather than a single element scale.

\paragraph{Class-dependent shapes.}
The shape distributions differ substantially across semantic classes. Caption, Table, and Abstract are predominantly horizontal, with median width-to-height ratios of 11.50, 15.26, and 16.08, respectively. In contrast, Figure and Author are approximately square (median ratios 1.17 and 1.02), with 58.1\% and 67.2\% of their instances lying in the near-square range $[2/3,1.5]$. We report these class-specific profiles as properties of the generated corpus; synthetic-to-real correspondence is not inferred from the marginal distributions alone.

\clearpage
\section{Layout Analysis Examples}
\label{sec:det_example_docstructbench}

\begin{figure*}[h]
    \includegraphics[width=1.0\textwidth]{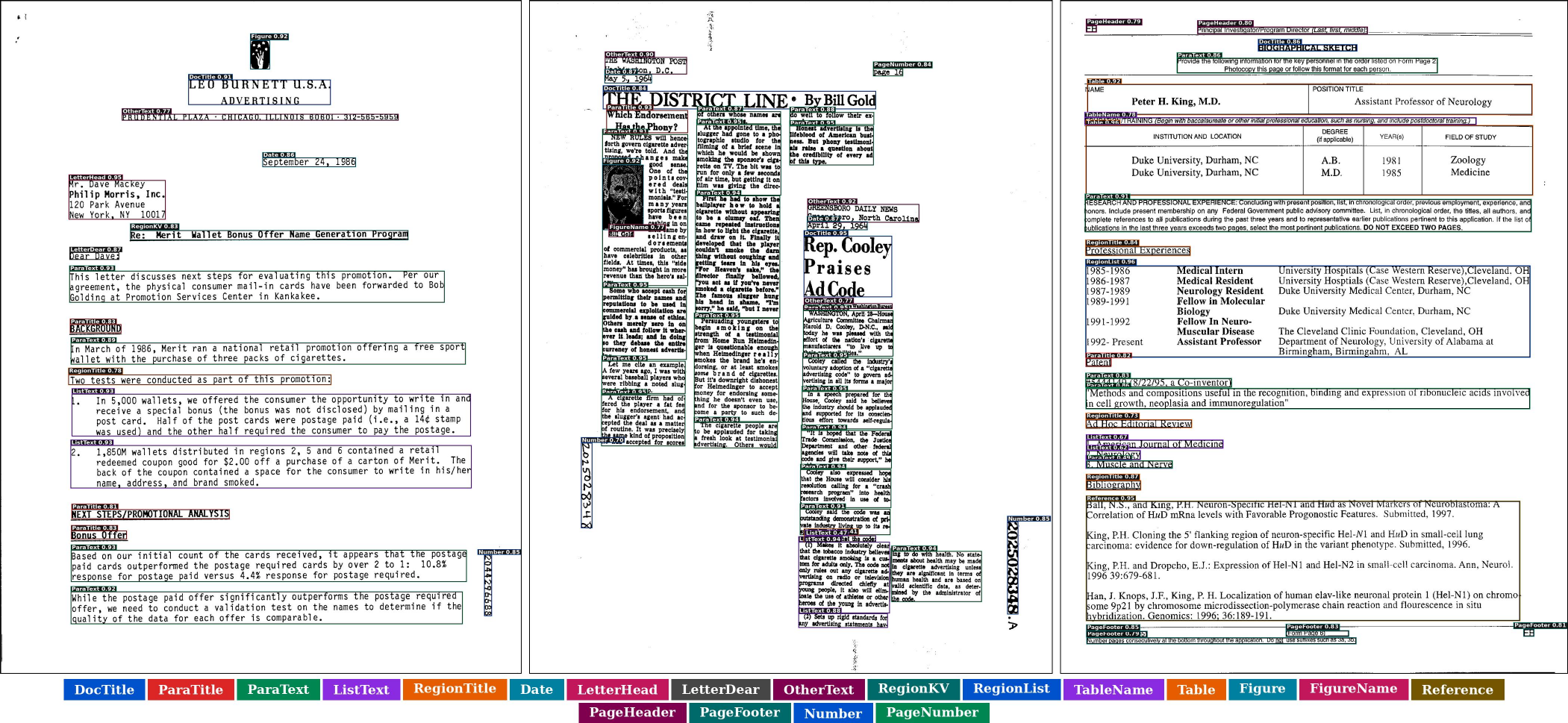}
    \caption{Layout analysis examples on D\textsuperscript{4}LA.}
    \label{det_d4la}
\end{figure*}

\begin{figure*}[h]
    \includegraphics[width=1.0\textwidth]{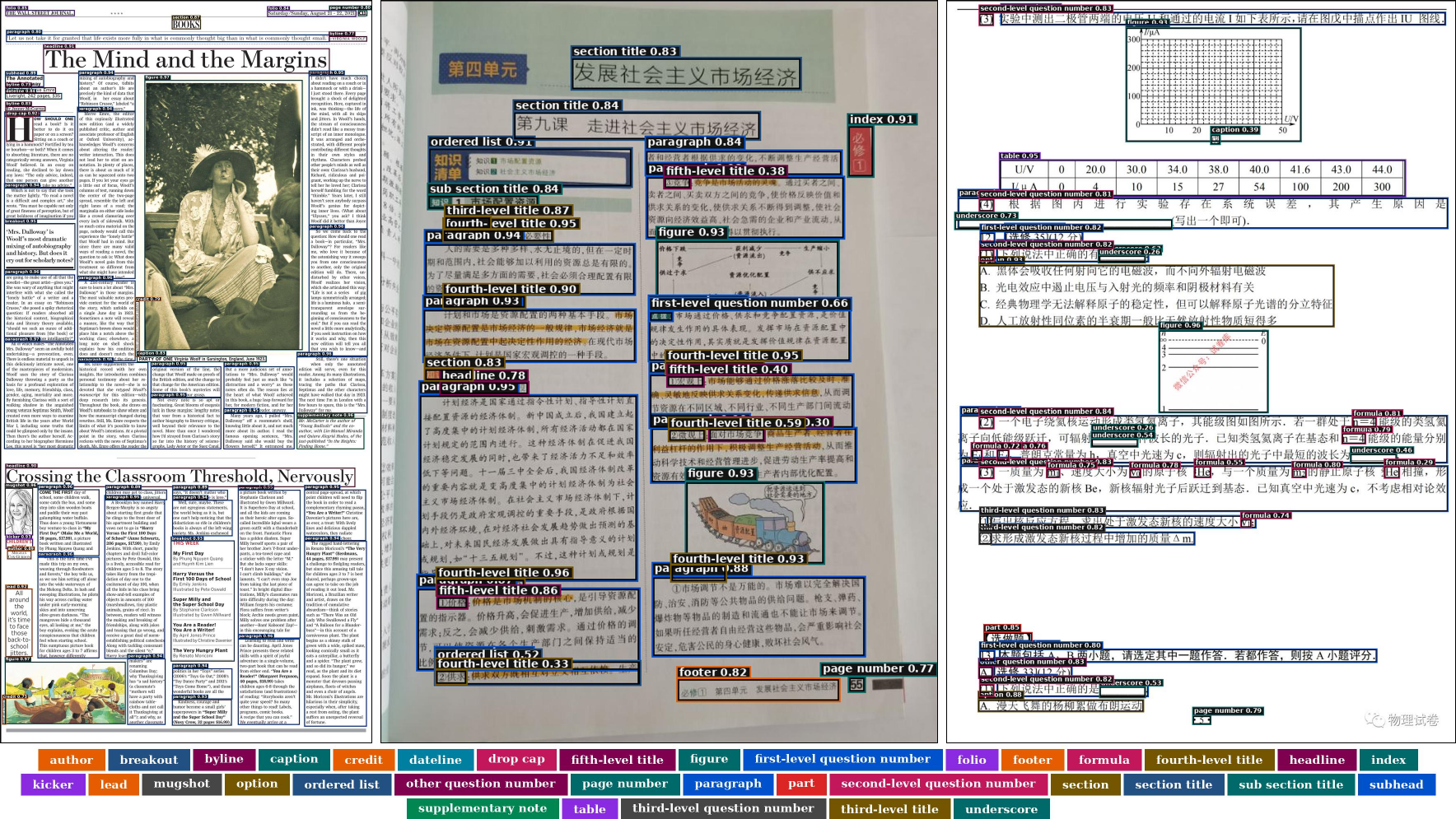}
    \caption{Layout analysis examples on M\textsuperscript{6}Doc.}
    \label{det_m6doc}
\end{figure*}

\begin{figure*}[h]
    \includegraphics[width=1.0\textwidth]{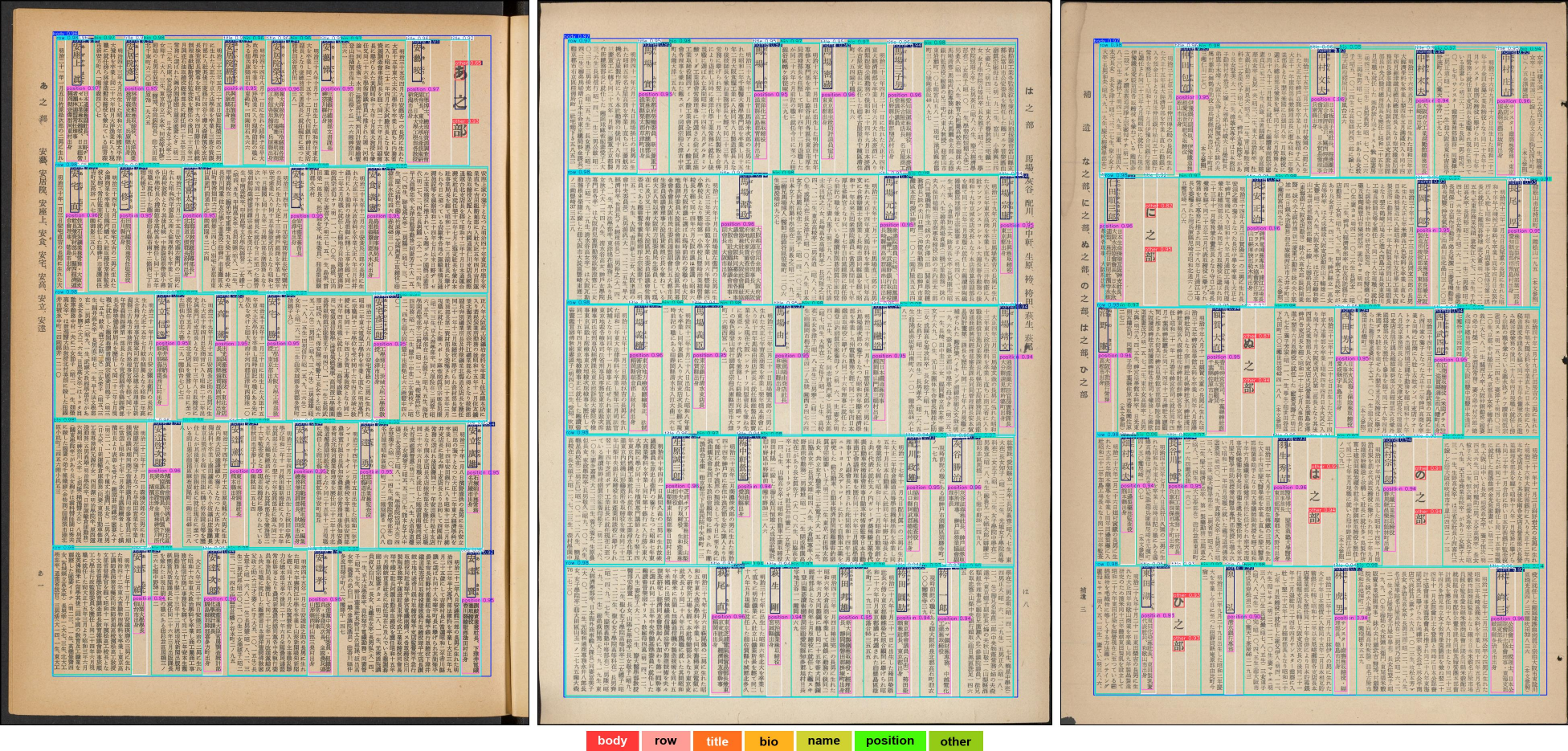}
    \caption{Layout analysis examples on HJDataset.}
    \label{det_hjdataset}
\end{figure*}

\begin{figure*}[h]
    \includegraphics[width=1.0\textwidth]{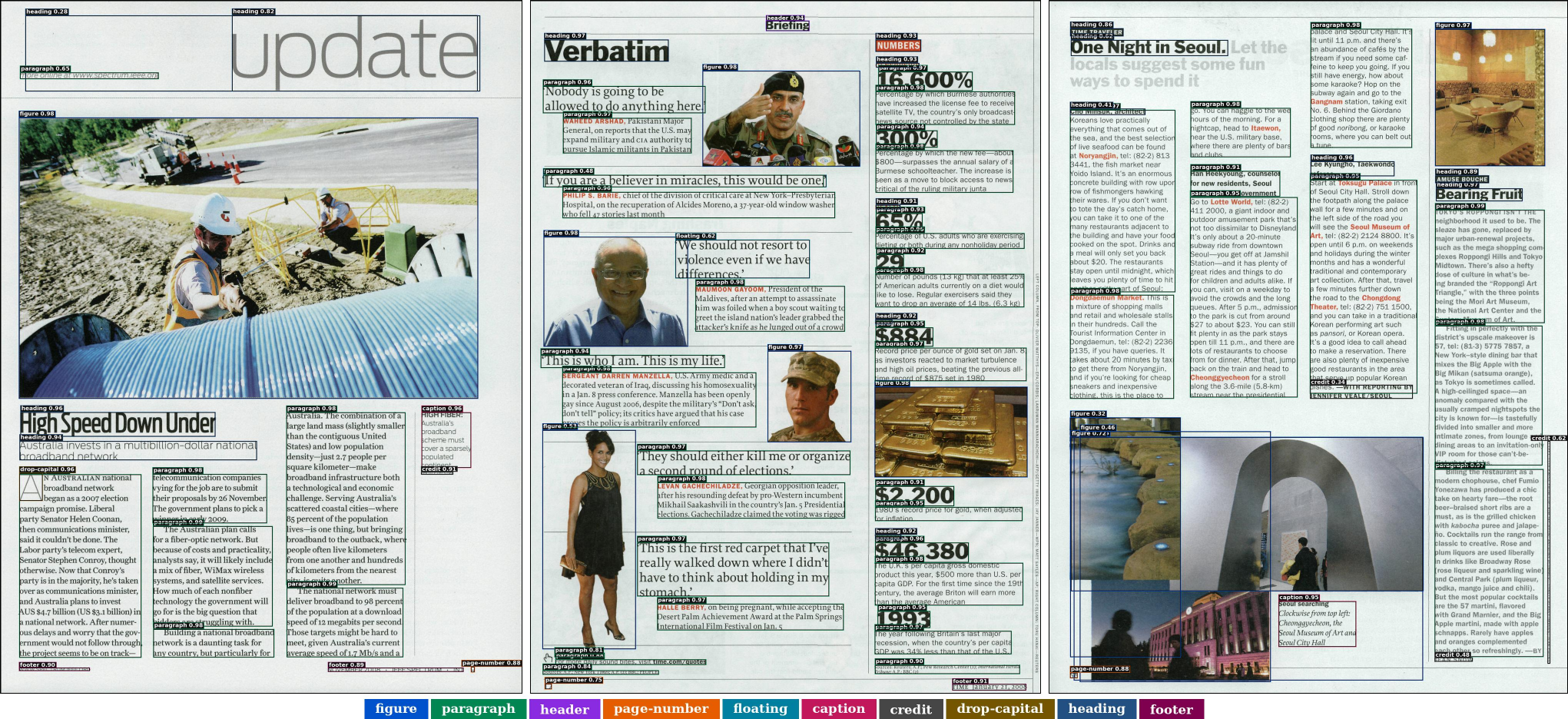}
    \caption{Layout analysis examples on PRIMA-LAD.}
    \label{det_prima}
\end{figure*}

\begin{figure*}[h]
    \includegraphics[width=1.0\textwidth]{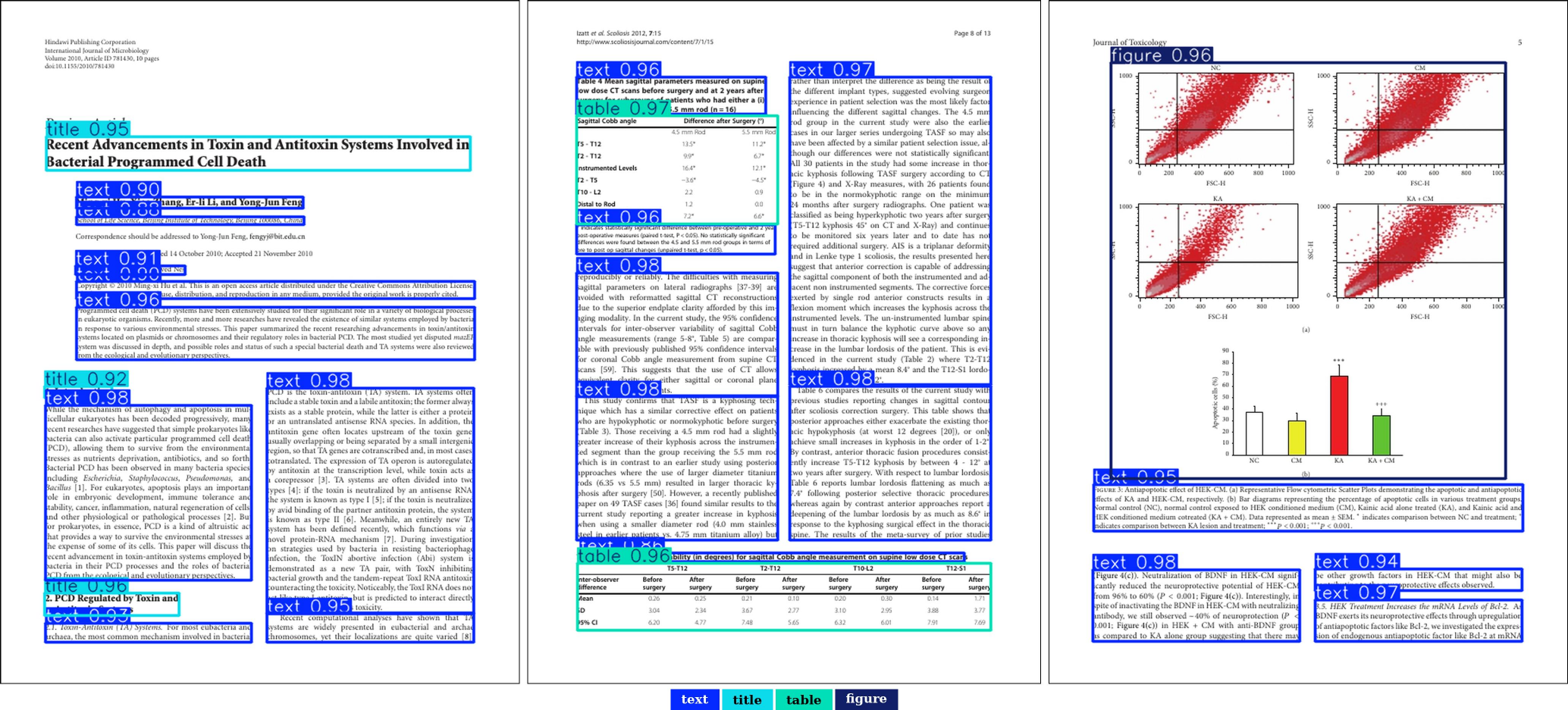}
    \caption{Layout analysis examples on PubLayNet.}
    \label{det_pub}
\end{figure*}

\begin{figure*}[h]
    \includegraphics[width=1.0\textwidth]{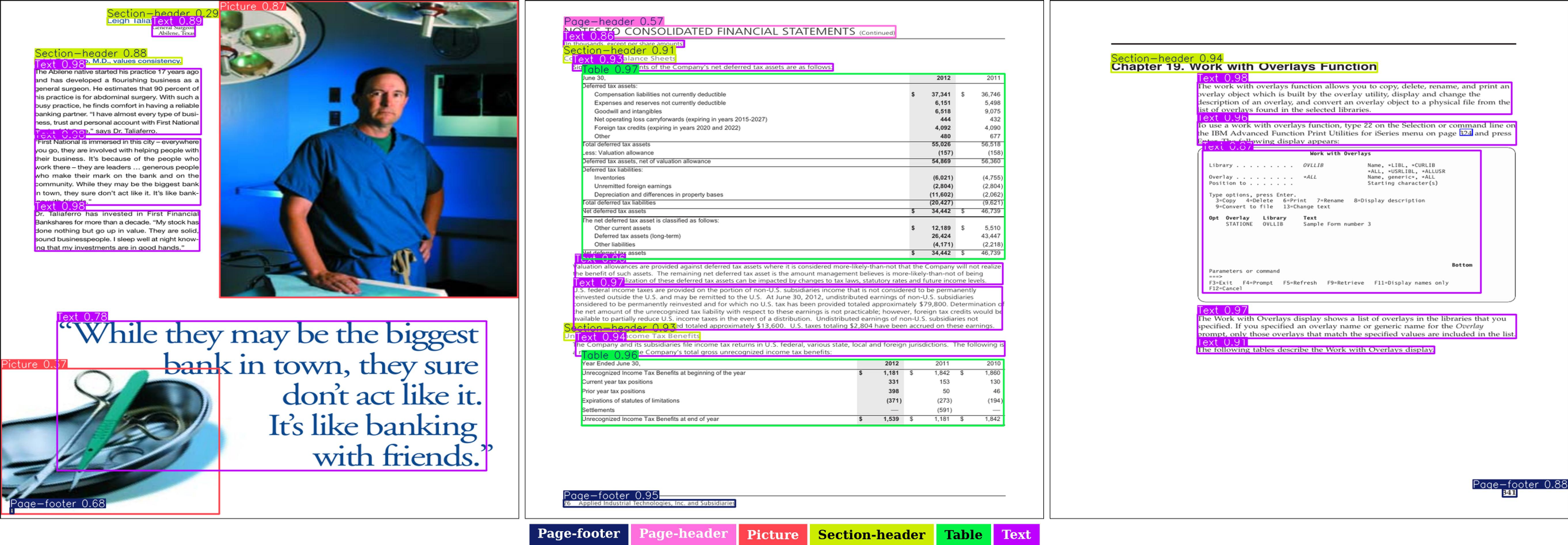}
    \caption{Layout analysis examples on DocLayNet.}
    \label{det_doclaynet}
\end{figure*}

\begin{figure*}[h]
    \includegraphics[width=1.0\textwidth]{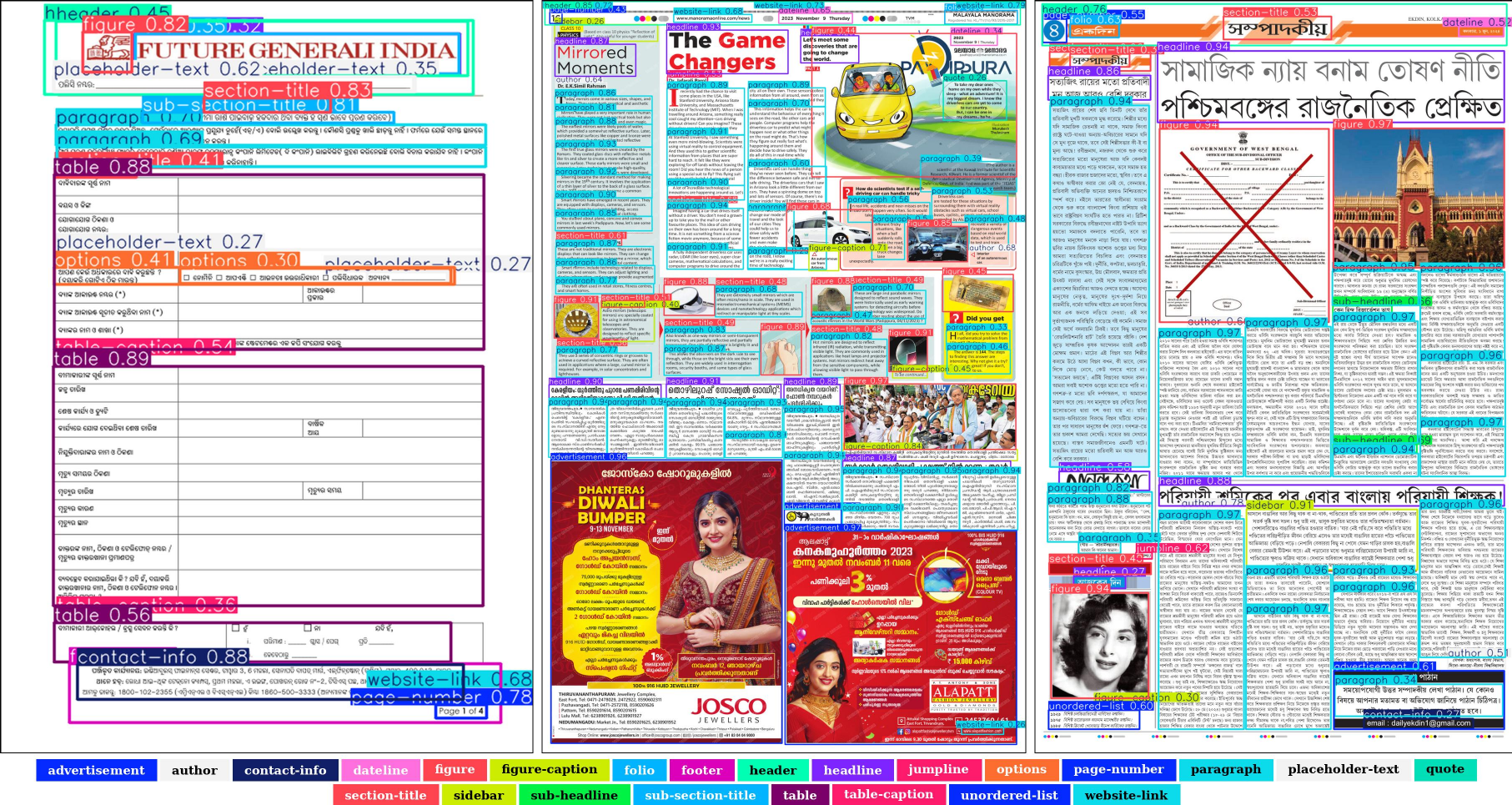}
    \caption{Layout analysis examples on IndicDLP.}
    \label{det_indicdlp}
\end{figure*}

Here, qualitative layout analysis examples on D\textsuperscript{4}LA~(Fig.~\ref{det_d4la}), M\textsuperscript{6}Doc~(Fig.~\ref{det_m6doc}), HJDataset~(Fig.~\ref{det_hjdataset}), PRIMA-LAD~(Fig.~\ref{det_prima}), PubLayNet~(Fig.~\ref{det_pub}), DocLayNet~(Fig.~\ref{det_doclaynet}), IndicDLP~(Fig.~\ref{det_indicdlp}) are illustrated.

\clearpage
\section{Limitations \& Future Work}

\textbf{Efficiency across deployment environments.}
PALA achieves favorable practical efficiency with optimized attention implementations. Its actual speedup may vary across hardware platforms and inference backends due to differences in kernel optimization and memory access. Further operator-level optimization could improve its portability and efficiency across broader deployment environments.

\textbf{Modeling scope of PALA.}
PALA is designed around the dominant horizontal and vertical organization of document layouts, enabling efficient long-range context modeling along both page axes. While this design covers common document structures well, highly irregular layouts with strong rotation, perspective distortion, or non-axis-aligned arrangements may benefit from more flexible interaction patterns. Extending PALA with adaptive directional modeling is a promising direction for such scenarios.

\textbf{Synthesis scope of MeshLayout.}
MeshLayout focuses on increasing structural diversity through controllable geometric layout synthesis, with its current element pool primarily constructed from DocBank. Expanding the element sources could further increase diversity in content, language, and visual appearance. In addition, incorporating richer semantic relationships and realistic document variations could further narrow the gap between synthetic and real-world documents.

\textbf{Future Work.}
Future work will explore more portable implementations of PALA and adaptive context modeling for irregular document structures. For MeshLayout, we plan to expand multilingual and multi-source element pools and incorporate richer semantic and visual variations, further improving the diversity and realism of synthetic document pretraining.

\end{document}